%% file: acl_latex.tex
\documentclass[11pt]{article}

\usepackage[final]{acl}

\usepackage{times}
\usepackage{latexsym}

\usepackage[T1]{fontenc}

\usepackage[utf8]{inputenc}

\usepackage{microtype}

\usepackage{inconsolata}
\usepackage{caption}
\usepackage{subcaption}
\usepackage{graphicx}
\usepackage{amsmath}
\usepackage{booktabs}

\usepackage{longtable}
\title{Transferring Textual Preferences to Vision-Language Understanding through Model Merging}

\author{Chen-An Li \quad Tzu-Han Lin \quad Yun-Nung Chen \quad  Hung-yi Lee \\
National Taiwan University, Taipei, Taiwan \\
\texttt{\{r13942069,r12944034\}@ntu.edu.tw} \quad \texttt{y.v.chen@ieee.org} \quad \texttt{hungyilee@ntu.edu.tw}
}

\begin{document}
\maketitle

\begin{abstract}
Large vision-language models (LVLMs) perform outstandingly across various multimodal tasks. However, their ability to evaluate generated content remains limited, and training vision-language reward models (VLRMs) with preference data is computationally expensive. This paper explores a training-free alternative by merging text-based reward models (RMs) with LVLMs to create VLRMs. Our approach shows that integrating these models leads to improved performance over LVLMs' scoring and text-based RMs, offering an efficient method for incorporating textual preferences into LVLMs. The code and data are publicly available at \url{https://github.com/lca0503/MergeToVLRM}.
\end{abstract}

\section{Introduction}
\label{sec:introduction}
Large vision-language models (LVLMs) have shown exceptional performance across a wide range of multimodal tasks~\citep{hurst2024gpt, team2024gemini, anthropic2024claude35}, primarily due to the implementation of reinforcement learning from human feedback (RLHF)~\citep{ouyang2022training}, which utilizes preference data~\citep{sun-etal-2024-aligning, li-etal-2024-vlfeedback}. This process often requires the use of reward models (RMs). However, LVLMs still struggle to assess generated content effectively~\citep{chen2024mllmasajudge, li2024vlrewardbench}, and training an RM with preference data is resource-intensive.

In this work, we investigate an alternative approach: \textit{Can knowledge derived from text-only preference data be transferred to LVLMs without additional training?} Several state-of-the-art LVLMs are built upon pre-trained language models with vision encoders and adapters~\citep{dubey2024llama, Qwen2.5-VL, lu2024deepseek}. This architectural design suggests that textual preferences learned by text-based RMs may potentially integrate into LVLMs through parameter merging.

\begin{figure}[t!]
    \centering    
    \includegraphics[width=0.89\linewidth]{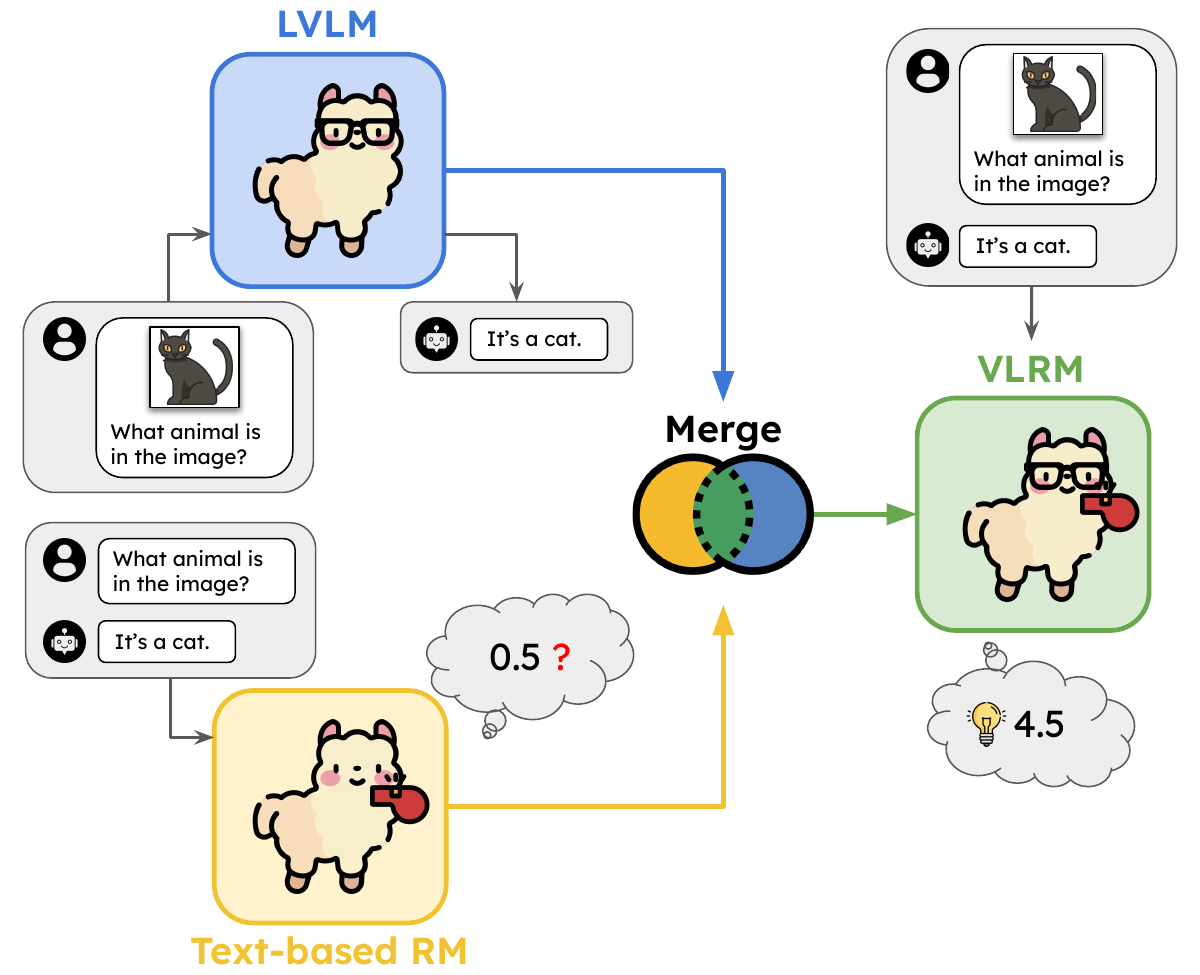}
    \vspace{-5pt}
    \caption{Framework for merging a text-based RM with an LVLM. LVLMs excel at visual tasks, while text-based RMs struggle to provide accurate rewards without visual cues. We transfer textual preferences to the vision-language understanding, resulting in a VLRM. All icons used in this figure are sourced from \url{https://www.flaticon.com/}}
    \vspace{-10pt}
    \label{fig:overview}
\end{figure}

Building on this idea, we propose merging LVLMs with text-based RMs to create vision-language reward models (VLRMs), as illustrated in Figure~\ref{fig:overview}. Our approach leverages existing RMs and LVLMs, eliminating the need for costly multimodal preference data collection and training. We explore various merging strategies, ranging from simple weighted averaging~\citep{wortsman2022model} to advanced techniques such as task arithmetic~\citep{ilharco2023editing}, TIES~\citep{yadav2024ties}, and DARE~\citep{yu2024language}.

We assess performance using VL-RewardBench \citep{li2024vlrewardbench} and Best-of-N sampling with TextVQA~\citep{singh2019towards} and MMMU-Pro~\citep{yue2024mmmu}. The results show that our combined VLRMs outperform scoring through LVLMs and reward generation with text-based RMs. Our approach offers a training-free method for transferring textual preferences to LVLMs via model merging, and we provide a detailed analysis of merging strategies, demonstrating its effectiveness across multiple benchmarks.

\section{Related Work}
\label{sec:related_work}
\paragraph{Preference Dataset} A common approach to train a reward model is to use the Bradley–Terry model~\citep{bradley1952rank}, which relies on paired data for learning. In NLP, many high-quality preference datasets are already available~\citep{learningtosummarize,bai2022traininghelpfulharmlessassistant,pmlr-v162-ethayarajh22a,oasst,cui2024ultrafeedback,zhu2024starlingb,wang2024helpsteer}. Similarly, in the vision-language domain, several preference datasets have been introduced~\citep{Yu_2024_CVPR,yu2024rlaifv,chen2024sharegpt4v,wijaya2024multimodalpreferencedatasynthetic,li2024vlfeedback,zhou2024aligning,xiao2024detecting}. In this work, we explore the potential of transferring textual preferences to LVLMs in a training-free manner, specifically through model merging.


\paragraph{LVLM-as-a-Judge \& Evaluation} LVLM-as-a-Judge refers to utilizing strong large vision-language models for evaluation and judgment. These LVLMs can be either closed-source~\citep{openai2023gpt4v,hurst2024gpt,team2024gemini,anthropic2024claude35} or open-source~\citep{lee-etal-2024-prometheus,dubey2024llama,deitke2024molmo,Qwen2.5-VL}. To assess LVLMs as generative reward models, \citet{chen2024mllmasajudge} established benchmarks and found that LVLMs exhibit high agreement with humans in pairwise comparison judgments, but perform poorly in scoring evaluation and batch ranking tasks. Recently, VL-RewardBench~\citep{li2024vlrewardbench} introduced challenging cases and complex multimodal reasoning tasks, revealing that most off-the-shelf LVLMs struggle with such evaluations.


\paragraph{Model Merging} Model merging is a common, training-free method for combining skills from multiple models within the parameter space. A basic approach involves simple weighted averaging~\citep{wortsman2022model}, while more advanced techniques have been developed~\citep{yadav2024ties,yu2024language,yang2024modelmergingllmsmllms}. These techniques have already proven effective in reward modeling~\citep{rame2024warm,lin-etal-2024-dogerm} and LLM-as-a-judge~\citep{kim-etal-2024-prometheus} in NLP. Recently, REMEDY~\citep{zhu2025remedy} introduced strategies for merging LVLMs. In contrast, our work focuses on merging textual reward models into the language modeling components of LVLMs.

\section{Methodology}
\label{sec:methodology}
We propose a training-free method to transfer textual preferences from a text-based RM $\theta^{\text{RM}}$ to a LVLM $\theta^{\text{LVLM}}$ through model merging.

Since both models originate from the same pre-trained language model $\theta^{\text{PRE}}$, we merge modules that appear in both models and preserve the LVLM’s vision capabilities and text-based RM reward function, resulting in a VLRM that can assess textual and visual content without additional training. Below, we outline the components and merging strategies involved.

\subsection{Model Components}
\input{tables/main_tulu25}

The pre-trained language model consists of:
\begin{equation*}
\theta^{\text{PRE}} = \{\theta^{\text{PRE}}_{\text{emb}}, \theta^{\text{PRE}}_{\text{trans}}, \theta^{\text{PRE}}_{\text{lm}}\},
\end{equation*}
where $\theta^{\text{PRE}}_{\text{emb}}$ is the embedding layer, $\theta^{\text{PRE}}_{\text{trans}}$ is the transformer, and $\theta^{\text{PRE}}_{\text{lm}}$ is the language modeling head, which maps the final hidden state of the transformer to the vocabulary.

The LVLM expands upon this with:
\begin{equation*}
\theta^{\text{LVLM}} = \{\theta^{\text{LVLM}}_{\text{venc}}, \theta^{\text{LVLM}}_{\text{adapt}}, \theta^{\text{LVLM}}_{\text{emb}}, \theta^{\text{LVLM}}_{\text{trans}}, \theta^{\text{LVLM}}_{\text{lm}}\},
\end{equation*}
where $\theta^{\text{LVLM}}_{\text{venc}}$ is the vision encoder, and $\theta^{\text{LVLM}}_{\text{adapt}}$ is the adapter that integrates the vision encoder outputs into the language model.

Similarly, the text-based RM is defined as:
\begin{equation*}
\theta^{\text{RM}} = \{\theta^{\text{RM}}_{\text{emb}}, \theta^{\text{RM}}_{\text{trans}}, \theta^{\text{RM}}_{\text{rm}}\},
\end{equation*}
where $\theta^{\text{RM}}_{\text{rm}}$ is the reward modeling head, which projects the transformer's final hidden state to a scalar value as the reward for a given input.

\subsection{Merging Strategies}

We explore four merging strategies.
\paragraph{Weighted Averaging}
The weighted averaging strategy is defined as:
\begin{equation*}
    \theta^{\text{MERGE}}_{\text{trans}} = \lambda \cdot \theta^{\text{LVLM}}_{\text{trans}} + (1 - \lambda) \cdot \theta^{\text{RM}}_{\text{trans}},
\end{equation*}
where $\lambda$ is a hyperparameter that controls the weight distribution between the two terms.
\paragraph{Task Arithmetic}
Task arithmetic strategy is defined as:
\begin{equation*}
    \begin{aligned}
        &\tau^{\text{LVLM}} = \theta^{\text{LVLM}}_{\text{trans}} - \theta^{\text{PRE}}_{\text{trans}}, \\
        &\tau^{\text{RM}} = \theta^{\text{RM}}_{\text{trans}} - \theta^{\text{PRE}}_{\text{trans}}, \\
        &\theta^{\text{MERGE}}_{\text{trans}} = \theta^{\text{PRE}}_{\text{trans}} + \lambda \cdot \tau_{\text{LVLM}} + \lambda \cdot \tau_{\text{RM}},
    \end{aligned}
\end{equation*}
where $\tau^{\text{LVLM}}$ represents the task vector derived from instruction tuning, and $\tau^{\text{RM}}$ is the task vector obtained from reward modeling. The hyperparameter $\lambda$ controls the contribution of the task vectors.

\paragraph{TIES \& DARE}
For the TIES and DARE strategies, we simplify the expression to:
\begin{equation*}
    \theta^{\text{MERGE}}_{\text{trans}} = \theta^{\text{PRE}}_{\text{trans}} + \lambda \cdot f(\tau^{\text{LVLM}}, d) + \lambda \cdot f(\tau^{\text{RM}}, d),
\end{equation*}
where $f(\cdot)$ denotes the function for trimming, selecting, and rescaling the task vector, and $d$ is the density determining how many parameters are retained. The two strategies apply different methods for trimming, selecting, and rescaling. See Appendix~\ref{sec:appendix:merging_details} for more details on TIES and DARE.

\subsection{Merged VLRM}

The merged embedding parameters, $\theta^{\text{MERGE}}_{\text{emb}}$ are obtained following standard embedding merging techniques outlined in MergeKit~\citep{goddard-etal-2024-arcees}, as detailed in Appendix~\ref{sec:appendix:merging_details}. 

Finally, the merged VLRM $\theta^{\text{MERGE}}$ is obtained by combining several components:
\begin{equation*}
    \theta^{\text{MERGE}} = \{\theta^{\text{LVLM}}_{\text{venc}}, \theta^{\text{LVLM}}_{\text{adapt}}, \theta^{\text{MERGE}}_{\text{emb}}, \theta^{\text{MERGE}}_{\text{trans}}, \theta^{\text{RM}}_{\text{rm}}\},
\end{equation*}
As a result, the merged VLRM can be used to provide rewards for both text and image content.

\section{Experiments}
\label{sec:experiments}

\subsection{Experimental Setup}
\label{subsec:experimental_setup}

\subsubsection{Models}
\label{subsubsec:models}
In this paper, we employ \texttt{Llama-3.2-11B-Vision\\-Instruct}~\citep{dubey2024llama} as our LVLM, referred to as \texttt{Llama-3.2-Vision}. For text-based RMs, we use \texttt{Llama-3.1-Tulu-2-8B-uf-mean-\\rm}~\citep{ivison2024unpacking} and \texttt{Llama-3.1-Tulu-3-\\8B-RM}~\citep{lambert2024t}, which we denote as \texttt{Tulu-2.5-RM} and \texttt{Tulu-3-RM}, respectively. All models derive from the same pre-trained language model \texttt{Llama-3.1-8B}. Our main results focus on \texttt{Tulu-2.5-RM} since it outperforms \texttt{Tulu-3-RM} on several VQA tasks with text-based input. Please refer to Appendix~\ref{sec:appendix:open_soruce_details} for the model details.

\subsubsection{Model Merging}
\label{subsubsec:model_merging}
We use MergeKit for model merging and apply several techniques: weighted averaging, task arithmetic, TIES, and DARE—labeled as \texttt{Linear}, \texttt{Task Vec.}, \texttt{TIES}, and \texttt{DARE}, respectively. Additionally, we explore combining DARE with task arithmetic and TIES for a more thorough analysis. To determine the optimal merging hyperparameters, we conduct a hyperparameter search and sample 400 instances from the RLAIF-V~\citep{yu2024rlaifv} training set as our validation set. More details are provided in Appendix~\ref{sec:appendix:merging_details}.

\subsection{Reward Model Evaluation}
\label{subsec:evaluation}

\subsubsection{VL-RewardBench}
\label{subsubsec:vl_rewardbench}
We assess the merged VLRMs using VL-RewardBench~\citep{li2024vlrewardbench}, a benchmark that includes three domains: general multimodal instructions, hallucination-related tasks, and multimodal reasoning tasks. Each instance includes a multimodal query that consists of an image and a user prompt, along with a chosen response and a rejected response.

\subsubsection{Best-of-N Sampling}
\label{subsubsec:best_of_n}
We assess our reward model's effectiveness in enhancing performance through reranking using Best-of-N sampling, where N = 8 in our work. This method scores and ranks responses to check if the highest-scoring one matches the correct answer. Specifically, we use \texttt{Llama-3.2-11B-Vision-Instruct} to generate eight candidates for the TextVQA~\citep{singh2019towards} and MMMU-Pro~\citep{yue2024mmmu} datasets. See Appendix~\ref{sec:appendix:dataset_details} for dataset details.

\subsection{Main Results}
\label{subsec:results}
Table~\ref{tab:main_tulu25} demonstrates the effectiveness of merging methods for combining an LVLM with a text-based RM. The baseline approaches include \texttt{Llama-3.2-Vision}, which utilizes the LVLM for direct scoring—pairwise scoring in VL-RewardBench and verbalized scoring in Best-of-N sampling tasks. Another baseline method, \texttt{Tulu-2.5-RM}, utilizes the text-based RM that focuses solely on evaluating the textual elements of questions and responses. We also incorporate a \texttt{Random} baseline that randomly selects responses. Furthermore, we implement a \texttt{Cascade} approach that employs a two-stage process: it first uses the LVLM to generate text descriptions of images based on the given question, then passes these descriptions with the original text inputs through the text-based RM to produce final scores.

As shown in Table~\ref{tab:main_tulu25}, merged VLRMs consistently outperform \texttt{Llama-3.2-Vision} and \texttt{Tulu-2.5-RM} across nearly all merging methods and benchmarks. This result demonstrates that combining a text-based RM with an LVLM effectively transfers textual preferences without training. Different merging strategies achieve the highest scores in different benchmarks, but overall, more advanced methods outperform simpler ones, highlighting the advantages of structured merging techniques. 
Additionally, in several benchmarks, merged VLRMs surpass or match the strong \texttt{Cascade} baseline, suggesting that model merging captures more information than merely cascading two models. Furthermore, as shown in Table~\ref{tab:vlrb_sub}, our merged VLRMs even exceed the performance of the 90B LVLM and achieve results comparable to commercial models.
A similar trend emerges when using \texttt{Tulu-3-RM} as the text-based RM; further details are provided in Appendix~\ref{subsec:appendix:main_results}.
\input{tables/VLRB_sub}

\subsection{Analysis}
\label{subsec:analysis}

\paragraph{Without Image Input}
\label{paragraph:without_image_input}
\input{tables/cmp_img_small_tulu25}
To further investigate whether the merged VLRMs effectively use the vision encoder, we conduct an ablation study by evaluating the models without image input. As shown in Table~\ref{tab:cmp_image_small_tulu25}, most models with image input outperform those without it across various merging techniques. This result suggests that the vision encoder plays an active role after merging, with performance gains not solely attributed to the text-based RM. These findings highlight how merging methods effectively combine textual and visual information. However, image input does not improve performance in the MMMU-Pro Standard set, likely because this set emphasizes reasoning, where reward assessments depend more on textual coherence than visual understanding, limiting the vision encoder's contribution. A similar trend occurs when using \texttt{Tulu-3-RM} as the text-based RM; see Appendix~\ref{subsec:appendix:without_image_input} for details.

\begin{figure}[t]
    \centering
    \begin{subfigure}[t!]{0.49\linewidth}
        \centering
        \includegraphics[width=\textwidth]{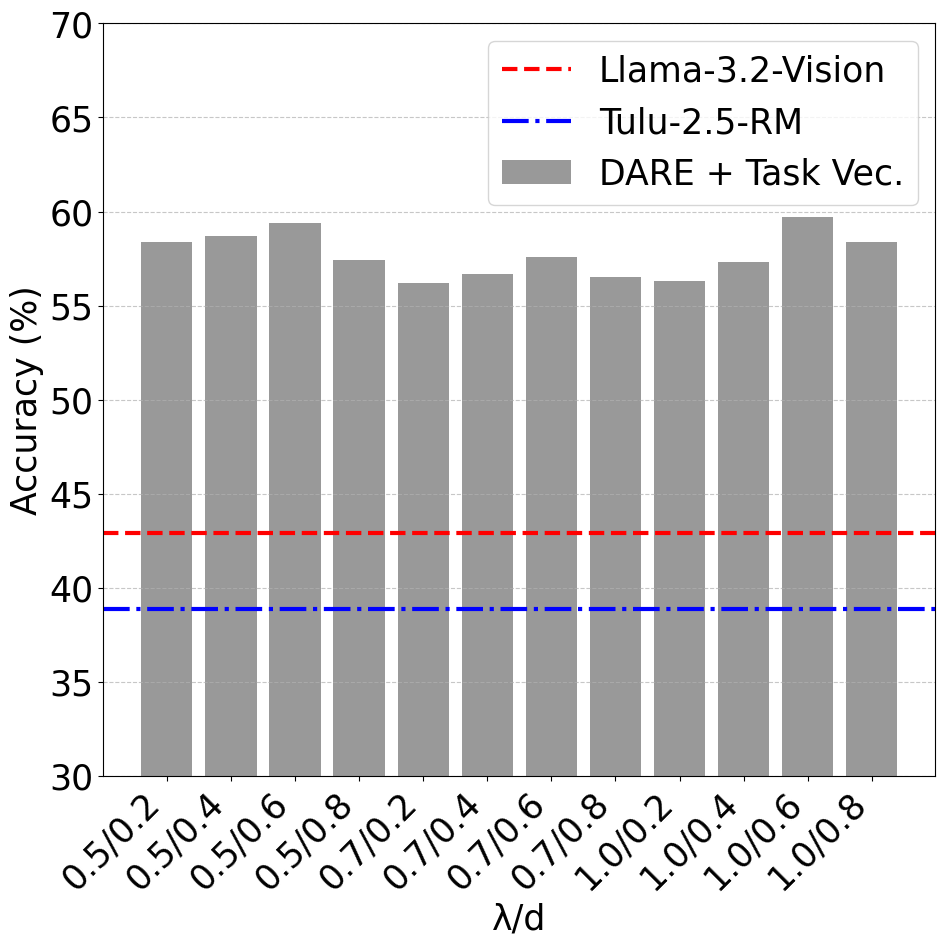}
        \caption{VL-RewardBench}
        \label{fig:daretv_vlrb}
    \end{subfigure}
    \hfill
    \begin{subfigure}[t!]{0.49\linewidth}
        \centering
        \includegraphics[width=\textwidth]{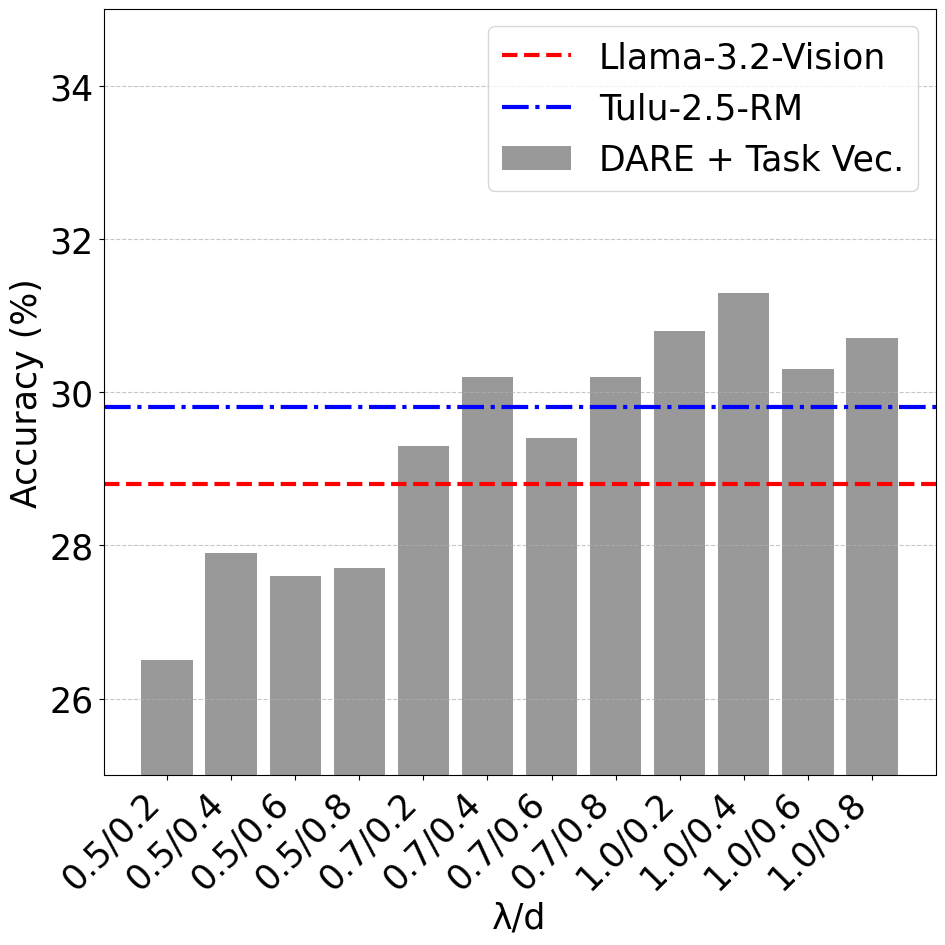}
        \caption{MMMU-Pro (Standard)}
        \label{fig:daretv_mps}
    \end{subfigure}
    \vspace{-3pt}
    \caption{Effect of \texttt{Dare + Task Vec.} merging hyperparameters with \texttt{Tulu-2.5-RM} as the text-based RM.}
    \vspace{-8pt}
    \label{fig:hyperparameters_effect}
\end{figure}

\paragraph{Effect of Merging Hyperparameters}
\label{paragraph:hyperparameters_effect}
We also investigate how merging hyperparameters impacts performance. Figure~\ref{fig:hyperparameters_effect} presents the results of searching for $d$ within the range [0.2, 0.4, 0.6, 0.8] and $\lambda$ within [0.5, 0.7, 1.0] for \texttt{DARE} + \texttt{Task Vec.}. 
Our findings indicate that optimal hyperparameter values vary across benchmarks. For example, in VL-RewardBench, $\lambda$ values do not have a significant effect, but in the MMMU-Pro standard set, we observe that $\lambda=1.0$ performs best.
This variation indicates that the choice of hyperparameters affects the performance of the final merged VLRM differently across tasks. Consequently, it highlights the importance of a well-curated validation set when selecting the optimal hyperparameters, which could be further explored in future research. 

Furthermore, our results for $d$ align with previous studies on TIES and DARE: even when task vectors are trimmed to lower rates (e.g., 0.4, 0.2), the merged VLRMs maintain strong performance, consistent with the findings on LLM merging. For further hyperparameter search results across other methods and benchmarks, refer to Appendix~\ref{subsec:appendix:hyperparameters_effect}.

\paragraph{Computation Overhead}
\label{paragraph:computation_overhead}
In our experiments, model merging is done entirely on CPUs (Intel Xeon Silver 4216) using a system with 128 GB of RAM. Using 11 different $\lambda$ values for weighted averaging takes about 1.5 hours of CPU time. The task arithmetic method takes a similar amount of time when using the same number of $\lambda$ values. Applying 12 combinations of $\lambda$ and density $d$ for the TIES method takes about 6 hours of CPU time, while DARE takes around 3 hours to handle the same number of combinations.

We evaluate the models on a validation set of 400 examples from the RLAIF-V dataset. We run model inference on GPUs with 24 GB of memory (Nvidia GeForce RTX 3090). Across all configurations and merging methods, inference takes approximately 1.5 hours of GPU time per method.

Overall, merging and evaluation require much less computing time than training a reward model from scratch. Since merging is the most time-consuming step and runs only on the CPU, the total computational cost stays relatively low. Also, both merging and evaluation can be run in parallel on multiple machines to reduce the actual runtime.

\section{Conclusion}
\label{sec:conclusion}
This work presents a training-free approach for integrating text-based RMs into LVLMs through model merging. Our method enables the efficient transfer of textual preferences without the expensive multimodal preference data collection or additional training. Experimental results show that our approach outperforms LVLM scoring and text-based RMs in multimodal reward assessment tasks. 

\section*{Limitations}
\label{sec:limitations}
Our study has several limitations. First, we focused on a specific 11B vision-language model paired with an 8B text-based reward model, primarily due to limitations in computational resources. Additionally, we focused solely on the LLaMA architecture and did not explore alternatives like Qwen~\citep{bai2023qwen,bai2023qwenvl} due to the absence of a suitable Qwen-based reward model for our experiments. Furthermore, we did not perform extensive ablation studies on the validation set. Our experimental results highlight the importance of a well-curated validation set in selecting optimal hyperparameters, which could be explored further in future research. Finally, due to the sensitivity of RLHF to hyperparameter tuning and our computational constraints, we did not implement algorithms like PPO~\citep{schulman2017proximal}. Future work could explore integrating RLHF with merged VLRMs to assess its potential impact.

\section*{Ethics Statement}
\label{sec:ethics_statement}
Our approach leverages pre-trained language and reward models, which may inherit biases from the training data. While merging models can enhance efficiency, it does not inherently mitigate existing biases. We encourage further research to evaluate and address potential biases in merged models to ensure fairness across diverse user groups. 

\section*{Acknowledgements}
We thank the reviewers for their insightful comments. This work was financially supported by the National Science and Technology Council (NSTC) in Taiwan, under Grant 113-2628-E-002-033. We thank to National Center for High-performance Computing (NCHC) of National Applied Research Laboratories (NARLabs) in Taiwan for providing computational and storage resources. We are also grateful to Yu-Xiang Lin from National Taiwan University for his valuable advice and thoughtful discussions.

\bibliography{custom}

\appendix

\section{Merging Details}
\label{sec:appendix:merging_details}

\paragraph{Weighted Averaging}
\label{sec:appendix:paragraph:weighted_averaging}
\citet{wortsman2022model} showed that combining the weights of multiple models fine-tuned with varying hyperparameter settings often leads to improved accuracy and robustness. In this work, we employ a weighted averaging strategy as a straightforward method to merge a large vision-language model with a text-based reward model. The weighted averaging strategy is formally defined as:
\begin{equation*}
    \theta^{\text{MERGE}}_{\text{trans}} = \lambda \cdot \theta^{\text{LVLM}}_{\text{trans}} + (1 - \lambda) \cdot \theta^{\text{RM}}_{\text{trans}},
\end{equation*}
where $\lambda$ is a hyperparameter that determines the weight distribution between the two models. We explore $\lambda$ values in the range: [0.0, 0.1, 0.2, 0.3, 0.4, 0.5, 0.6, 0.7, 0.8, 0.9, 1.0].

\paragraph{Task Arithmetic}
\label{sec:appendix:paragraph:task_arithmetic}
\citet{ilharco2023editing} demonstrated that the task vector, obtained by subtracting the weights of a pre-trained model from those of the same model after fine-tuning for a specific task, defines the task direction. Utilizing this task vector can improve task performance. We also apply the task arithmetic approach to develop a vision-language reward model. The task arithmetic strategy is formally defined as:
\begin{equation*}
    \begin{aligned}
        &\tau^{\text{LVLM}} = \theta^{\text{LVLM}}_{\text{trans}} - \theta^{\text{PRE}}_{\text{trans}}, \\
        &\tau^{\text{RM}} = \theta^{\text{RM}}_{\text{trans}} - \theta^{\text{PRE}}_{\text{trans}}, \\
        &\theta^{\text{MERGE}}_{\text{trans}} = \theta^{\text{PRE}}_{\text{trans}} + \lambda \cdot \tau_{\text{LVLM}} + \lambda \cdot \tau_{\text{RM}},
    \end{aligned}
\end{equation*}
where $\tau^{\text{LVLM}}$ denotes the task vector derived from instruction tuning, and $\tau^{\text{RM}}$ refers to the task vector obtained from reward modeling. The hyperparameter $\lambda$ controls the relative contribution of task vectors. We explore $\lambda$ values in the range: [0.0, 0.1, 0.2, 0.3, 0.4, 0.5, 0.6, 0.7, 0.8, 0.9, 1.0].

\paragraph{TIES}
\label{sec:appendix:paragraph:ties}
\citet{yadav2024ties} consider the interference between parameters from different models during the model merging process. Their approach consists of three main steps. First, they prune task vector values based on magnitude, retaining only a proportion $d$ of the task vector. Second, they resolve sign conflicts by calculating the total magnitude of parameter values in positive and negative directions and selecting the direction with the larger total magnitude. Only values that match the chosen sign are retained. Finally, they compute the mean of the retained values to determine the final parameter value. The TIES method can be simply expressed as:
\begin{equation*} 
    \theta^{\text{MERGE}}_{\text{trans}} = \theta^{\text{PRE}}_{\text{trans}} + \lambda \cdot f(\tau^{\text{LVLM}}, d) + \lambda \cdot f(\tau^{\text{RM}}, d),
\end{equation*}
where $f(\cdot)$ denotes the function for trimming, selecting, and rescaling the task vector, and $d$ is the density determining how many parameters are retained. We search for optimal values of $\lambda$ within the range [0.5, 0.7, 1.0] and $d$ within the range [0.2, 0.4, 0.6, 0.8].

\paragraph{DARE}
\label{sec:appendix:paragraph:dare}
\citet{yu2024language} also addresses the interference between parameters from different models during the model merging process. They randomly drop delta parameters with a probability of $p$ and rescale the remaining ones by $1/(1-p)$. The DARE method can be combined with both the Task Arithmetic and TIES approaches. When combined with Task Arithmetic, a proportion $p$ of task vectors is randomly dropped, and the remaining ones are rescaled by $1/(1-p)$. When DARE is combined with TIES, a proportion $p$ of task vectors is randomly dropped, and the sign of each parameter is determined by comparing the total magnitude in the positive and negative directions. The sign corresponding to the larger total magnitude is selected, and only values matching this sign are retained. Their mean is then computed as the final parameter value, and the result is rescaled by $1/(1-p)$. The DARE method can also be expressed as:
\begin{equation*} 
    \theta^{\text{MERGE}}_{\text{trans}} = \theta^{\text{PRE}}_{\text{trans}} + \lambda \cdot f(\tau^{\text{LVLM}}, d) + \lambda \cdot f(\tau^{\text{RM}}, d),
\end{equation*}
where $d$ represents the density, determining the proportion of retained parameters, with $d=1-p$. We search for optimal values of $\lambda$ within the range [0.5, 0.7, 1.0] and $d$ within the range [0.2, 0.4, 0.6, 0.8].

\paragraph{Merging Embeddings}
\label{sec:appendix:paragraph:merger_embeds}
We follow the embedding merging procedure from MergeKit~\citep{goddard-etal-2024-arcees}. The process is as follows:

\begin{enumerate}
\item If a token exists in the pre-trained model, we use its embedding from that model.
\item If a token appears in only one model (either the LVLM or the text-based RM), we use its embedding from that model.
\item If a token appears in multiple models, we compute the average of its embeddings.
\end{enumerate}

Notably, the pre-trained model is not required for the weighted averaging method. Therefore, we omit the first step when applying this merging approach.

\paragraph{Merging Hyperparameter Selection}
\label{sec:appendix:paragraph:hyper_select}
We select the merging hyperparameter by using a sampled set of 400 instances from the RLAIF-V~\citep{yu2024rlaifv} training set as our validation set. In case of a tie in scores, an additional 100 sampled instances will be used for evaluation. Results are discussed in Appendix~\ref{subsec:appendix:hyperparameters_effect}.

\section{Dataset Details}
\label{sec:appendix:dataset_details}

\paragraph{VL-RewardBench}
\label{sec:appendix:paragraph:vl_rewardbench}
VL-RewardBench~\citep{li2024vlrewardbench} is a benchmark comprising 1,250 high-quality examples spanning three domains: general multimodal instructions, hallucination-related tasks, and multimodal reasoning tasks. Each example includes a multimodal query—consisting of an image and a user prompt—along with a selected response and a rejected response.

\paragraph{TextVQA}
\label{sec:appendix:paragraph:textvqa}
TextVQA~\citep{singh2019towards} is a dataset designed to evaluate the ability of visual question-answering (VQA) models to read and reason about text within images. We use its validation set, which contains 5,000 instances, to assess our merged VLRMs.

\paragraph{MMMU-Pro}
\label{sec:appendix:paragraph:mmmu_pro}
MMMU-Pro~\citep{yue2024mmmu} is an advanced benchmark designed to assess the understanding and reasoning abilities of multimodal models. It is derived from the original MMMU~\citep{yue2024mmmu1} dataset and consists of two subsets: a standard set, which includes image and text queries with 10 answer options, and a vision set, which features a vision-only input scenario. In the vision set, the questions are embedded within screenshots or photos, with no explicit text provided.

\paragraph{RLAIF-V}
\label{sec:appendix:paragraph:rlaif_v}
RLAIF-V~\citep{yu2024rlaifv} preference dataset is created by generating multiple candidate responses for a given prompt and image using various random seeds. Each response is divided into individual claims, which are then assessed using an open-source large vision-language model. This model assigns confidence scores to each claim, which are combined to form an overall response score. Preference pairs are generated by comparing the response scores for the same prompt, selecting the preferred response and the less favorable one based on the score differences. Pairs with significant length disparities are excluded to avoid bias. We select 400 instances from this preference dataset to serve as our validation set for selecting the hyperparameters of merging methods.

\section{Best-of-N Sampling Details}
\label{sec:appendix:best_of_n}
We use lmms-eval~\citep{zhang2024lmms} for response generation with the Best-of-N sampling technique. For the TextVQA dataset, we set both the temperature and top-p to 1.0, sampling 8 responses. To encourage concise answers, we append ``Answer the question using a single word or phrase.'' after the generation prompt. For the MMMU-Pro dataset, we also set the temperature and top p to 1.0, with a maximum token limit of 4096, to sample 8 responses. Additionally, we apply chain-of-thought (CoT) for generating both answers and their reasoning.

\section{Prompt Template}
\label{sec:appendix:prompt_template}
For Best-of-N sampling using \texttt{LLaMA-3.2-Vision} as the generative reward model, the prompt template is provided in Table~\ref{tab:prompt_templates}. For image captioning with \texttt{LLaMA-3.2-Vision} and reward modeling using \texttt{Tulu-3-RM} and \texttt{Tulu-2.5-RM}, the detailed prompt template can also be found in Table~\ref{tab:prompt_templates}.

\section{Open-Source Model Details}
\label{sec:appendix:open_soruce_details}

\paragraph{\texttt{Llama-3.2-11B-Vision-Instruct}}
\texttt{Llama-3.2\\-11B-Vision-Instruct}~\citep{dubey2024llama} is an 11B-parameter LVLM consisting of three main components: a vision encoder, an adapter, and a pre-trained language model. The language model is based on \texttt{Llama-3.1-8B-Instruct}. The adapter incorporates cross-attention layers to integrate image representations into the language model. During adapter training, the language model remains frozen, enabling seamless drop-in replacement for Llama-3.1 series models without requiring re-training.

\paragraph{\texttt{Tulu-2.5-RM}}
\texttt{Tulu-2.5-RM}~\citep{ivison2024unpacking} is a reward model initialized from \texttt{Llama-3.1-8B} and fine-tuned using the Tulu 2 recipe~\citep{ivison2023camels}. It is adapted for reward modeling by replacing the language modeling head with a linear layer and fine-tuning it on preference data from diverse sources, including Ultrafeedback~\citep{cui2024ultrafeedback}, Nectar~\citep{zhu2024starlingb}, HH-RLHF~\citep{bai2022traininghelpfulharmlessassistant}, and AlpacaFarm~\citep{dubois2023alpacafarm}, among others.

\paragraph{\texttt{Tulu-3-RM}}
\texttt{Tulu-3-RM}~\citep{lambert2024t} is another reward model initialized from \texttt{Llama-3.1-8B} and fine-tuned following the Tulu 3 recipe~\citep{lambert2024t}. Like \texttt{Tulu-2.5-RM}, it is adapted for reward modeling by replacing the language modeling head with a linear layer. However, \texttt{Tulu-3-RM} is trained on a mixture of on-policy and off-policy preference data collected through an enhanced version of the Ultrafeedback~\citep{cui2024ultrafeedback} pipeline. This dataset includes prompts from various sources, such as the SFT dataset in the Tulu 3 recipe, WildChat~\citep{zhao2024wildchat}, Ultrafeedback~\citep{cui2024ultrafeedback}, and synthetic persona-augmented instructions.

\section{Qualitative Results}
\label{sec:appendix:qualitative_results}
We investigate reward model behavior before and after merging, and we evaluate qualitatively on VL-RewardBench. Tables~\ref{tab:qualitative_results} and~\ref{tab:qualitative_results_2} present results for \texttt{Tulu-2.5-RM}, while Tables~\ref{tab:qualitative_results_3} and~\ref{tab:qualitative_results_4} show \texttt{Tulu-3-RM}. \textcolor{Red}{Red text} indicates misalignment with the image. Before merging, the text-based reward model made incorrect predictions. After merging, the vision-language reward models correctly identified the better response. In most cases, more advanced merging methods—such as task arithmetic, TIES, and DARE—produce larger reward differences between chosen and rejected responses than simple weighted averaging.

\section{Full Results}
\label{sec:appendix:full_results}
\subsection{Main Results}
\label{subsec:appendix:main_results}
The main results of merging with \texttt{Tulu-2.5-RM} are discussed in Section~\ref{subsec:results} of the main text. As shown in Table~\ref{tab:main_tulu25}, merged VLRMs consistently outperform \texttt{Llama-3.2-Vision} and \texttt{Tulu-2.5-RM} across nearly all merging methods and benchmarks. Notably, in VL-RewardBench, they show the greatest improvement in the Hallucination domain. In Best-of-N evaluation, they perform well in both TextVQA and MMMU-Pro. Additionally, merged VLRMs match or surpass the strong \texttt{Cascade} baseline, suggesting that merging captures more information than simply cascading two models.

A similar trend is observed when merging with \texttt{Tulu-3-RM}. As shown in Table~\ref{tab:main_tulu3}, merged VLRMs outperform \texttt{Llama-3.2-Vision} and \texttt{Tulu-3-RM} across most methods and benchmarks. In VL-RewardBench, they improve mainly in the General and Hallucination domains. For Best-of-N evaluation, they perform well in MMMU-Pro, but only a few achieve results comparable to \texttt{Llama-3.2-Vision} in TextVQA, likely due to \texttt{Tulu-3-RM}’s weaker performance in this task. While merging with \texttt{Llama-3.2-Vision} enhances performance over \texttt{Tulu-3-RM}, it does not surpass \texttt{Llama-3.2-Vision}'s score. Additionally, merged VLRMs exceed the strong \texttt{Cascade} baseline in other benchmarks and remain competitive with it in TextVQA.

In Table~\ref{tab:vlrb_full}, we compare our merged VLRMs with large open-source LVLMs and commercial systems on VL-RewardBench. Surprisingly, our merged VLRMs outperform 90B LVLMs and achieve performance comparable to commercial models, demonstrating the effectiveness of transferring textual preferences from text-based RMs to LVLMs.

\subsection{Without Image Input}
\label{subsec:appendix:without_image_input}
We conduct an ablation study by evaluating models without image input. Full results with \texttt{Tulu-2.5-RM} are shown in Table~\ref{tab:cmp_image_full_tulu25}. Models with image input consistently outperform those without it across various merging techniques, suggesting that the vision encoder actively contributes after merging rather than performance gains being solely due to the text-based RM. This indicates that merged VLRMs effectively utilize the vision encoder in most cases. Notably, in VL-RewardBench, merged VLRMs match or surpass those without image input, especially in the hallucination domain, where image input significantly improves performance. In Best-of-N evaluation, models with image input perform better in the TextVQA and MMMU-Pro Vision sets. However, in the MMMU-Pro Standard set, image input does not provide an advantage, likely because this set emphasizes text reasoning, where reward assessments depend more on textual coherence than visual information.

Full results with \texttt{Tulu-3-RM} are shown in Table~\ref{tab:cmp_image_full_tulu3}, following a similar trend. In VL-RewardBench, merged VLRMs outperform those without image input in the hallucination domain and are comparable to or surpass them in general and reasoning domains. Image input also enhances Best-of-N evaluation, particularly in TextVQA and MMMU-Pro Vision. However, in the MMMU-Pro Standard, image input does not provide a clear advantage, reaffirming that this set prioritizes text reasoning over visual input.

\subsection{Effect of Merging Hyperparameters}
\label{subsec:appendix:hyperparameters_effect}
In this study, we optimize hyperparameter merging using sampled instances from RLAIF-V. The results, based on 400 sampled RLAIF-V instances used as a validation set, are presented in Tables~\ref{tab:hyper_select_linear_tulu25} to \ref{tab:hyper_select_dare_ties_tulu3}. Bold text highlights the best performance, while \textbf{text with *} indicates cases where scores are tied. In these cases, an additional 100 samples are used, and * marks the top-performing result among them.

Figures~\ref{fig:full_tulu2.5_linear} to \ref{fig:full_tulu3_dare_ties} show the effect of hyperparameters across various benchmarks, merging methods, and text-based RMs. The results reveal that optimal hyperparameters differ across these factors, emphasizing the importance of a well-constructed validation set. Future research could further explore this. For example, Figure~\ref{fig:full_tulu2.5_linear} shows the results of searching for $\lambda$ values between 0 and 1 for the \texttt{Linear} method using \texttt{Tulu-2.5-RM}. In the VL-RewardBench, a mid-range $\lambda$ produces the best performance, while in the MMMU-Pro vision set, a smaller $\lambda$ yields better results. This variation suggests that hyperparameter choices influence the performance of the final merged VLRMs differently depending on the task.

Moreover, we observe a trend consistent with prior studies~\citep{yadav2024ties, yu2024language}: even when task vectors are reduced to lower rates (e.g., 0.4, 0.2), merged VLRMs continue to perform well, aligning with findings on LLM merging.

\input{tables/prompt_templates}

\input{tables/qualitatives/qualitative_1}
\input{tables/qualitatives/qualitative_2}
\input{tables/qualitatives/qualitative_3}
\input{tables/qualitatives/qualitative_4}

\input{tables/main_tulu3}
\input{tables/cmp_image_full_tulu25}
\input{tables/cmp_image_full_tulu3}
\input{tables/VLRB_full}
\input{figures/full_results}
\input{tables/hyper_select}

\end{document}

%% file: tables/main_tulu25.tex
\begin{table*}[hbtp]
    \centering
    \resizebox{0.85\textwidth}{!}{
        \begin{tabular}{l c c c c c c c c}
            \toprule
            & \multicolumn{5}{c}{\textbf{VL-RewardBench}} & \multicolumn{1}{c}{\textbf{TextVQA}} & \multicolumn{2}{c}{\textbf{MMMU-Pro}} \\
            \cmidrule(lr){2-6} \cmidrule(lr){7-7} \cmidrule(lr){8-9}
            \textbf{Method} & General & Hallucination & Reasoning & Overall & Macro Avg. & Overall & Standard & Vision \\
            \midrule
            \texttt{Llama-3.2-Vision}          & 33.3* & 38.4* & 56.6* & 42.9* & 42.8* & 46.4 & 28.8 & 19.8 \\
            \texttt{Tulu-2.5-RM}               & 43.2 & 31.4 & 54.1 & 38.9 & 42.9 & 42.6 & 29.8 & 21.4 \\
            \midrule
            \texttt{Random}                    & \textbf{50.0} & 50.0 & 50.0 & 50.0 & 50.0 & 48.2 & 29.2 & 18.4 \\
            \texttt{Cascade}                   & 44.8 & 37.8 & 57.2 & 43.8 & 46.6 & 43.2 & 30.9 & \textbf{23.4} \\
            \midrule
            \texttt{Linear}                    & 39.3 & 52.3 & 54.4 & 51.0 & 48.7 & 54.7 & 27.8 & 22.1 \\
            \texttt{Task Vec.}                 & 48.6 & 59.4 & 59.7 & 57.9 & 55.9 & 59.0 & 31.0 & 22.7 \\
            \texttt{TIES}                      & 43.7 & 58.2 & 58.5 & 56.2 & 53.5 & \textbf{64.2} & 29.1 & 22.6 \\
            \texttt{DARE} + \texttt{Task Vec.} & 49.2 & \textbf{61.7} & \textbf{61.0} & \textbf{59.7} & \textbf{57.3} & 58.8 & 30.3 & 22.4 \\
            \texttt{DARE} + \texttt{TIES}      & 49.2 & 59.1 & 58.2 & 57.4 & 55.5 & 57.3 & \textbf{31.6} & 22.0 \\
            \bottomrule
        \end{tabular}
    }
    \vspace{-3pt}
    \caption{Comparison of merging methods across the VL-RewardBench, TextVQA, and MMMU-Pro datasets using \texttt{TULU-2.5-RM} for merging. *Indicates results from~\citet{li2024vlrewardbench}.}
    \vspace{-5pt}
    \label{tab:main_tulu25}
\end{table*}

%% file: tables/VLRB_sub.tex
\begin{table}[hbtp]
    \centering
    \resizebox{0.48\textwidth}{!}{
        \begin{tabular}{l | c c c}
            \toprule
            \textbf{Method} & General & Hallucination & Reasoning \\
            \midrule
            \multicolumn{4}{c}{\textit{Open-Source Models*}} \\
            \midrule
            \texttt{Llama-3.2-Vision (11B)}  & 33.3 & 38.4 & 56.6 \\
            \texttt{Llama-3.2-Vision (90B)}  & 42.6 & 57.3 & 61.7 \\
            \midrule
            \multicolumn{4}{c}{\textit{Proprietary Models*}} \\
            \midrule
            \texttt{Gemini-1.5-Flash}  & 47.8 & 59.6 & 58.4 \\
            \texttt{Gemini-1.5-Pro}    & 50.8 & 72.5 & 64.2 \\
            \texttt{GPT-4o-mini}       & 41.7 & 34.5 & 58.2 \\
            \texttt{GPT-4o}            & 49.1 & 67.6 & 70.5 \\
            \midrule
            \multicolumn{4}{c}{\textit{Using TULU-2.5-RM for merging}} \\
            \midrule
            \texttt{Linear}                    & 39.3 & 52.3 & 54.4 \\
            \texttt{Task Vec.}                 & 48.6 & 59.4 & 59.7 \\
            \texttt{TIES}                      & 43.7 & 58.2 & 58.5 \\
            \texttt{DARE + Task Vec.}          & 49.2 & 61.7 & 61.0 \\
            \texttt{DARE + TIES}               & 49.2 & 59.1 & 58.2 \\
            \bottomrule
        \end{tabular}
    }
    \caption{VL-RewardBench results comparing open-source and proprietary models with our reward model using TULU-2.5-RM for merging. *Indicates results from~\citet{li2024vlrewardbench}. Full results are provided in Table~\ref{tab:vlrb_full}}
    \label{tab:vlrb_sub}
\end{table}

%% file: tables/cmp_img_small_tulu25.tex
\begin{table}[t!]
    \centering
    \resizebox{\columnwidth}{!}{
        \begin{tabular}{l c c c c}
            \toprule
            & \multicolumn{1}{c}{\textbf{VL-RB}} & \multicolumn{1}{c}{\textbf{TextVQA}} & \multicolumn{2}{c}{\textbf{MMMU-Pro}} \\
            \cmidrule(lr){2-2} \cmidrule(lr){3-3} \cmidrule(lr){4-5}
            \textbf{Method} & Overall & Overall & Standard & Vision \\
            \midrule
            \texttt{Linear}                    & 51.0 & 54.7 & 27.8 & 22.1 \\
            \texttt{w/o image input}           & 39.8 & 45.8 & 29.1 & 21.6 \\
            \midrule
            \texttt{Task Vec.}                 & 57.9 & 59.0 & 31.0 & 22.7 \\
            \texttt{w/o image input}           & 44.9 & 38.7 & 31.8 & 21.0 \\
            \midrule
            \texttt{TIES}                      & 56.2 & 64.2 & 29.1 & 22.6 \\
            \texttt{w/o image input}           & 42.7 & 40.9 & 31.2 & 21.0 \\
            \midrule
            \texttt{DARE} + \texttt{Task Vec.} & 59.7 & 58.8 & 30.3 & 22.4 \\
            \texttt{w/o image input}           & 44.5 & 36.2 & 32.1 & 20.8 \\
            \midrule
            \texttt{DARE} + \texttt{TIES}      & 57.4 & 57.3 & 31.6 & 22.0 \\
            \texttt{w/o image input}           & 45.6 & 36.9 & 32.1 & 20.8 \\
            \bottomrule
        \end{tabular}
    }
    \caption{Comparison of merging methods with and without image input, using \texttt{Tulu-2.5-RM} for merging. VL-RB stands for VL-RewardBench.}
    \vspace{-5pt}
    \label{tab:cmp_image_small_tulu25}
\end{table}

%% file: tables/prompt_templates.tex
\begin{table*}[hbtp]
    \begin{center}
        \begin{tabular}{|p{\textwidth}|}
            \hline
            \textbf{Best-of-N Sampling Prompt template for \texttt{Llama-3.2-Vision}} \\
            \hline
            \textless$\vert$start\_header\_id$\vert$\textgreater system\textless$\vert$end\_header\_id$\vert$\textgreater \\
            You are a highly capable multimodal AI assistant tasked with evaluating answers to visual questions. \\
            \textless$\vert$eot\_id$\vert$\textgreater \textless$\vert$start\_header\_id$\vert$\textgreater user\textless$\vert$end\_header\_id$\vert$\textgreater \\ \\
            Please analyze the following image and question, then evaluate the provided answer: \\ \\
            Question: \\ \\
            \{INSTRUCTION\} \\ \\
            Answer: \\ \\
            \{RESPONSE\} \\ \\
            Evaluate the answer based on the following criteria: \\
            1. Accuracy: How well does the answer align with the visual information in the image? \\
            Score: [1 (Poor) to 5 (Excellent)] \\ \\
            2. Completeness: Does the answer fully address all aspects of the question? \\
            Score: [1 (Poor) to 5 (Excellent)] \\ \\
            3. Clarity: Is the answer well-articulated and easy to understand? \\
            Score: [1 (Poor) to 5 (Excellent)] \\ \\
            4. Relevance: Does the answer directly relate to the question and the image? \\
            Score: [1 (Poor) to 5 (Excellent)] \\ \\
            After your evaluation, please include: \\
            1. Reasoning: A detailed explanation for each criterion, highlighting why you assigned the given score. \\
            2. Overall Assessment: Provide a n overall quality score (1 to 5) for the answer. \\
            \textless$\vert$eot\_id$\vert$\textgreater \\
            \hline
        \end{tabular}
        \begin{tabular}{|p{\textwidth}|}
            \hline
            \textbf{Image Captioning Prompt template using \texttt{Llama-3.2-Vision}} \\
            \hline
            \textless$\vert$start\_header\_id$\vert$\textgreater system\textless$\vert$end\_header\_id$\vert$\textgreater \\
            You are a highly capable multimodal AI assistant tasked with evaluating answers to visual questions. \\
            \textless$\vert$eot\_id$\vert$\textgreater \textless$\vert$start\_header\_id$\vert$\textgreater user\textless$\vert$end\_header\_id$\vert$\textgreater \\ \\
            \{IMAGE\} Please describe this image according to the given question: \{INSTRUCTION\} \\
            \textless$\vert$eot\_id$\vert$\textgreater \\
            \hline
        \end{tabular}
        \begin{tabular}{|p{\textwidth}|}
            \hline
            \textbf{Prompt template for reward modeling} \\
            \hline
            \textless$\vert$start\_header\_id$\vert$\textgreater user\textless$\vert$end\_header\_id$\vert$\textgreater \\ \\
            \{INSTRUCTION\} \\
            \textless$\vert$eot\_id$\vert$\textgreater \textless$\vert$start\_header\_id$\vert$\textgreater assistant\textless$\vert$end\_header\_id$\vert$\textgreater \\ \\
            \{RESPONSE\} \\
            \textless$\vert$eot\_id$\vert$\textgreater \\
            \hline
        \end{tabular}
    \end{center}
    \caption{Prompt template for best-of-n sampling, image captioning and reward modeling.} 
    \label{tab:prompt_templates}
\end{table*}

%% file: tables/qualitatives/qualitative_1.tex
\begin{table*}[hbtp]
    \begin{center}
        \begin{tabular}{|p{\textwidth}|}
        \hline
        \textbf{Image:}
        \begin{center}
            \includegraphics[width=0.5\textwidth]{figures/qualitatives/VLRB1.jpg}
        \end{center} \\
        \hline
        \textbf{Query:} Describe this image in detail. \\
        \hline
        \textbf{Response 1:} \\
        The image showcases a lively scene of a couple of kids playing soccer on a field. They are kicking a sports ball around while several people, including young men and other onlookers, are watching and enjoying the game. \\ \\
        The spectators are spread throughout the scene, with some standing in different spots around the grassy area. There are also numerous chairs set up in the vicinity, possibly for the audience to take a rest or gather at a later time. Furthermore, various personal items, such as backpacks and a handbag, can be observed near the chairs.\\ \\ 
        Additionally, a few cars and an umbrella are present in the background, hinting at an outdoor, casual setting for this friendly soccer match. \\
        \hline
        \textbf{Response 2:} \\
        The image showcases a lively scene of a couple of kids playing soccer on a field. They are kicking a sports ball around while several people, including young men and other onlookers, are watching and enjoying the game. The spectators are spread throughout the scene, with some standing in different spots around the grassy area. There are also numerous chairs set up in the vicinity, possibly for the audience to take a rest or gather at a later time. \textcolor{Red}{Moreover, the soccer goalposts stand tall, ready for the players to score.} Additionally, the grassy terrain adds to the natural charm of the setting, making it an ideal place for a friendly soccer match. \\
        \hline
        \textbf{Score comparison:}
        \begin{center}
            \begin{tabular}{|l|c|c|}
                \hline
                \textbf{Method} & \textbf{Response 1 Score} & \textbf{Response 2 Score} \\
                \hline
                \texttt{Tulu-2.5-RM} & 2.17188 & \textbf{2.27930} \\
                \texttt{Linear} & \textbf{1.89551} & 1.60156 \\
                \texttt{Task Vec.} & \textbf{3.50391} & 1.82422 \\
                \texttt{TIES} & \textbf{2.67773} & 1.96387 \\
                \texttt{DARE + Task Vec.} & \textbf{3.51758} & 1.69141 \\
                \texttt{DARE + TIES} & \textbf{3.44336} & 1.65430 \\
                \hline
            \end{tabular} \\
        \end{center} \\
        \hline
        \end{tabular}
    \end{center}
    \caption{Qualitative results on VL-RewardBench using \texttt{TULU-2.5-RM} for merging.} 
    \label{tab:qualitative_results}
\end{table*}

%% file: tables/qualitatives/qualitative_2.tex
\begin{table*}[hbtp]
    \begin{center}
        \begin{tabular}{|p{\textwidth}|}
        \hline
        \textbf{Image:}
        \begin{center}
            \includegraphics[width=0.5\textwidth]{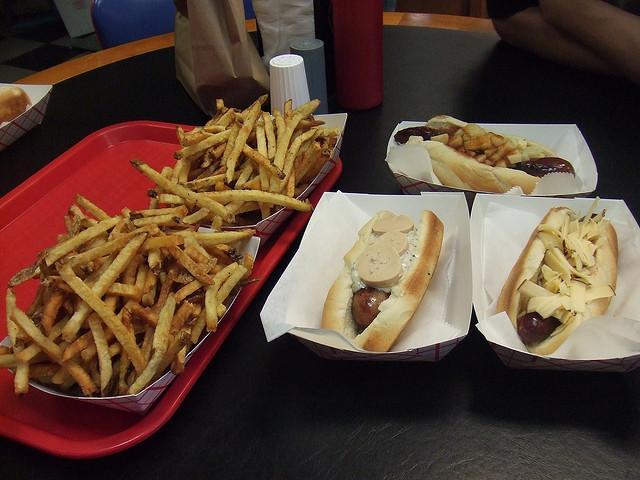}
        \end{center} \\
        \hline
        \textbf{Query:} What is still on the tray? \\
        \hline
        \textbf{Response 1:} \\
        There are still two portions of fries left on the tray. \\
        \hline
        \textbf{Response 2:} \\
        There are still \textcolor{Red}{two hot dogs} and some fries left on the tray. \\
        \hline
        \textbf{Score comparison:}
        \begin{center}
            \begin{tabular}{|l|c|c|}
                \hline
                \textbf{Method} & \textbf{Response 1 Score} & \textbf{Response 2 Score} \\
                \hline
                \texttt{Tulu-2.5-RM} & 0.56885 & \textbf{1.20508} \\
                \texttt{Linear} & \textbf{0.60986} & 0.38477 \\
                \texttt{Task Vec.} & \textbf{2.47461} & 0.55273 \\
                \texttt{TIES} & \textbf{1.76953} & 1.10840 \\
                \texttt{DARE + Task Vec.} & \textbf{2.5918} & 0.62988 \\
                \texttt{DARE + TIES} & \textbf{2.46484} & 0.43701 \\
                \hline
            \end{tabular} \\
        \end{center} \\
        \hline
        \end{tabular}
    \end{center}
    \caption{Qualitative results on VL-RewardBench using \texttt{TULU-2.5-RM} for merging.} 
    \label{tab:qualitative_results_2}
\end{table*}

%% file: tables/qualitatives/qualitative_3.tex
\begin{table*}[hbtp]
    \begin{center}
        \begin{tabular}{|p{\textwidth}|}
        \hline
        \textbf{Image:}
        \begin{center}
            \includegraphics[width=0.5\textwidth]{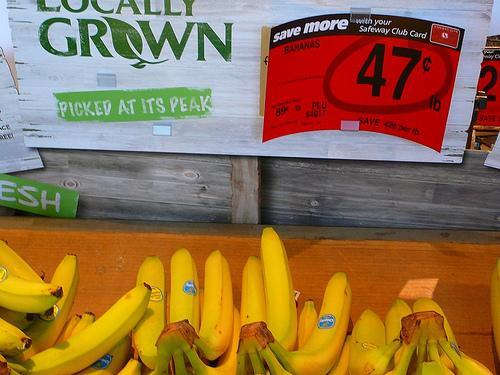}
        \end{center} \\
        \hline
        \textbf{Query:} How many bananas are there in the image? \\
        \hline
        \textbf{Response 1:} \\
        In the picture, there are at least 26 bananas. The bananas are divided into several bunches, which may indicate that they vary in size and ripeness. \\
        \hline
        \textbf{Response 2:} \\
        There are \textcolor{Red}{47} bananas in the picture. The bananas are divided into several bunches, which may indicate that their size and ripeness are different. \\
        \hline
        \textbf{Score comparison:}
        \begin{center}
            \begin{tabular}{|l|c|c|}
                \hline
                \textbf{Method} & \textbf{Response 1 Score} & \textbf{Response 2 Score} \\
                \hline
                \texttt{Tulu-3-RM} & 1.4043 & \textbf{2.33594} \\
                \texttt{Linear} & \textbf{1.33984} & 0.51074 \\
                \texttt{Task Vec.} & \textbf{5.58203} & 4.90234 \\
                \texttt{TIES} & \textbf{7.99219} & 7.55469 \\
                \texttt{DARE + Task Vec.} & \textbf{4.35938} & 2.88672 \\
                \texttt{DARE + TIES} & \textbf{5.51953} & 5.20312 \\
                \hline
            \end{tabular} \\
        \end{center} \\
        \hline
        \end{tabular}
    \end{center}
    \caption{Qualitative results on VL-RewardBench using \texttt{TULU-3-RM} for merging.} 
    \label{tab:qualitative_results_3}
\end{table*}

%% file: tables/qualitatives/qualitative_4.tex
\begin{table*}[hbtp]
    \begin{center}
        \begin{tabular}{|p{\textwidth}|}
        \hline
        \textbf{Image:}
        \begin{center}
            \includegraphics[width=0.5\textwidth]{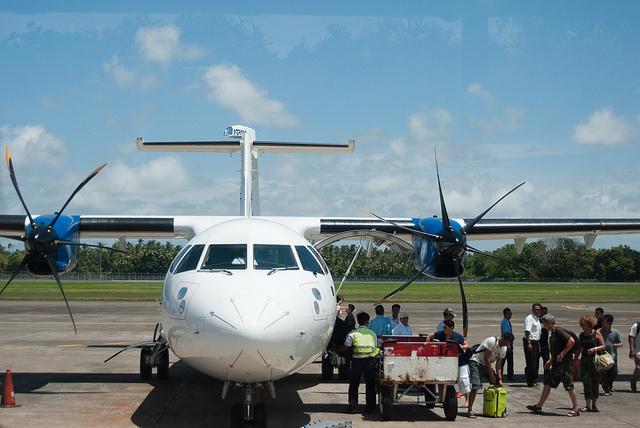}
        \end{center} \\
        \hline
        \textbf{Query:} Describe this image in detail. \\
        \hline
        \textbf{Response 1:} \\
        In the image, there is a twin-propeller airplane parked on the tarmac, surrounded by a crowd of people. Some people seem to be disembarking from the plane, while others are walking around in the vicinity. There are at least 15 people in total, spread out near the airplane.\\ \\
        Luggage is being unloaded, with a suitcase visible near the center of the scene and a handbag located towards the right side. It appears that the passengers are either arriving at their destination or preparing to board the aircraft, as the luggage is being sorted and attended to.\\
        \hline
        \textbf{Response 2:} \\
        \textcolor{Red}{In the image, the flight crew consisting of a pilot and co-pilot can be seen conversing near the twin-propeller airplane parked on the tarmac}. A crowd of people surrounds them, with some individuals disembarking from the plane and others preparing to board. \textcolor{Red}{Ground support equipment such as baggage carts and fuel trucks are visible in the background near the airport terminal building.} Luggage is being unloaded and sorted, with a suitcase and handbag visible in the foreground. The blue stripe on the airplane adds a pop of color to the scene. It appears that the passengers are either arriving at their destination or preparing to depart on their journey. \textcolor{Red}{An information sign can be seen towards the left side of the image.}\\
        \hline
        \textbf{Score comparison:}
        \begin{center}
            \begin{tabular}{|l|c|c|}
                \hline
                \textbf{Method} & \textbf{Response 1 Score} & \textbf{Response 2 Score} \\
                \hline
                \texttt{Tulu-3-RM} & 3.94531 & \textbf{4.74219} \\
                \texttt{Linear} & \textbf{3.66016} & 2.74414 \\
                \texttt{Task Vec.} & \textbf{5.23828} & 2.99219 \\
                \texttt{TIES} & \textbf{7.72656} & 5.67188 \\
                \texttt{DARE + Task Vec.} & \textbf{4.67188} & 2.24414 \\
                \texttt{DARE + TIES} & \textbf{5.79688} & 2.88477 \\
                \hline
            \end{tabular} \\
        \end{center} \\
        \hline
        \end{tabular}
    \end{center}
    \caption{Qualitative results on VL-RewardBench using \texttt{TULU-3-RM} for merging.} 
    \label{tab:qualitative_results_4}
\end{table*}

%% file: tables/main_tulu3.tex
\begin{table*}[hbtp]
    \centering
    \resizebox{\textwidth}{!}{
        \begin{tabular}{l c c c c c c c c}
            \toprule
            & \multicolumn{5}{c}{\textbf{VL-RewardBench}} & \multicolumn{1}{c}{\textbf{TextVQA}} & \multicolumn{2}{c}{\textbf{MMMU-Pro}} \\
            \cmidrule(lr){2-6} \cmidrule(lr){7-7} \cmidrule(lr){8-9}
            \textbf{Method} & General & Hallucination & Reasoning & Overall & Macro Avg. & Overall & Standard & Vision \\
            \midrule
            \texttt{Llama-3.2-Vision}          & 33.3* & 38.4* & 56.6* & 42.9* & 42.8* & 46.4 & 28.8 & 19.8 \\
            \texttt{Tulu-3-RM}                 & 45.4 & 36.6 & 56.6 & 43.0 & 46.2 & 27.4 & 29.4 & 20.4 \\
            \midrule
            \texttt{Random}                    & 50.0 & 50.0 & 50.0 & 50.0 & 50.0 & \textbf{48.2} & 29.2 & 18.4 \\
            \texttt{Cascade}                   & 54.1 & 40.5 & 57.2 & 46.7 & 50.6 & 38.3 & 31.3 & \textbf{23.7} \\
            \midrule
            \texttt{Linear}                    & 47.5 & 51.0 & 55.0 & 51.5 & 51.2 & 45.8 & 29.1 & 19.0 \\
            \texttt{Task Vec.}                 & 63.4 & 66.4 & 57.5 & 63.7 & 62.4 & 36.0 & \textbf{31.6} & 20.9 \\
            \texttt{TIES}                      & 59.0 & \textbf{74.1} & 50.9 & \textbf{66.0} & 61.4 & 28.3 & 30.7 & 20.6 \\
            \texttt{DARE} + \texttt{Task Vec.} & 63.4 & 68.9 & \textbf{58.5} & 65.4 & \textbf{63.6} & 36.1 & 30.2 & 20.9 \\
            \texttt{DARE} + \texttt{TIES}      & \textbf{63.9} & 65.6 & 57.2 & 63.2 & 62.2 & 56.9 & 31.4 & 21.8 \\
            \bottomrule
        \end{tabular}
    }
    \caption{Comparison of merging methods across the VL-RewardBench, TextVQA, and MMMU-Pro datasets using \texttt{TULU-3-RM} for merging. *Indicates results from~\citet{li2024vlrewardbench}.}
    \label{tab:main_tulu3}
\end{table*}

%% file: tables/cmp_image_full_tulu25.tex
\begin{table*}[hbtp]
    \centering
    \resizebox{\textwidth}{!}{
        \begin{tabular}{l c c c c c c c c}
            \toprule
            & \multicolumn{5}{c}{\textbf{VL-RewardBench}} & \multicolumn{1}{c}{\textbf{TextVQA}} & \multicolumn{2}{c}{\textbf{MMMU-Pro}} \\
            \cmidrule(lr){2-6} \cmidrule(lr){7-7} \cmidrule(lr){8-9}
            \textbf{Method} & General & Hallucination & Reasoning & Overall & Macro Avg. & Overall & Standard & Vision \\
            \midrule
            \texttt{Linear}                    & 39.3 \textcolor{Red}{(-2.2)} & 52.3 \textcolor{Green}{(+20.8)} & 54.4 \textcolor{Red}{(-4.1)} & 51.0 \textcolor{Green}{(+11.2)} & 48.7 \textcolor{Green}{(+4.9)} & 54.7 \textcolor{Green}{(+8.9)} & 27.8 \textcolor{Red}{(-1.3)} & 22.1 \textcolor{Green}{(+0.5)} \\
            \texttt{w/o image input}           & 41.5 & 31.5 & 58.5 & 39.8 & 43.8 & 45.8 & 29.1 & 21.6 \\
            \midrule
            \texttt{Task Vec.}                 & 48.6 \textcolor{Green}{(+4.3)} & 59.4 \textcolor{Green}{(+20.4)} & 59.7 \textcolor{Green}{(+0.6)} & 57.9 \textcolor{Green}{(+13.0)} & 55.9 \textcolor{Green}{(+8.4)} & 59.0 \textcolor{Green}{(+20.3)} & 31.0 \textcolor{Red}{(-0.8)} & 22.7 \textcolor{Green}{(+1.7)} \\
            \texttt{w/o image input}           & 44.3 & 39.0 & 59.1 & 44.9 & 47.5 & 38.7 & 31.8 & 21.0 \\
            \midrule
            \texttt{TIES}                      & 43.7 \textcolor{Red}{(-1.1)} & 58.2 \textcolor{Green}{(+23.0)} & 58.5 \textcolor{Red}{(-0.6)} & 56.2 \textcolor{Green}{(+13.5)} & 53.5 \textcolor{Green}{(+7.1)} & 64.2 \textcolor{Green}{(+23.3)} & 29.1 \textcolor{Red}{(-2.1)} & 22.6 \textcolor{Green}{(+1.6)} \\
            \texttt{w/o image input}           & 44.8 & 35.2 & 59.1 & 42.7 & 46.4 & 40.9 & 31.2 & 21.0 \\
            \midrule
            \texttt{DARE} + \texttt{Task Vec.} & 49.2 \textcolor{Green}{(+4.4)} & 61.7 \textcolor{Green}{(+23.4)} & 61.0 \textcolor{Green}{(+2.2)} & 59.7 \textcolor{Green}{(+15.2)} & 57.3 \textcolor{Green}{(+10.0)} & 58.8 \textcolor{Green}{(+22.6)} & 30.3 \textcolor{Red}{(-1.8)} & 22.4 \textcolor{Green}{(+1.6)} \\
            \texttt{w/o image input}           & 44.8 & 38.3 & 58.8 & 44.5 & 47.3 & 36.2 & 32.1 & 20.8 \\
            \midrule
            \texttt{DARE} + \texttt{TIES}      & 49.2 \textcolor{Green}{(+3.3)} & 59.1 \textcolor{Green}{(+19.2)} & 58.2 \textcolor{Red}{(-0.6)} & 57.4 \textcolor{Green}{(+11.8)} & 55.5 \textcolor{Green}{(+7.3)} & 57.3 \textcolor{Green}{(+20.4)} & 31.6 \textcolor{Red}{(-0.5)} & 22.0 \textcolor{Green}{(+1.2)} \\
            \texttt{w/o image input}           & 45.9 & 39.9 & 58.8 & 45.6 & 48.2 & 36.9 & 32.1 & 20.8 \\
            \bottomrule
        \end{tabular}
    }
    \caption{Full results comparing merging methods with and without image input, using \texttt{TULU-2.5-RM} for merging.}
    \label{tab:cmp_image_full_tulu25}
\end{table*}

%% file: tables/cmp_image_full_tulu3.tex
\begin{table*}[hbtp]
    \centering
    \resizebox{\textwidth}{!}{
        \begin{tabular}{l c c c c c c c c}
            \toprule
            & \multicolumn{5}{c}{\textbf{VL-RewardBench}} & \multicolumn{1}{c}{\textbf{TextVQA}} & \multicolumn{2}{c}{\textbf{MMMU-Pro}} \\
            \cmidrule(lr){2-6} \cmidrule(lr){7-7} \cmidrule(lr){8-9}
            \textbf{Method} & General & Hallucination & Reasoning & Overall & Macro Avg. & Overall & Standard & Vision \\
            \midrule
            \texttt{Linear}                    & 47.5 \textcolor{Red}{(-1.1)} & 51.0 \textcolor{Green}{(+1.1)} & 55.0 \textcolor{Green}{(0.0)} & 51.5 \textcolor{Green}{(+0.5)} & 51.2 \textcolor{Green}{(0.0)} & 45.8 \textcolor{Green}{(+25.5)} & 29.1 \textcolor{Green}{(+0.5)} & 19.0 \textcolor{Red}{(-1.3)} \\
            \texttt{w/o image input}           & 48.6 & 49.9 & 55.0 & 51.0 & 51.2 & 20.3 & 28.6 & 20.3 \\
            \midrule
            \texttt{Task Vec.}                 & 63.4 \textcolor{Green}{(+3.8)} & 66.4 \textcolor{Green}{(+19.3)} & 57.5 \textcolor{Green}{(+4.4)} & 63.7 \textcolor{Green}{(+13.2)} & 62.4 \textcolor{Green}{(+9.1)} & 36.0 \textcolor{Green}{(+1.2)} & 31.6 \textcolor{Red}{(-0.1)} & 20.9 \textcolor{Green}{(+0.3)} \\
            \texttt{w/o image input}           & 59.6 & 47.1 & 53.1 & 50.5 & 53.3 & 34.8 & 31.7 & 20.6 \\
            \midrule
            \texttt{TIES}                      & 59.0 \textcolor{Red}{(-0.6)} & 74.1 \textcolor{Green}{(+33.5)} & 50.9 \textcolor{Red}{(-3.2)} & 66.0 \textcolor{Green}{(+19.2)} & 61.4 \textcolor{Green}{(+10.0)} & 28.3 \textcolor{Red}{(-0.3)} & 30.7 \textcolor{Red}{(-1.0)} & 20.6 \textcolor{Red}{(-0.9)} \\
            \texttt{w/o image input}           & 59.6 & 40.6 & 54.1 & 46.8 & 51.4 & 28.6 & 31.7 & 21.5 \\
            \midrule
            \texttt{DARE} + \texttt{Task Vec.} & 63.4 \textcolor{Green}{(+3.8)} & 68.9 \textcolor{Green}{(+18.4)} & 58.5 \textcolor{Green}{(+2.2)} & 65.4 \textcolor{Green}{(+12.1)} & 63.6 \textcolor{Green}{(+8.2)} & 36.1 \textcolor{Red}{(-5.8)} & 30.2 \textcolor{Red}{(-1.9)} & 20.9 \textcolor{Green}{(+0.7)} \\
            \texttt{w/o image input}           & 59.6 & 50.5 & 56.3 & 53.3 & 55.4 & 41.9 & 32.1 & 20.2 \\
            \midrule
            \texttt{DARE} + \texttt{TIES}      & 63.9 \textcolor{Green}{(+8.7)} & 65.6 \textcolor{Green}{(+20.9)} & 57.2 \textcolor{Green}{(+1.9)} & 63.2 \textcolor{Green}{(+14.2)} & 62.2 \textcolor{Green}{(+10.4)} & 56.9 \textcolor{Green}{(+29.2)} & 31.4 \textcolor{Green}{(+0.6)} & 21.8 \textcolor{Green}{(+1.4)} \\
            \texttt{w/o image input}           & 55.2 & 44.7 & 55.3 & 49.0 & 51.8 & 27.7 & 30.8 & 20.4 \\
            \bottomrule
        \end{tabular}
    }
    \caption{Full results comparing merging methods with and without image input, using \texttt{TULU-3-RM} for merging.}
    \label{tab:cmp_image_full_tulu3}
\end{table*}

%% file: tables/VLRB_full.tex
\begin{table*}[hbtp]
    \centering
    \resizebox{0.8\textwidth}{!}{
        \begin{tabular}{l | c c c | c c}
            \toprule
            \textbf{Method} & General & Hallucination & Reasoning & Overall & Macro Avg. \\
            \midrule
            \multicolumn{6}{c}{\textit{Open-Source Models*}} \\
            \midrule
            \texttt{Llama-3.2-Vision-11B-Instruct}  & 33.3 & 38.4 & 56.6 & 42.9 & 42.8 \\
            \texttt{Llama-3.2-Vision-90B-Instruct}  & 42.6 & 57.3 & 61.7 & 56.2 & 53.9 \\ 
            \texttt{Qwen2-VL-72B-Instruct}          & 38.1 & 32.8 & 58.0 & 39.5 & 43.0 \\
            \texttt{Molmo-72B-0924}                 & 33.9 & 42.3 & 54.9 & 44.1 & 43.7 \\
            \texttt{NVLM-D-72B}                     & 38.9 & 31.6 & 62.0 & 40.1 & 44.1 \\ 
            \midrule
            \multicolumn{6}{c}{\textit{Proprietary Models*}} \\
            \midrule
            \texttt{Gemini-1.5-Flash (2024-09-24)}  & 47.8 & 59.6 & 58.4 & 57.6 & 55.3 \\
            \texttt{Gemini-1.5-Pro (2024-09-24)}    & 50.8 & 72.5 & 64.2 & \textbf{67.2} & 62.5 \\
            \texttt{Claude-3.5-Sonnet (2024-06-22)} & 43.4 & 55.0 & 62.3 & 55.3 & 53.6 \\
            \texttt{GPT-4o-mini (2024-07-18)}       & 41.7 & 34.5 & 58.2 & 41.5 & 44.8 \\
            \texttt{GPT-4o (2024-08-06)}            & 49.1 & 67.6 & \textbf{70.5} & 65.8 & 62.4\\ 
            \midrule
            \multicolumn{6}{c}{\textit{Using TULU-2.5-RM for merging}} \\
            \midrule
            \texttt{Linear}                    & 39.3 & 52.3 & 54.4 & 51.0 & 48.7 \\
            \texttt{Task Vec.}                 & 48.6 & 59.4 & 59.7 & 57.9 & 55.9 \\
            \texttt{TIES}                      & 43.7 & 58.2 & 58.5 & 56.2 & 53.5 \\
            \texttt{DARE} + \texttt{Task Vec.} & 49.2 & 61.7 & 61.0 & 59.7 & 57.3 \\
            \texttt{DARE} + \texttt{TIES}      & 49.2 & 59.1 & 58.2 & 57.4 & 55.5 \\
            \midrule
            \multicolumn{6}{c}{\textit{Using TULU-3-RM for merging}} \\
            \midrule
            \texttt{Linear}                    & 47.5 & 51.0 & 55.0 & 51.5 & 51.2 \\
            \texttt{Task Vec.}                 & 63.4 & 66.4 & 57.5 & 63.7 & 62.4 \\
            \texttt{TIES}                      & 59.0 & \textbf{74.1} & 50.9 & 66.0 & 61.4 \\
            \texttt{DARE} + \texttt{Task Vec.} & 63.4 & 68.9 & 58.5 & 65.4 & \textbf{63.6} \\
            \texttt{DARE} + \texttt{TIES}      & \textbf{63.9} & 65.6 & 57.2 & 63.2 & 62.2 \\
            \bottomrule
        \end{tabular}
    }
    \caption{Full results on VL-RewardBench, compared with current strong large vision-language models. *Indicates results from~\citet{li2024vlrewardbench}.}
    \label{tab:vlrb_full}
\end{table*}

%% file: figures/full_results.tex
\begin{figure*}[ht]
     \centering
     \begin{subfigure}[b]{0.245\linewidth}
         \centering
         \includegraphics[width=\textwidth]{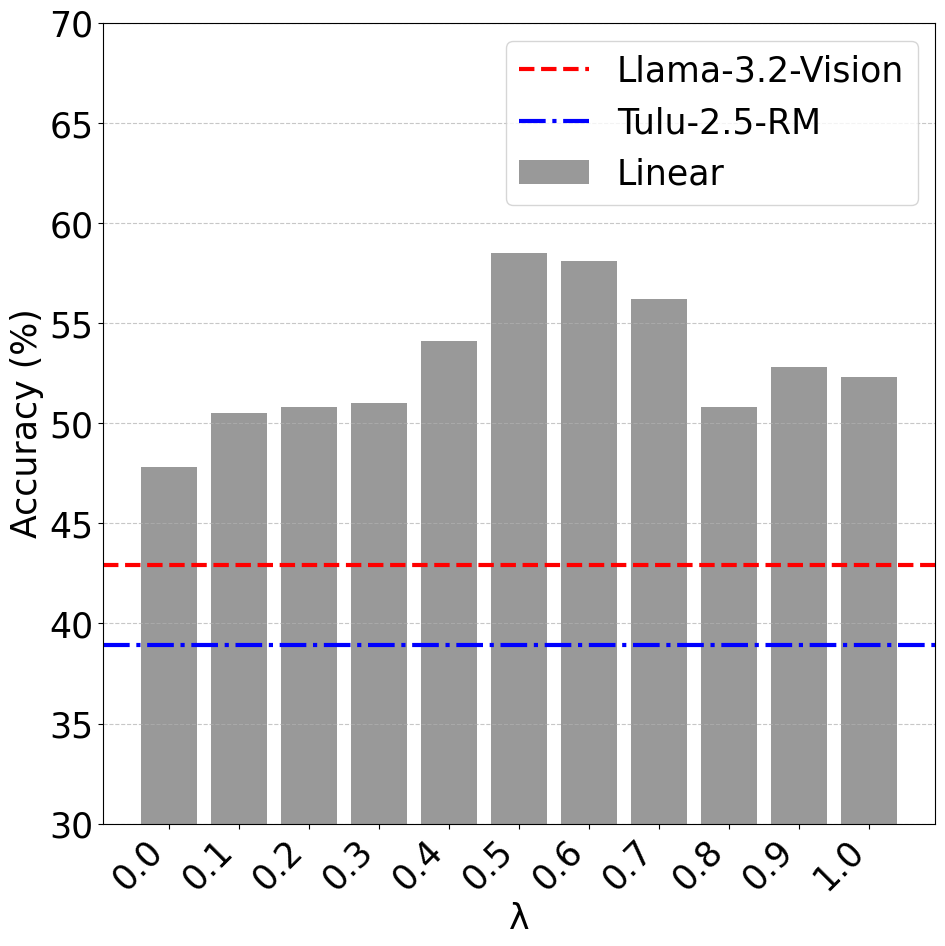}
         \caption{VL-RewardBench}
     \end{subfigure}
     \hfill
     \begin{subfigure}[b]{0.245\linewidth}
         \centering
         \includegraphics[width=\textwidth]{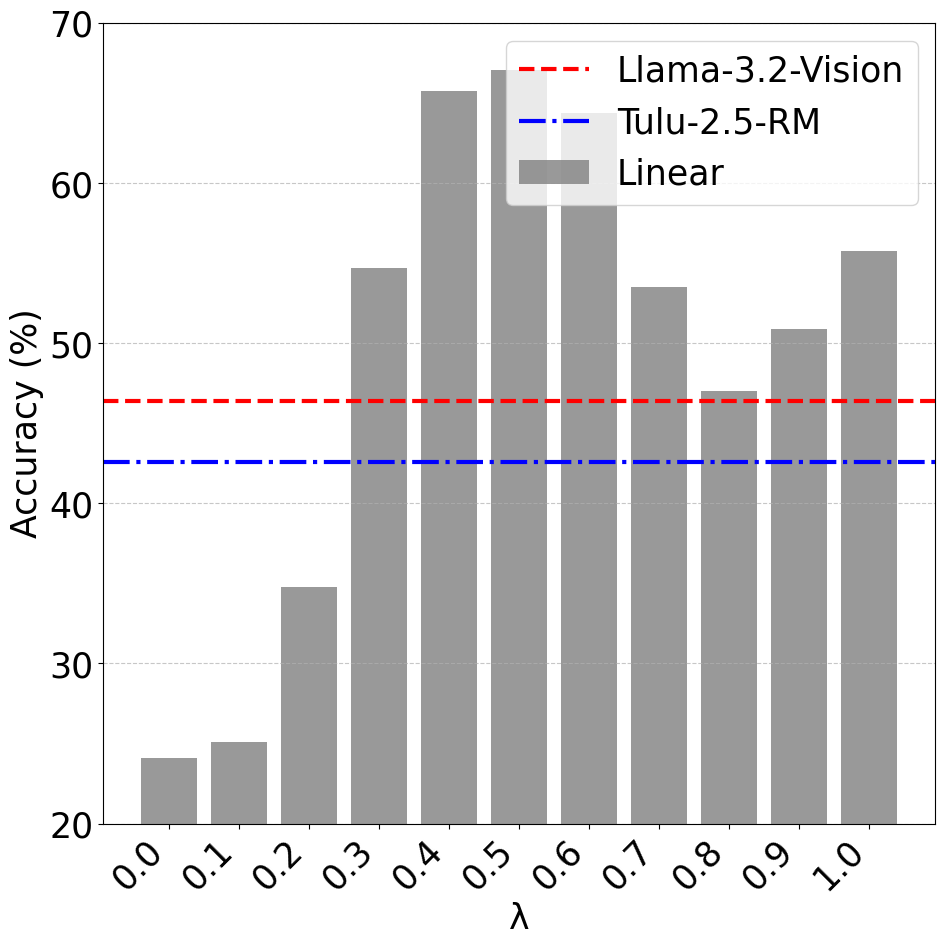}
         \caption{TextVQA}
     \end{subfigure}
     \hfill
      \begin{subfigure}[b]{0.245\linewidth}
         \centering
         \includegraphics[width=\textwidth]{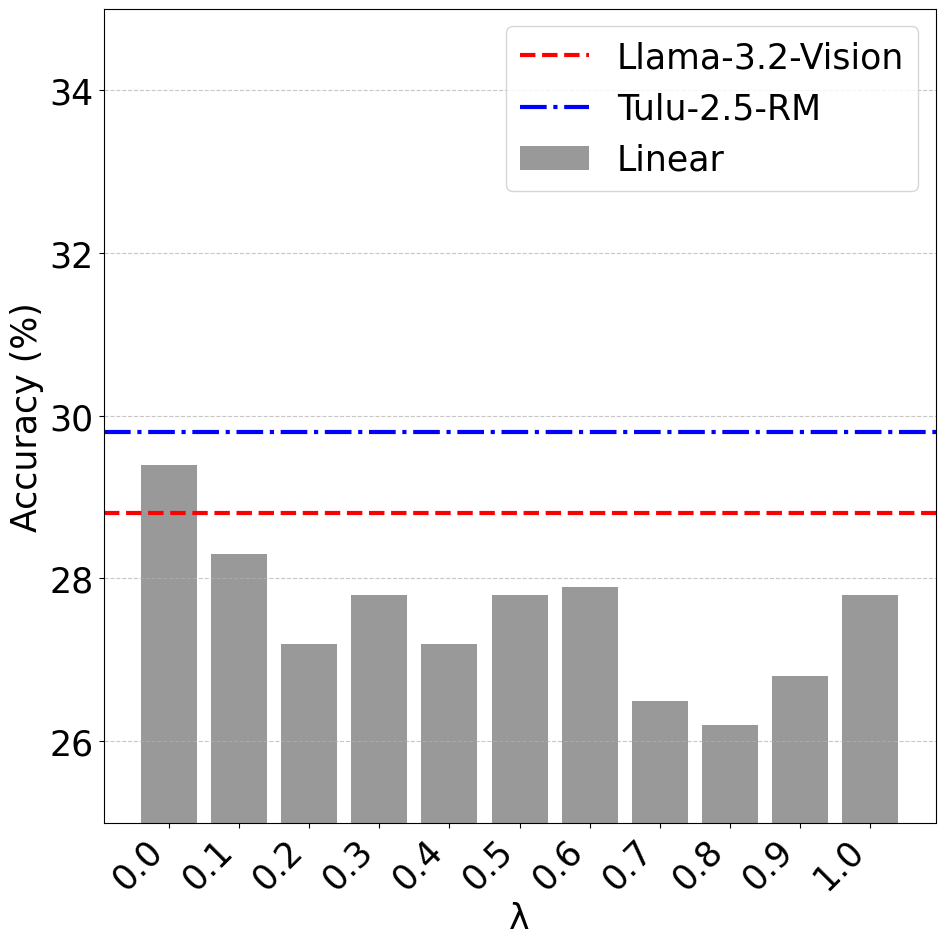}
         \caption{MMMU-Pro (Standard)}
     \end{subfigure}
     \hfill
     \begin{subfigure}[b]{0.245\linewidth}
         \centering
         \includegraphics[width=\textwidth]{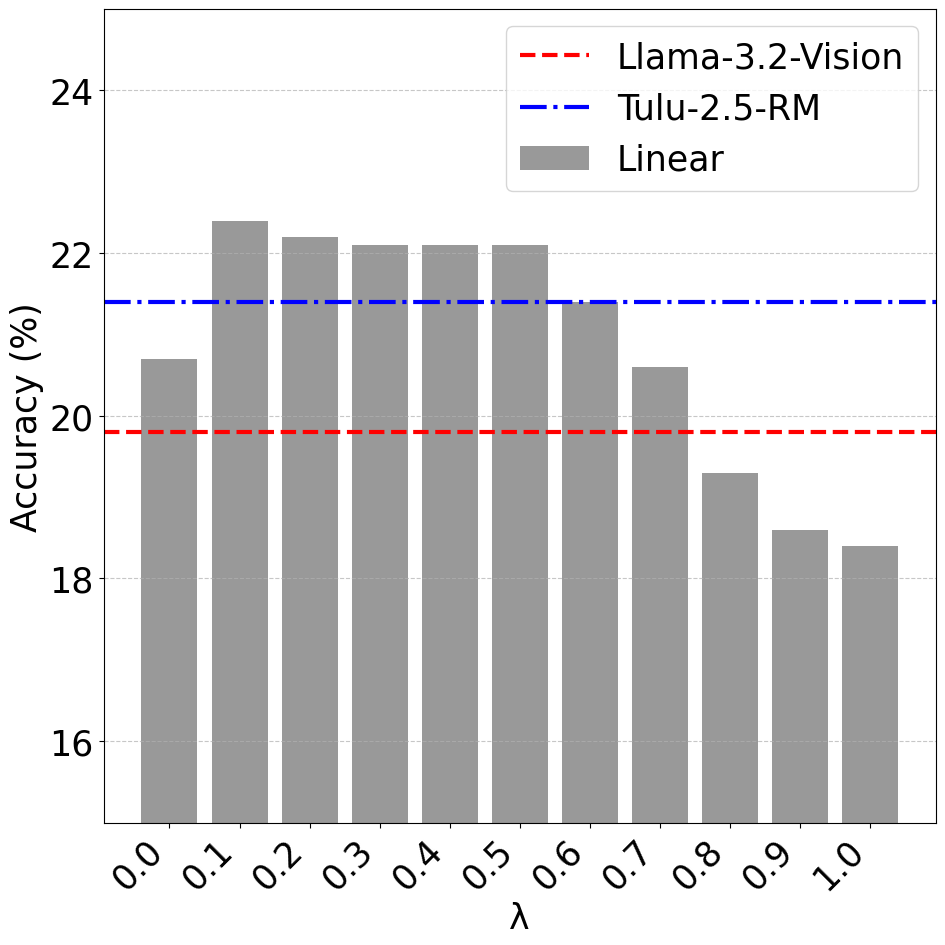}
         \caption{MMMU-Pro (Vision)}
     \end{subfigure}
        \caption{Full results of merging \texttt{Llama-3.2-Vision} and \texttt{Tulu-2.5-RM} (\texttt{Linear})}
        \vspace{-10pt}
        \label{fig:full_tulu2.5_linear}
\end{figure*}

\begin{figure*}[ht]
     \centering
     \begin{subfigure}[b]{0.245\linewidth}
         \centering
         \includegraphics[width=\textwidth]{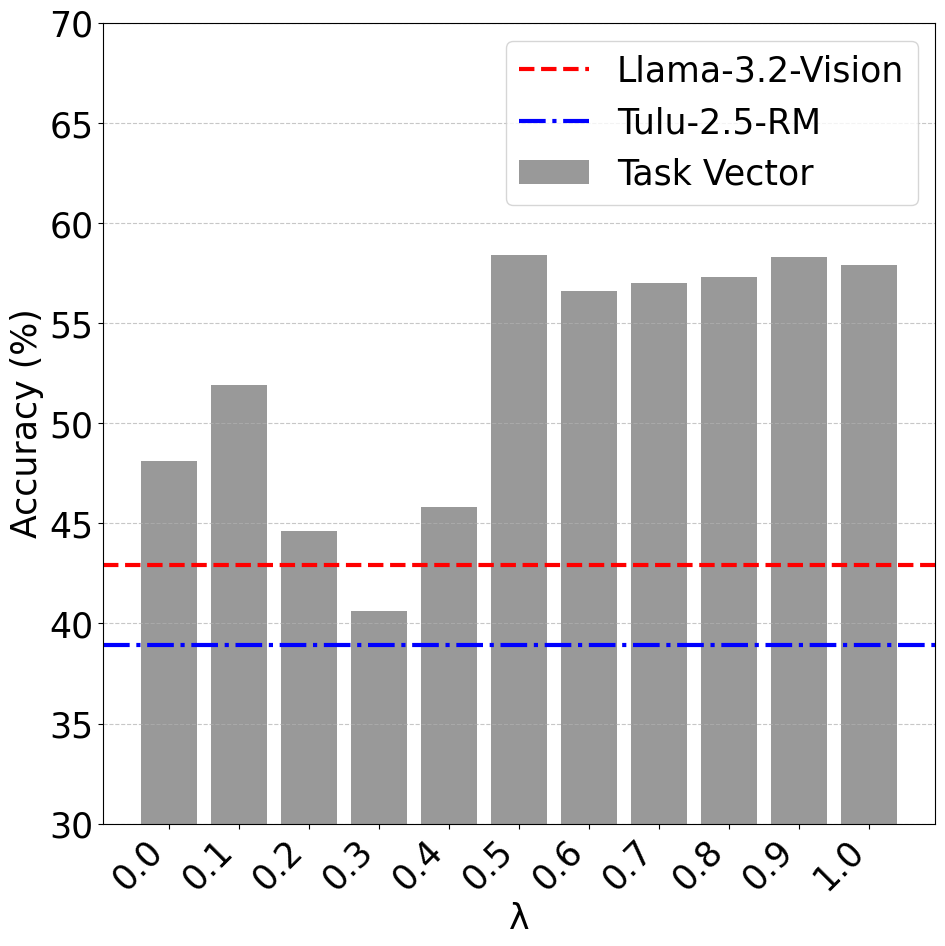}
         \caption{VL-RewardBench}
     \end{subfigure}
     \hfill
     \begin{subfigure}[b]{0.245\linewidth}
         \centering
         \includegraphics[width=\textwidth]{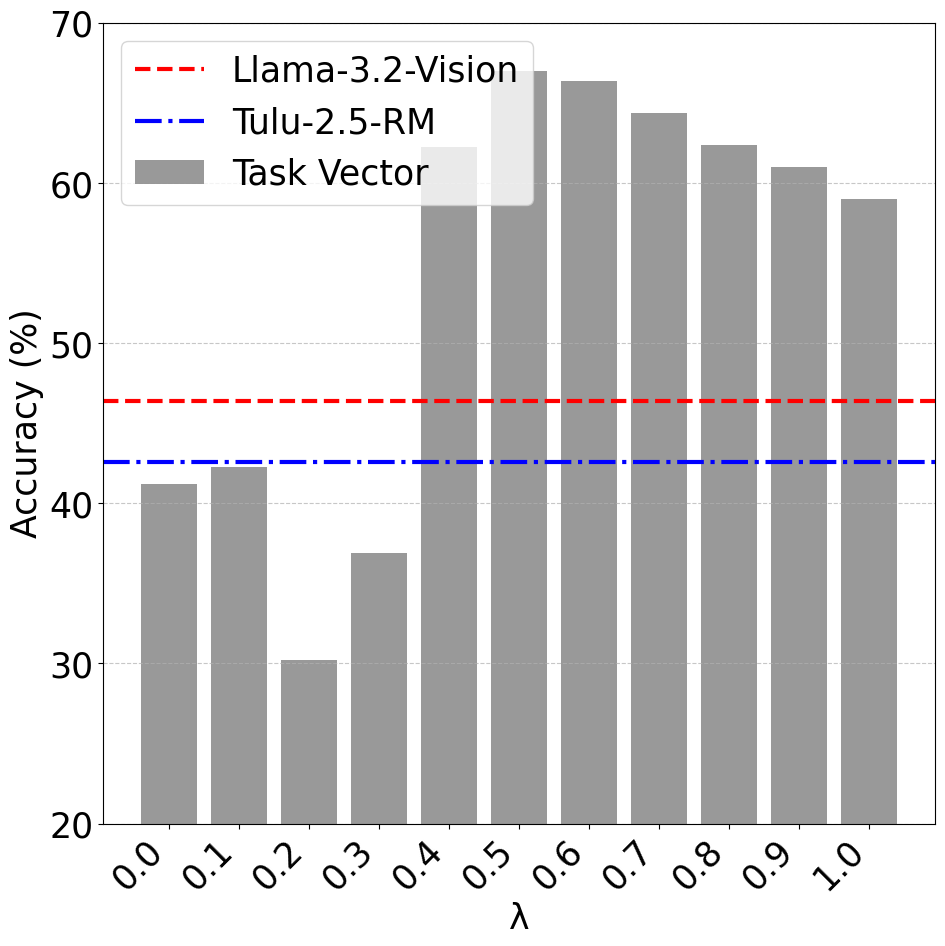}
         \caption{TextVQA}
     \end{subfigure}
     \hfill
      \begin{subfigure}[b]{0.245\linewidth}
         \centering
         \includegraphics[width=\textwidth]{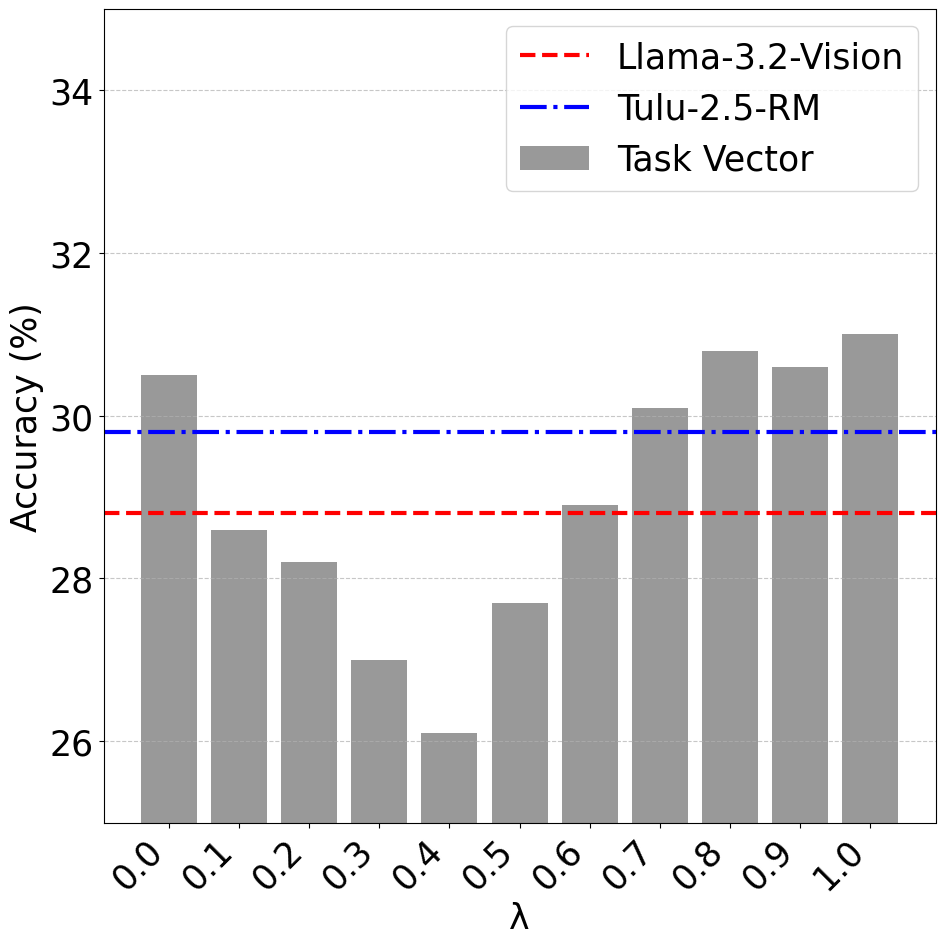}
         \caption{MMMU-Pro (Standard)}
     \end{subfigure}
     \hfill
     \begin{subfigure}[b]{0.245\linewidth}
         \centering
         \includegraphics[width=\textwidth]{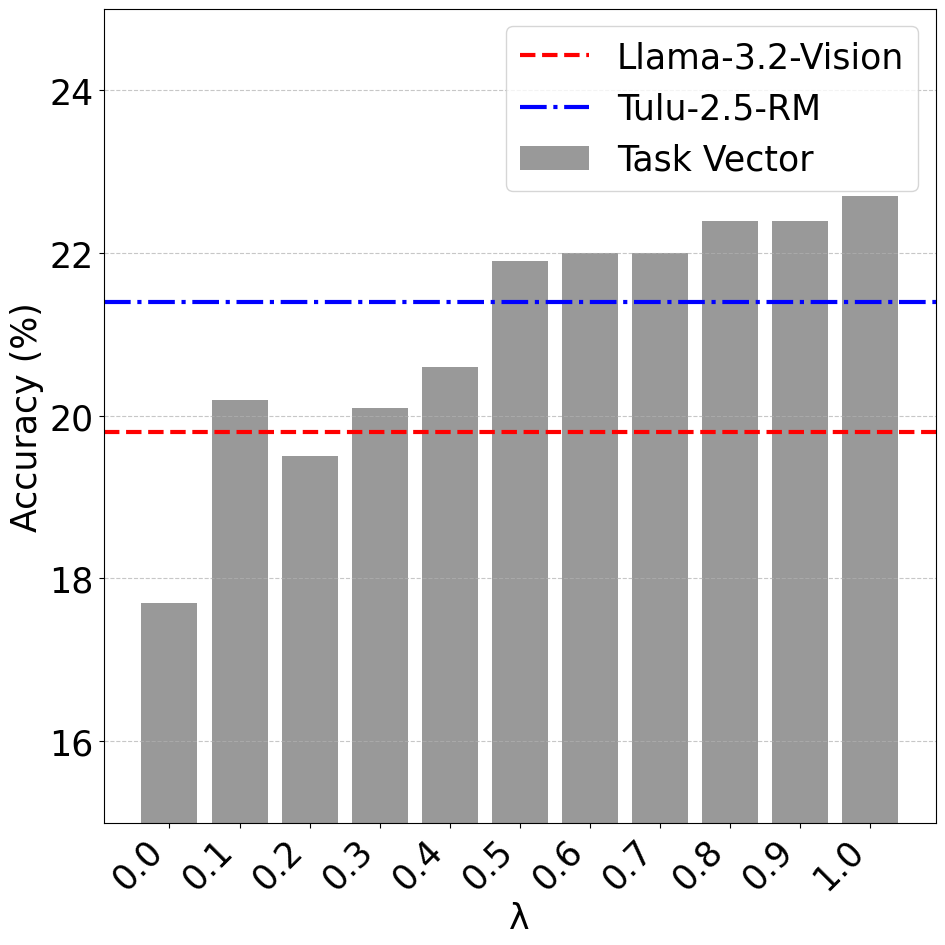}
         \caption{MMMU-Pro (Vision)}
     \end{subfigure}
        \caption{Full results of merging \texttt{Llama-3.2-Vision} and \texttt{Tulu-2.5-RM} (\texttt{Task Vec.})}
        \vspace{-10pt}
        \label{fig:full_tulu2.5_task_vector}
\end{figure*}

\begin{figure*}[ht]
     \centering
     \begin{subfigure}[b]{0.245\linewidth}
         \centering
         \includegraphics[width=\textwidth]{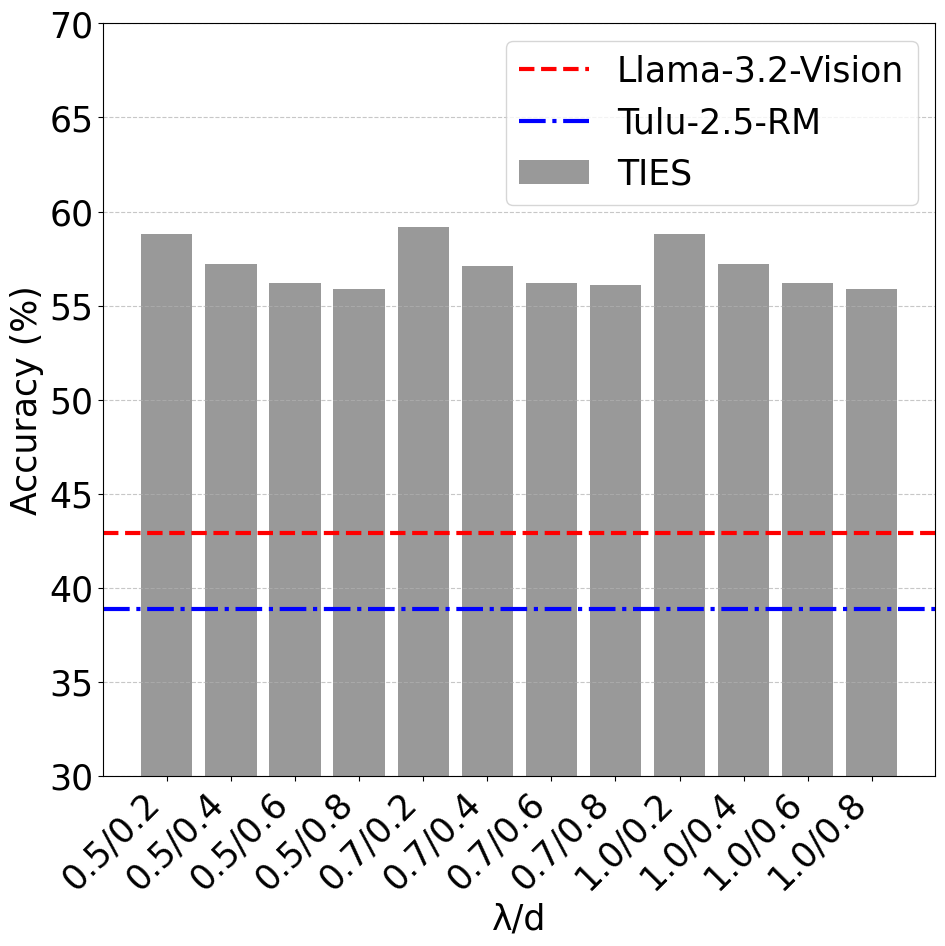}
         \caption{VL-RewardBench}
     \end{subfigure}
     \hfill
     \begin{subfigure}[b]{0.245\linewidth}
         \centering
         \includegraphics[width=\textwidth]{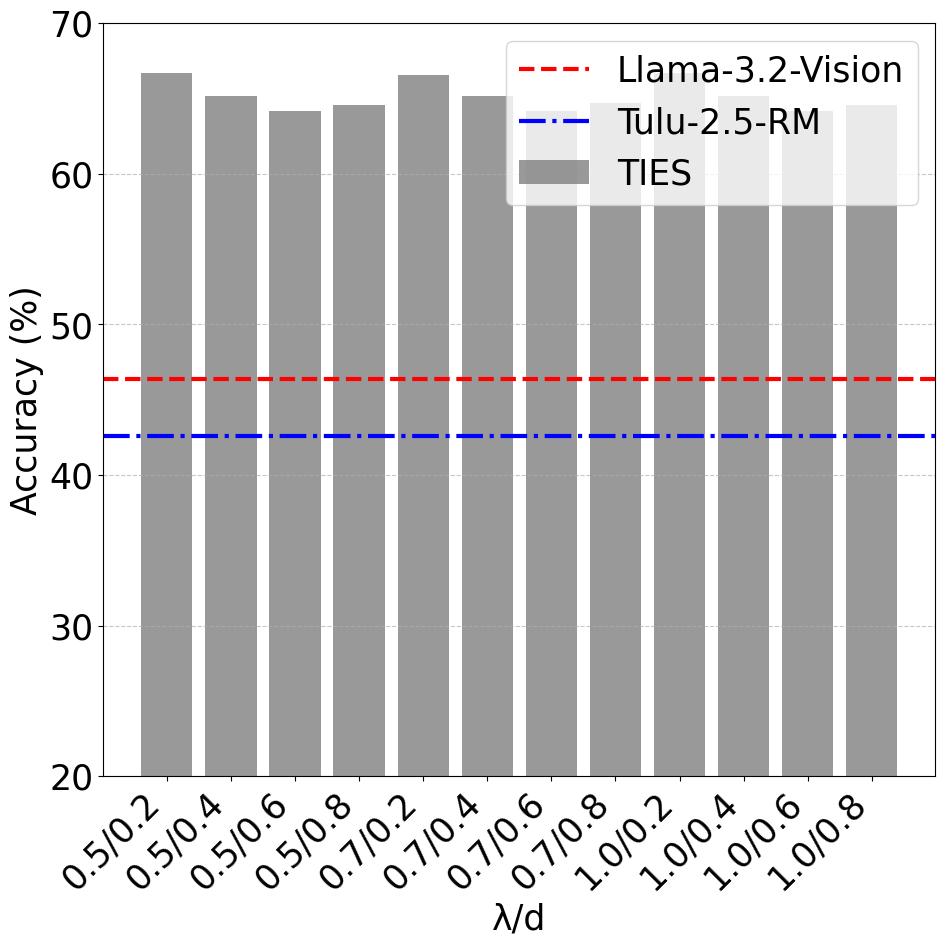}
         \caption{TextVQA}
     \end{subfigure}
     \hfill
      \begin{subfigure}[b]{0.245\linewidth}
         \centering
         \includegraphics[width=\textwidth]{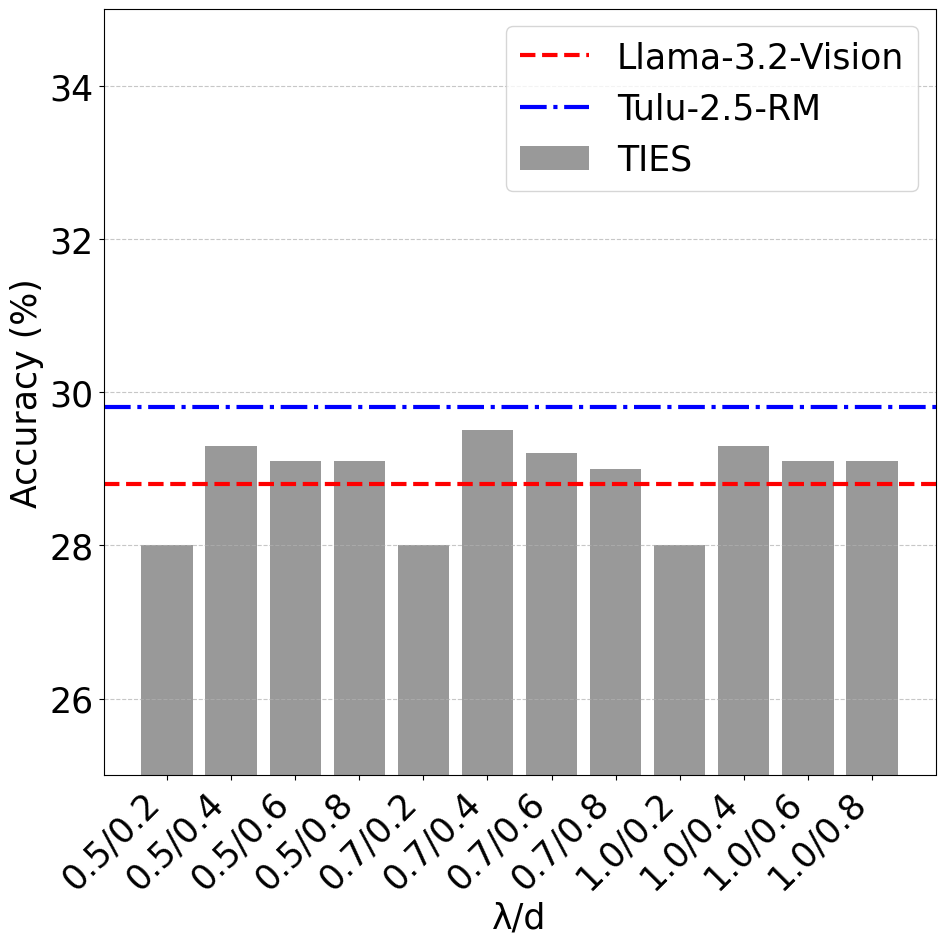}
         \caption{MMMU-Pro (Standard)}
     \end{subfigure}
     \hfill
     \begin{subfigure}[b]{0.245\linewidth}
         \centering
         \includegraphics[width=\textwidth]{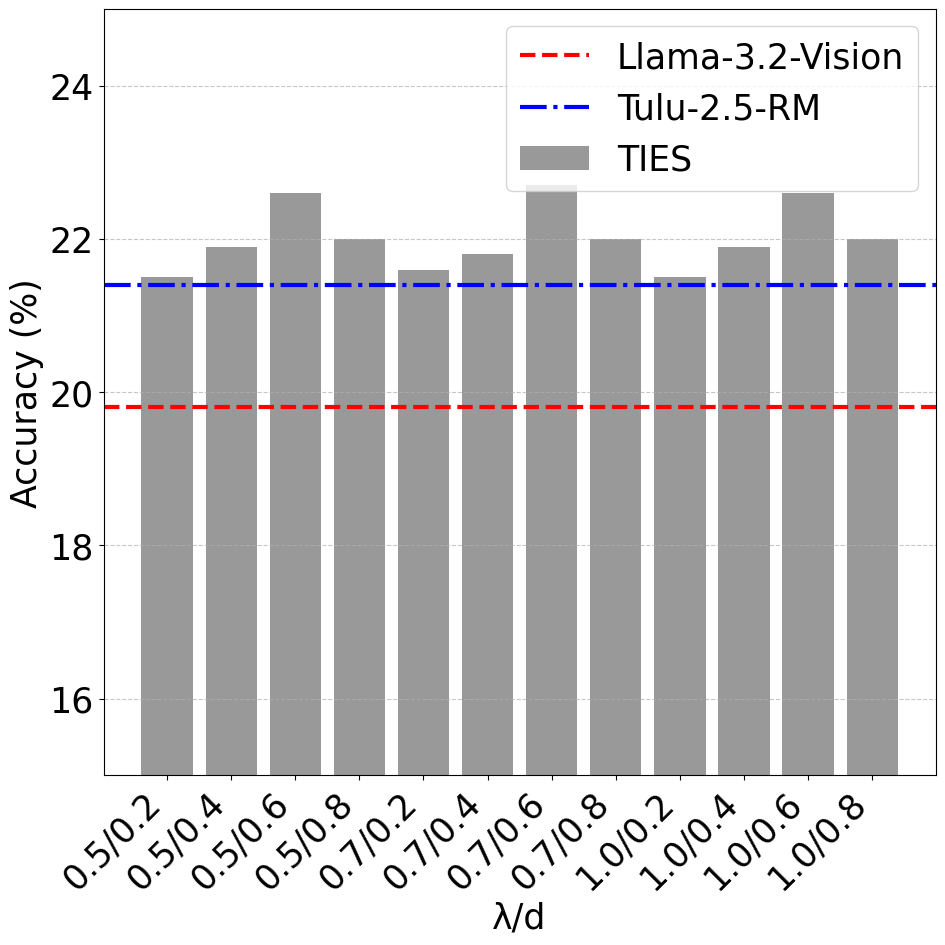}
         \caption{MMMU-Pro (Vision)}
     \end{subfigure}
        \caption{Full results of merging \texttt{Llama-3.2-Vision} and \texttt{Tulu-2.5-RM} (\texttt{TIES})}
        \vspace{-10pt}
        \label{fig:full_tulu2.5_ties}
\end{figure*}

\begin{figure*}[ht]
     \centering
     \begin{subfigure}[b]{0.245\linewidth}
         \centering
         \includegraphics[width=\textwidth]{figures/full_results/tulu2.5/dare_tv/vlrb.png}
         \caption{VL-RewardBench}
     \end{subfigure}
     \hfill
     \begin{subfigure}[b]{0.245\linewidth}
         \centering
         \includegraphics[width=\textwidth]{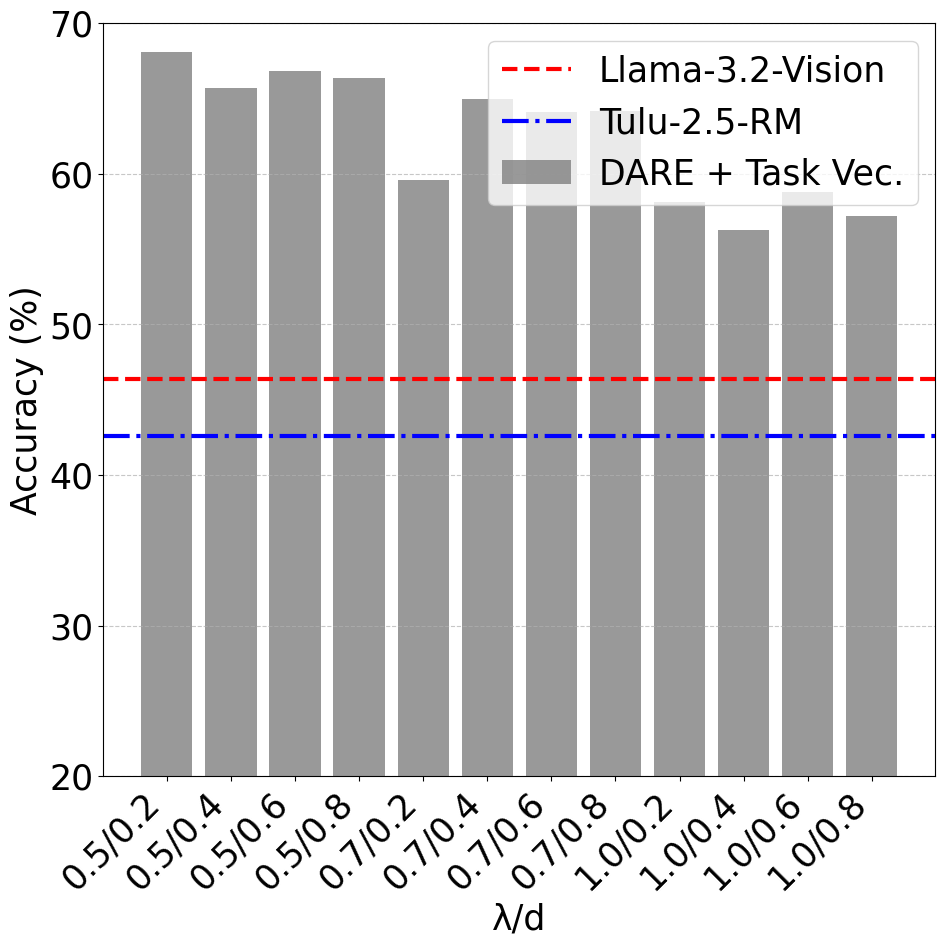}
         \caption{TextVQA}
     \end{subfigure}
     \hfill
      \begin{subfigure}[b]{0.245\linewidth}
         \centering
         \includegraphics[width=\textwidth]{figures/full_results/tulu2.5/dare_tv/mmmu_std.png}
         \caption{MMMU-Pro (Standard)}
     \end{subfigure}
     \hfill
     \begin{subfigure}[b]{0.245\linewidth}
         \centering
         \includegraphics[width=\textwidth]{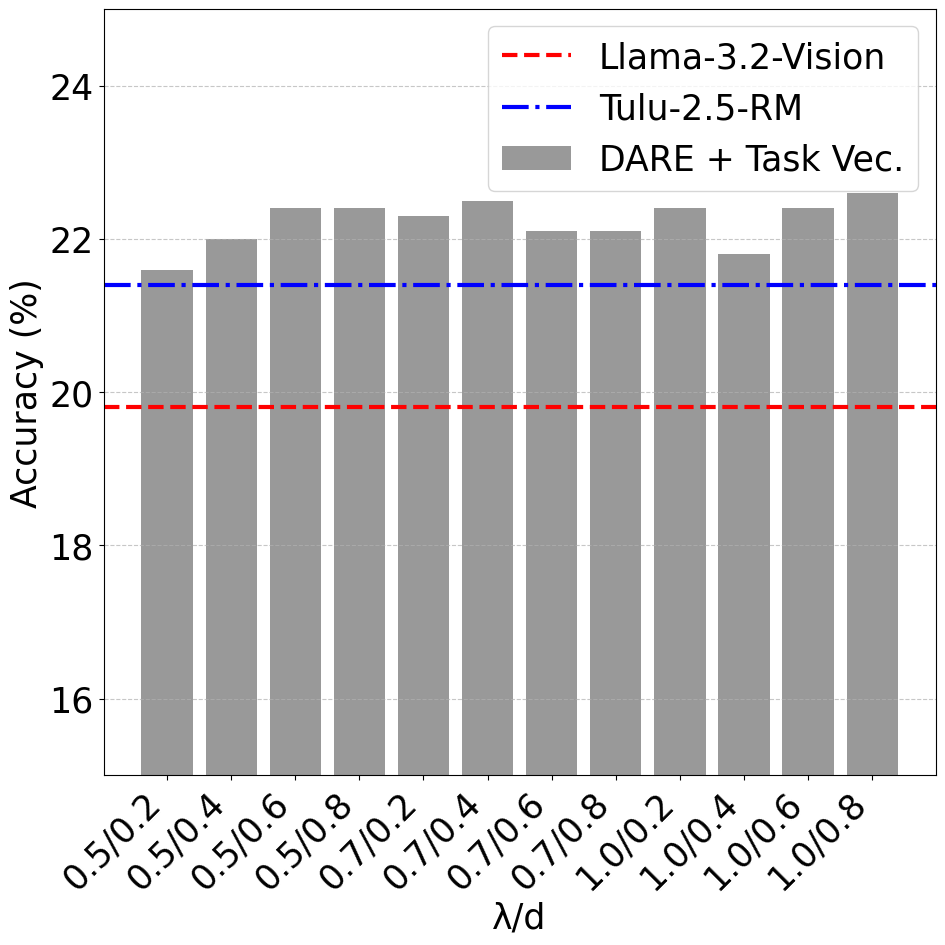}
         \caption{MMMU-Pro (Vision)}
     \end{subfigure}
        \caption{Full results of merging \texttt{Llama-3.2-Vision} and \texttt{Tulu-2.5-RM} (\texttt{DARE + Task Vec.})}
        \vspace{-10pt}
        \label{fig:full_tulu2.5_dare_tv}
\end{figure*}

\begin{figure*}[ht]
     \centering
     \begin{subfigure}[b]{0.245\linewidth}
         \centering
         \includegraphics[width=\textwidth]{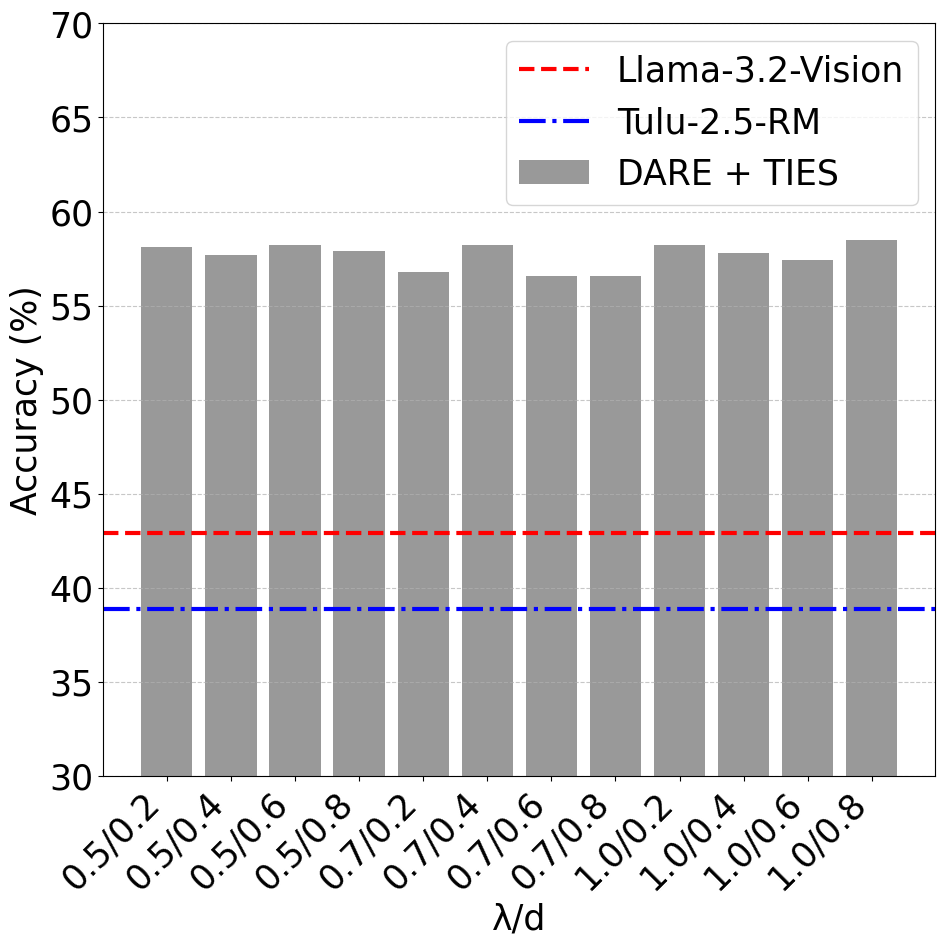}
         \caption{VL-RewardBench}
     \end{subfigure}
     \hfill
     \begin{subfigure}[b]{0.245\linewidth}
         \centering
         \includegraphics[width=\textwidth]{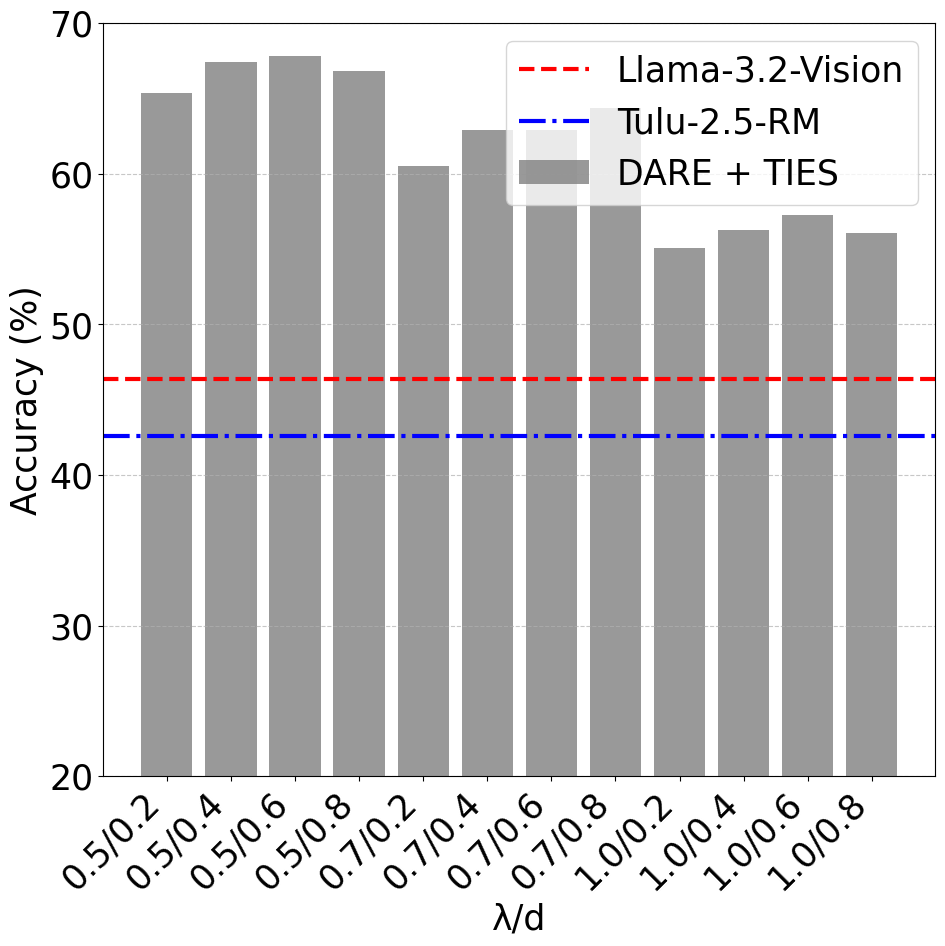}
         \caption{TextVQA}
     \end{subfigure}
     \hfill
      \begin{subfigure}[b]{0.245\linewidth}
         \centering
         \includegraphics[width=\textwidth]{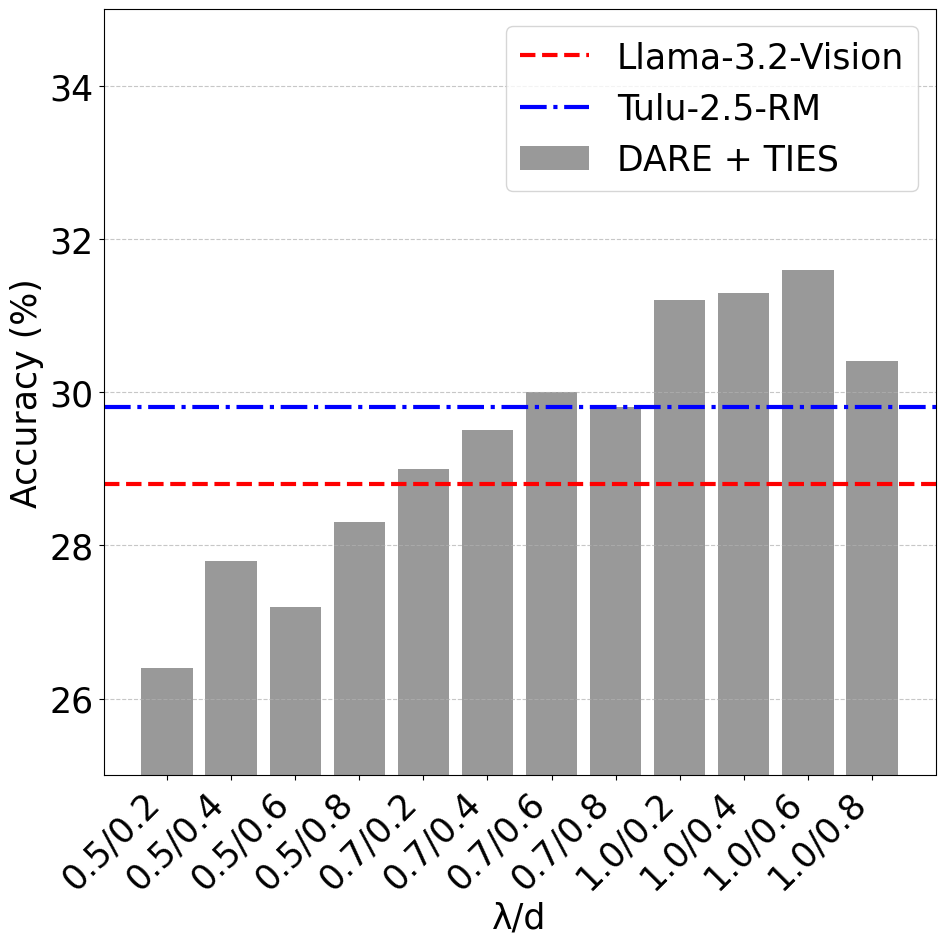}
         \caption{MMMU-Pro (Standard)}
     \end{subfigure}
     \hfill
     \begin{subfigure}[b]{0.245\linewidth}
         \centering
         \includegraphics[width=\textwidth]{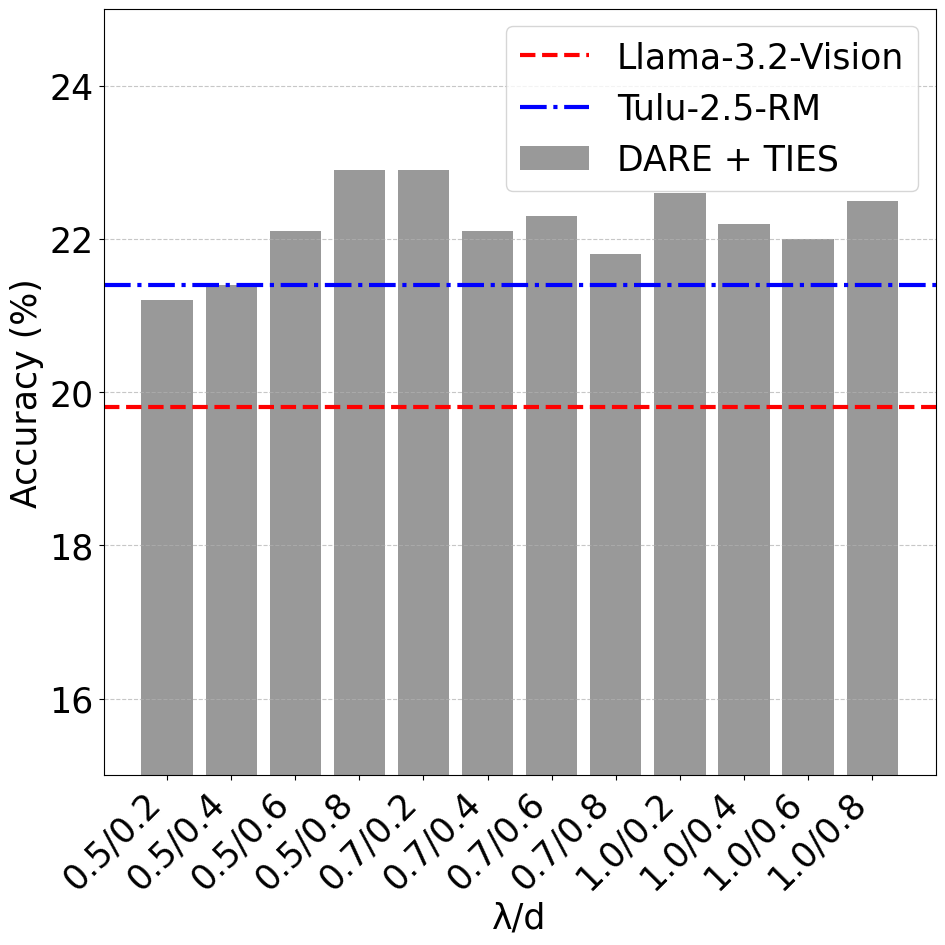}
         \caption{MMMU-Pro (Vision)}
     \end{subfigure}
        \caption{Full results of merging \texttt{Llama-3.2-Vision} and \texttt{Tulu-2.5-RM} (\texttt{DARE + TIES})}
        \vspace{-10pt}
        \label{fig:full_tulu2.5_dare_ties}
\end{figure*}

\begin{figure*}[ht]
     \centering
     \begin{subfigure}[b]{0.245\linewidth}
         \centering
         \includegraphics[width=\textwidth]{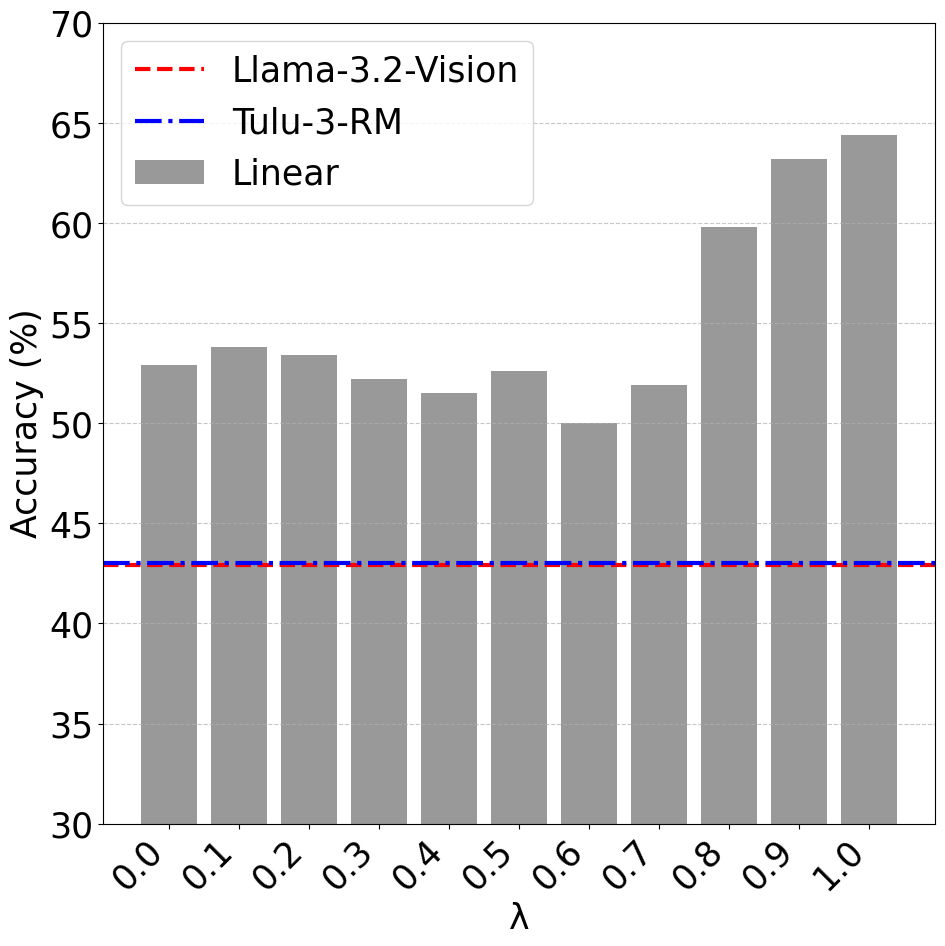}
         \caption{VL-RewardBench}
     \end{subfigure}
     \hfill
     \begin{subfigure}[b]{0.245\linewidth}
         \centering
         \includegraphics[width=\textwidth]{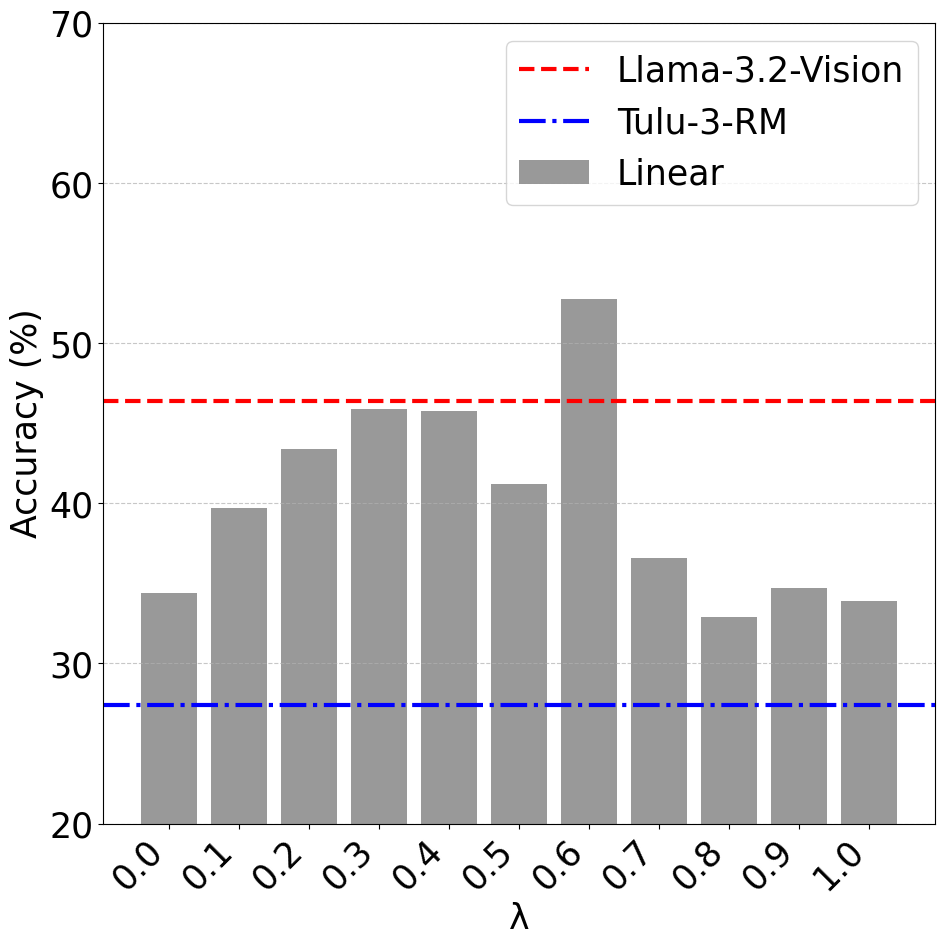}
         \caption{TextVQA}
     \end{subfigure}
     \hfill
      \begin{subfigure}[b]{0.245\linewidth}
         \centering
         \includegraphics[width=\textwidth]{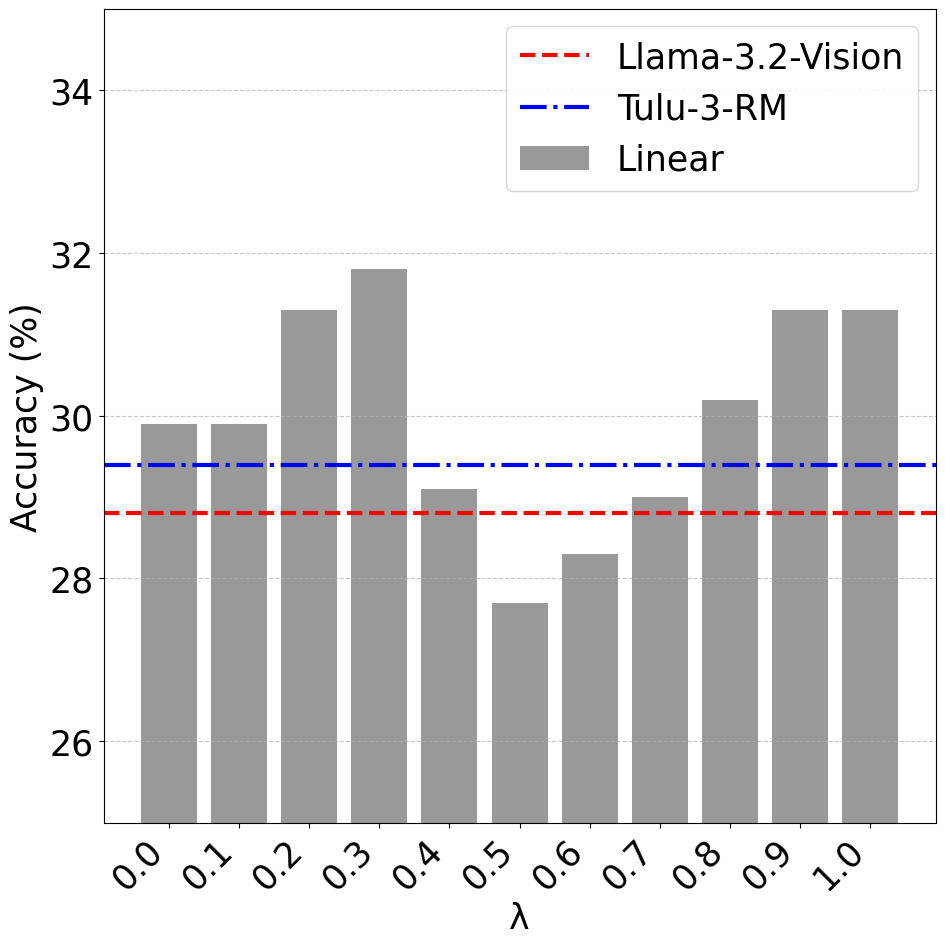}
         \caption{MMMU-Pro (Standard)}
     \end{subfigure}
     \hfill
     \begin{subfigure}[b]{0.245\linewidth}
         \centering
         \includegraphics[width=\textwidth]{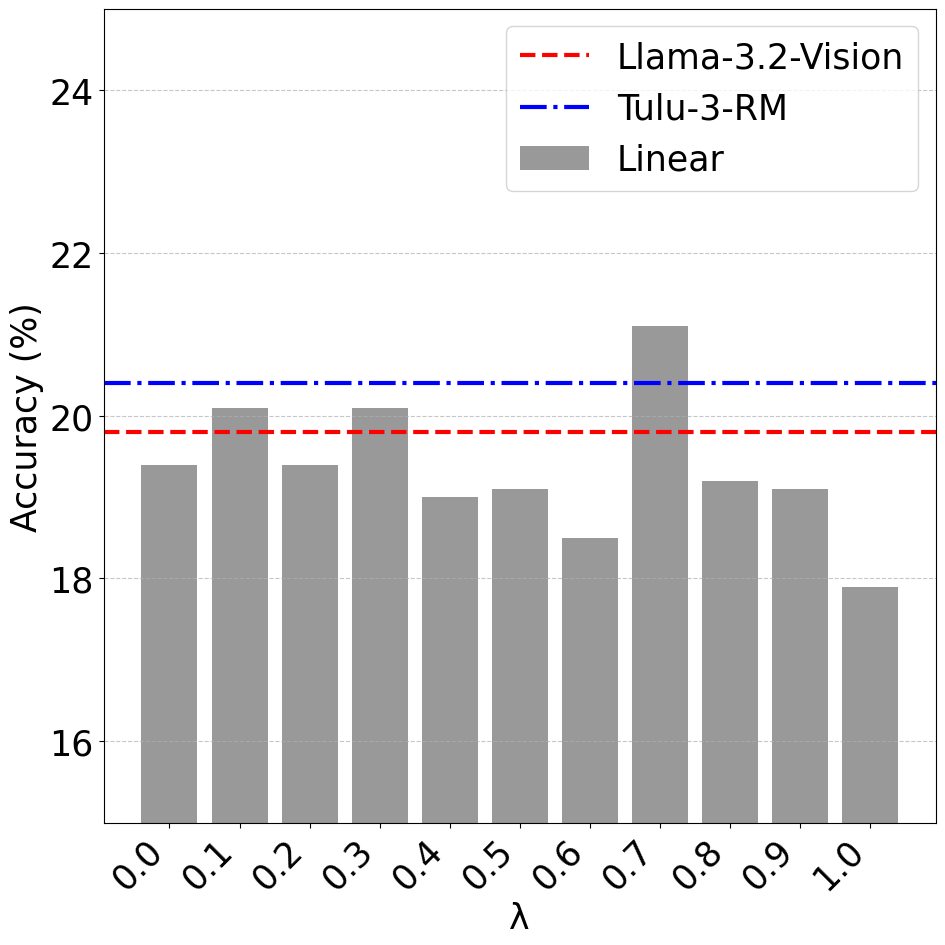}
         \caption{MMMU-Pro (Vision)}
     \end{subfigure}
        \caption{Full results of merging \texttt{Llama-3.2-Vision} and \texttt{Tulu-3-RM} (\texttt{Linear})}
        \vspace{-10pt}
        \label{fig:full_tulu3_linear}
\end{figure*}

\begin{figure*}[ht]
     \centering
     \begin{subfigure}[b]{0.245\linewidth}
         \centering
         \includegraphics[width=\textwidth]{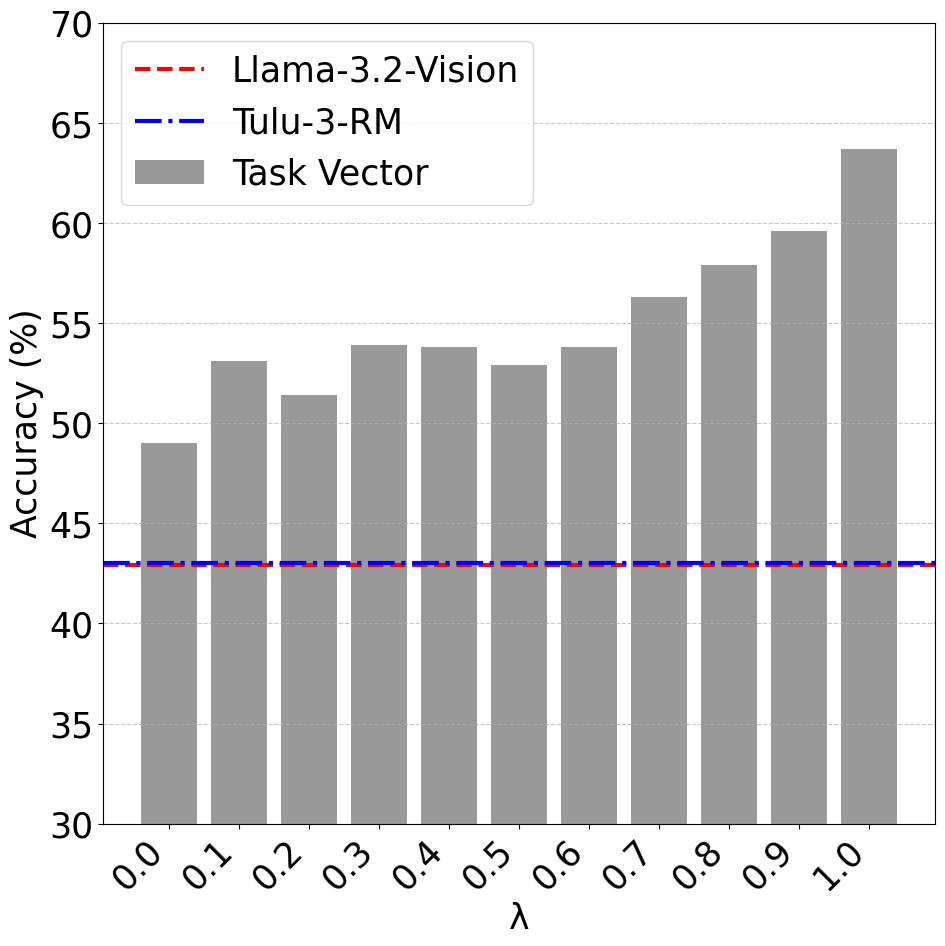}
         \caption{VL-RewardBench}
     \end{subfigure}
     \hfill
     \begin{subfigure}[b]{0.245\linewidth}
         \centering
         \includegraphics[width=\textwidth]{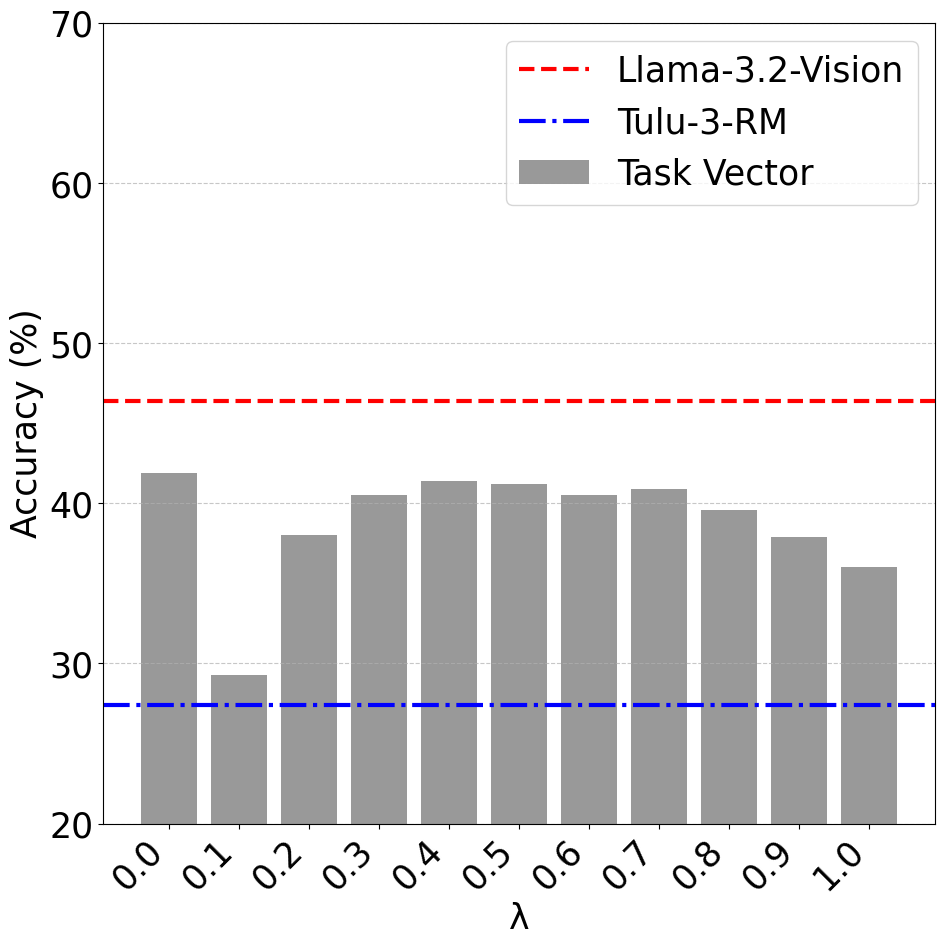}
         \caption{TextVQA}
     \end{subfigure}
     \hfill
      \begin{subfigure}[b]{0.245\linewidth}
         \centering
         \includegraphics[width=\textwidth]{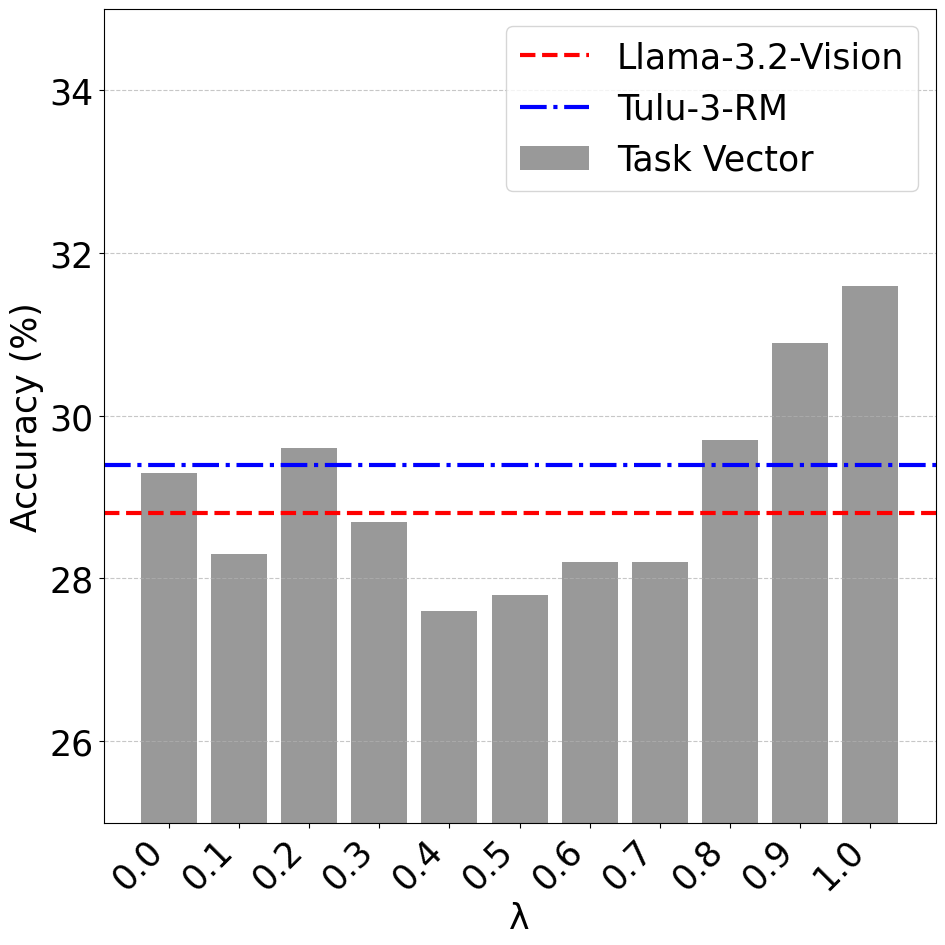}
         \caption{MMMU-Pro (Standard)}
     \end{subfigure}
     \hfill
     \begin{subfigure}[b]{0.245\linewidth}
         \centering
         \includegraphics[width=\textwidth]{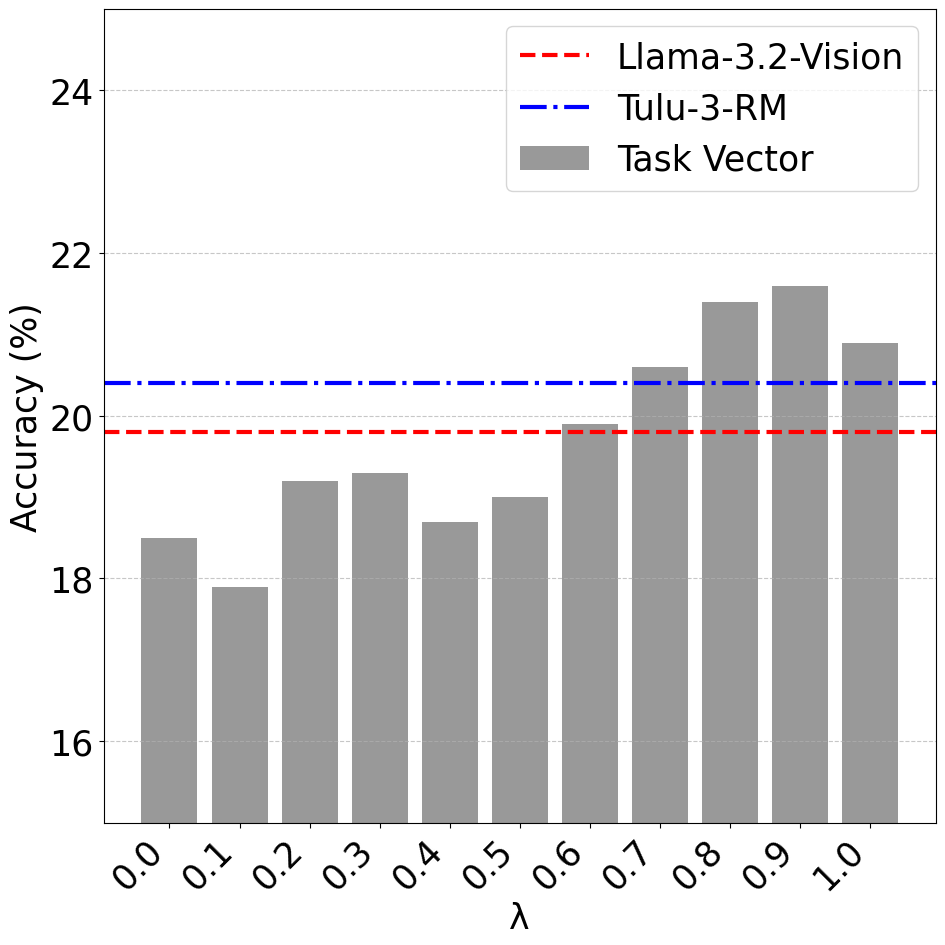}
         \caption{MMMU-Pro (Vision)}
     \end{subfigure}
        \caption{Full results of merging \texttt{Llama-3.2-Vision} and \texttt{Tulu-3-RM} (\texttt{Task Vec.})}
        \vspace{-10pt}
        \label{fig:full_tulu3_task_vector}
\end{figure*}

\begin{figure*}[ht]
     \centering
     \begin{subfigure}[b]{0.245\linewidth}
         \centering
         \includegraphics[width=\textwidth]{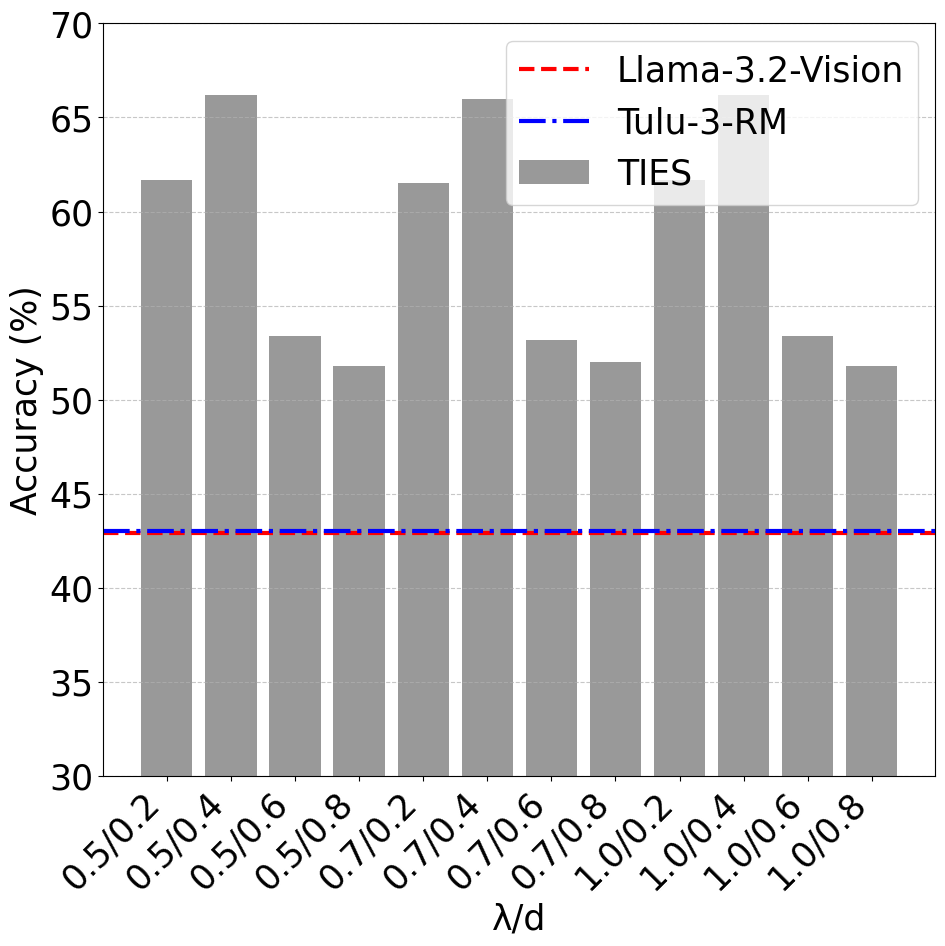}
         \caption{VL-RewardBench}
     \end{subfigure}
     \hfill
     \begin{subfigure}[b]{0.245\linewidth}
         \centering
         \includegraphics[width=\textwidth]{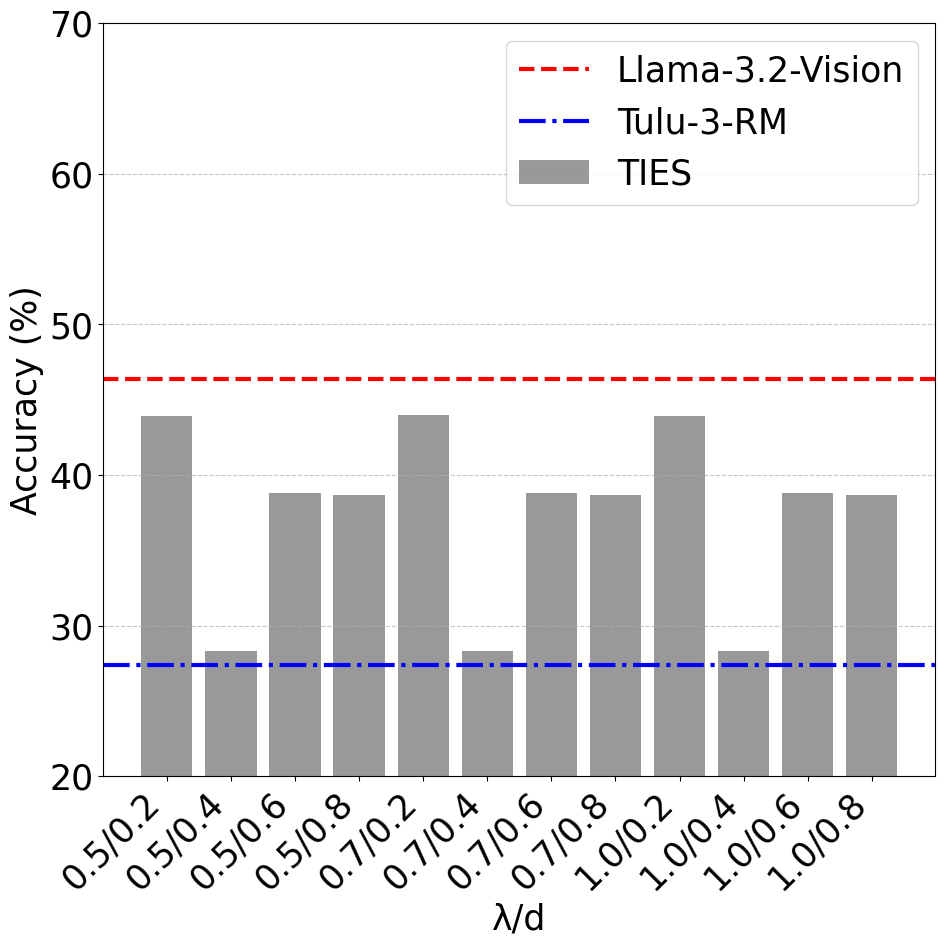}
         \caption{TextVQA}
     \end{subfigure}
     \hfill
      \begin{subfigure}[b]{0.245\linewidth}
         \centering
         \includegraphics[width=\textwidth]{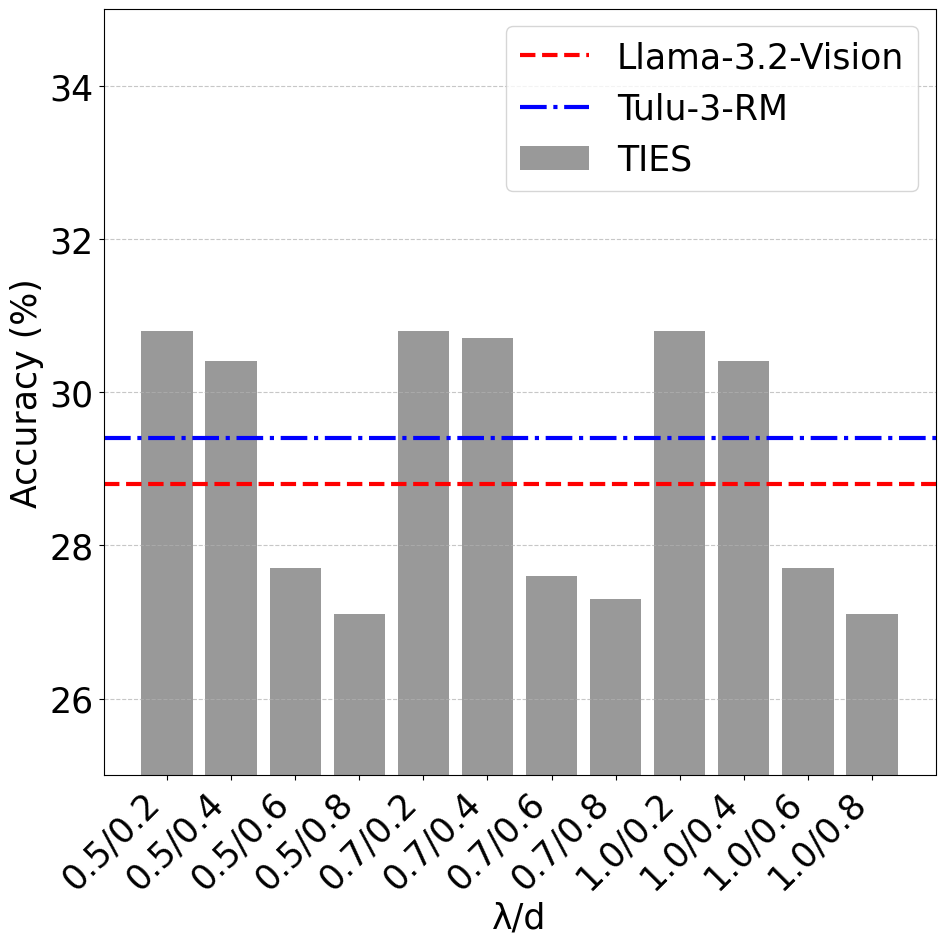}
         \caption{MMMU-Pro (Standard)}
     \end{subfigure}
     \hfill
     \begin{subfigure}[b]{0.245\linewidth}
         \centering
         \includegraphics[width=\textwidth]{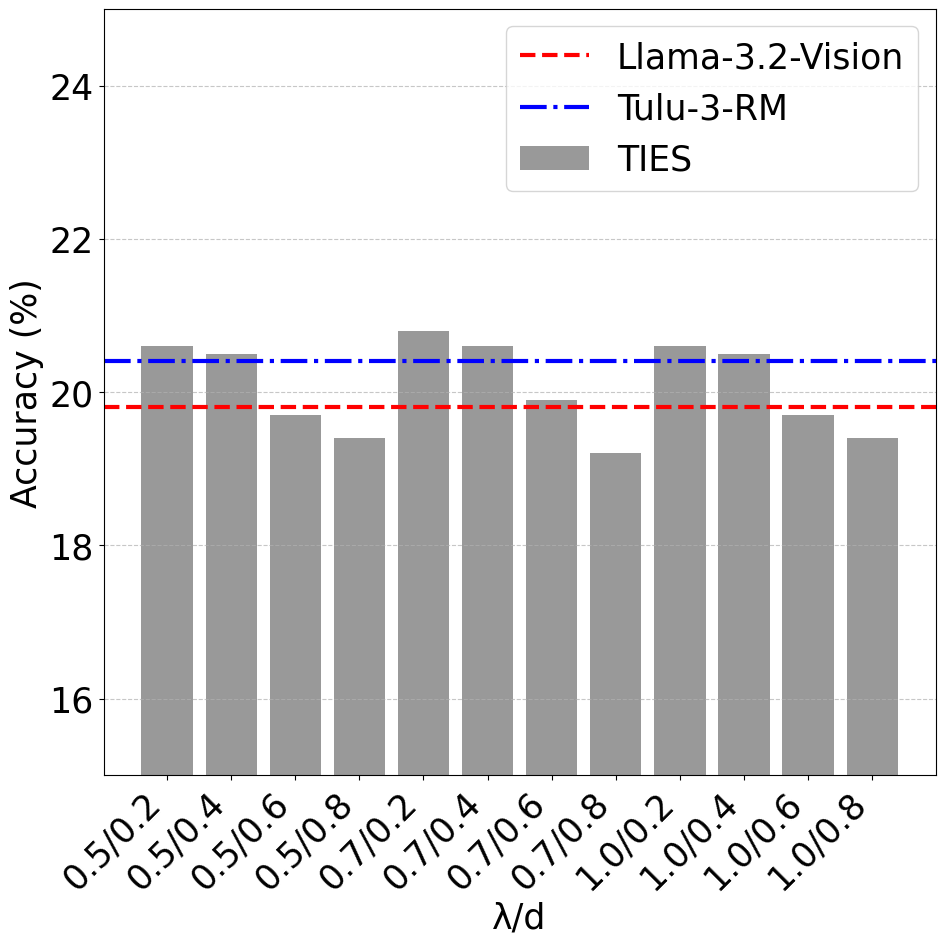}
         \caption{MMMU-Pro (Vision)}
     \end{subfigure}
        \caption{Full results of merging \texttt{Llama-3.2-Vision} and \texttt{Tulu-3-RM} (\texttt{TIES})}
        \vspace{-10pt}
        \label{fig:full_tulu3_ties}
\end{figure*}

\begin{figure*}[ht]
     \centering
     \begin{subfigure}[b]{0.245\linewidth}
         \centering
         \includegraphics[width=\textwidth]{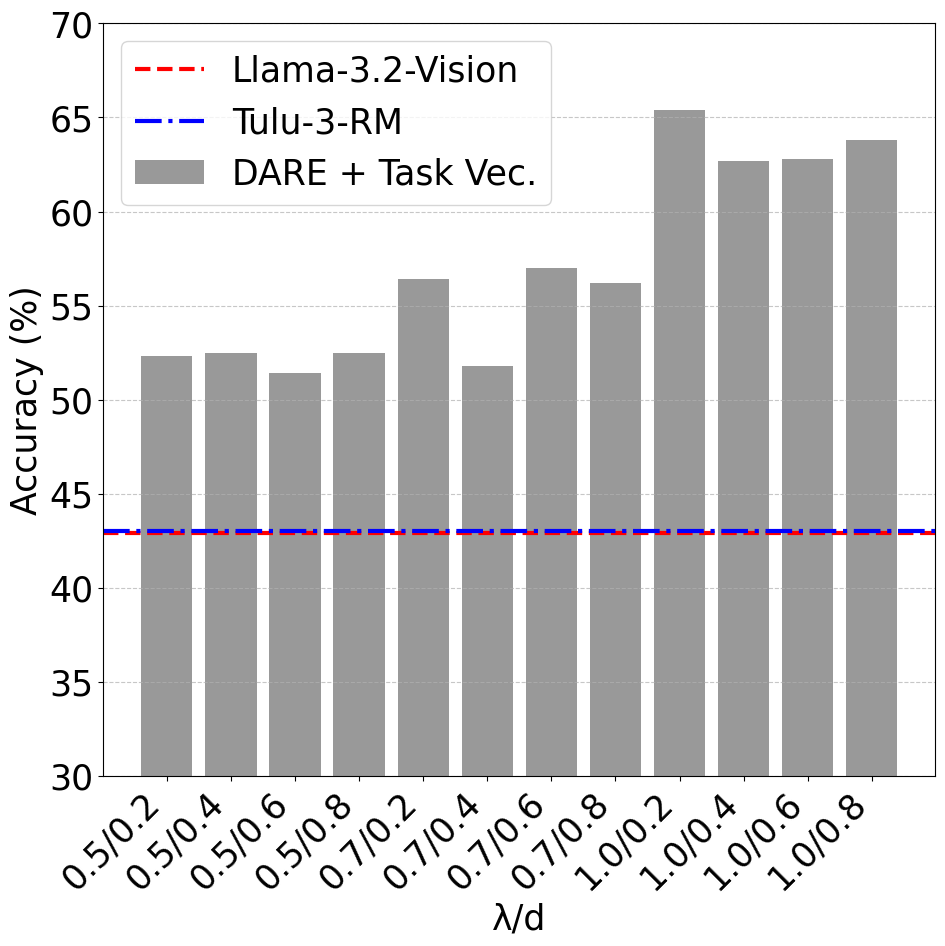}
         \caption{VL-RewardBench}
     \end{subfigure}
     \hfill
     \begin{subfigure}[b]{0.245\linewidth}
         \centering
         \includegraphics[width=\textwidth]{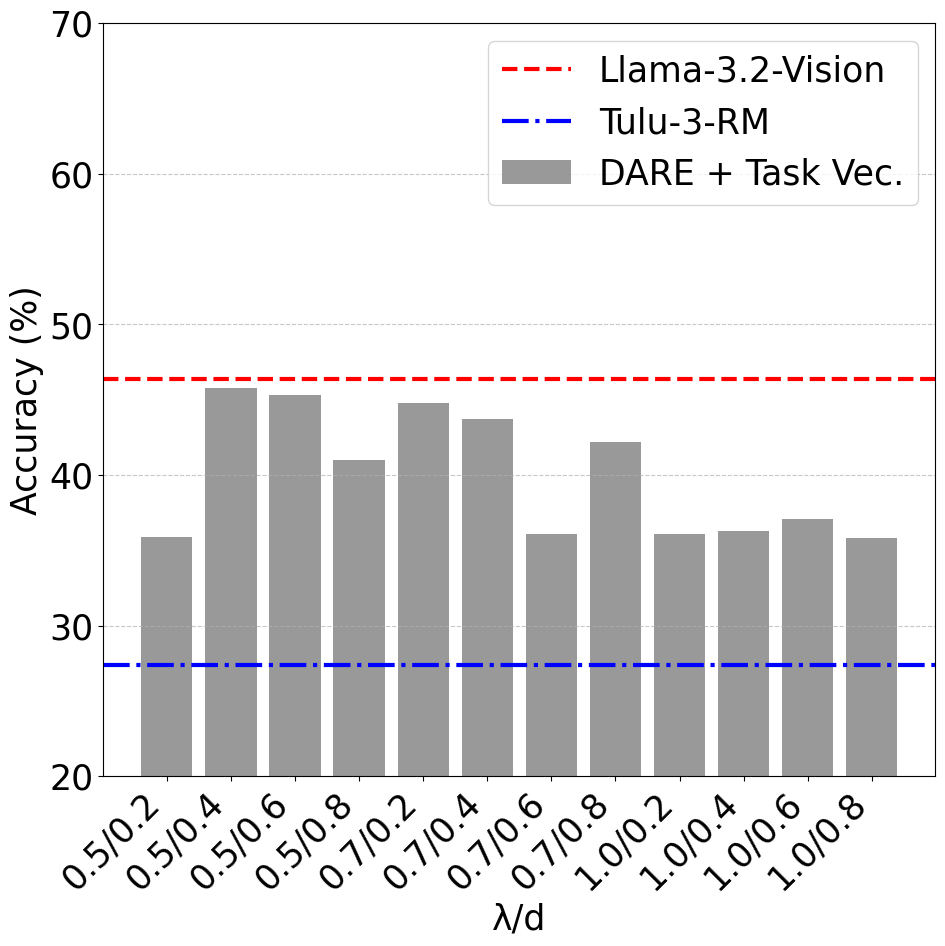}
         \caption{TextVQA}
     \end{subfigure}
     \hfill
      \begin{subfigure}[b]{0.245\linewidth}
         \centering
         \includegraphics[width=\textwidth]{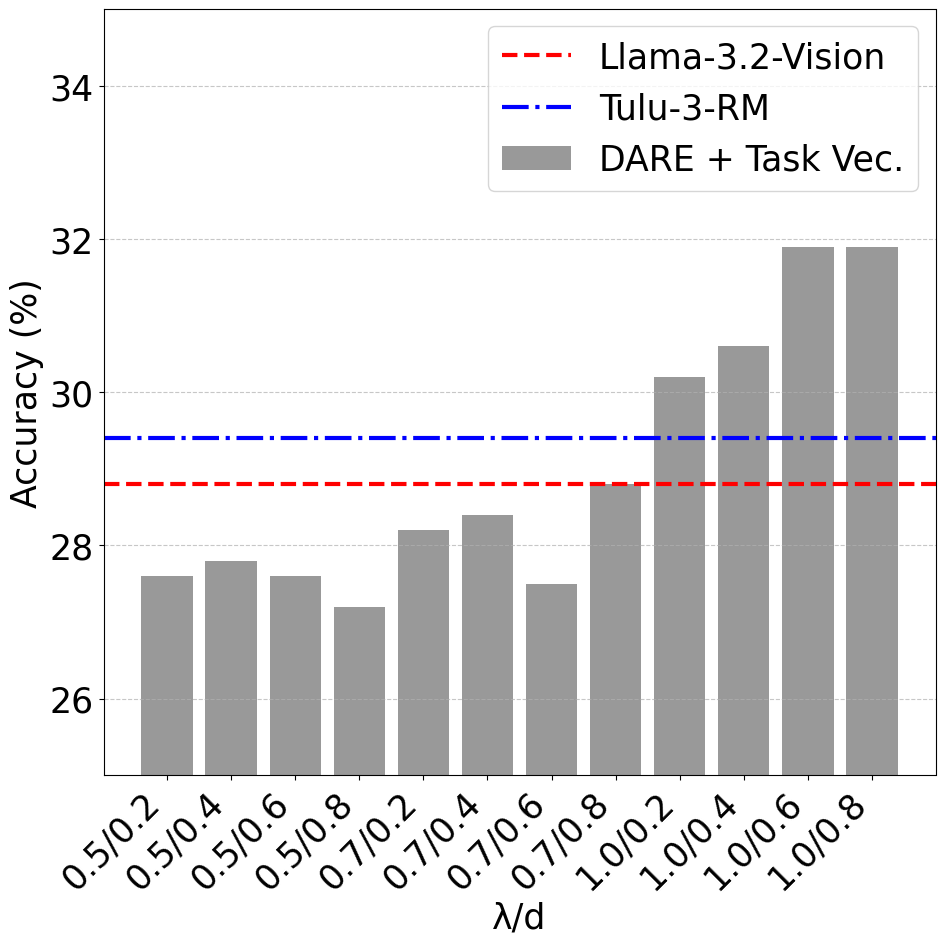}
         \caption{MMMU-Pro (Standard)}
     \end{subfigure}
     \hfill
     \begin{subfigure}[b]{0.245\linewidth}
         \centering
         \includegraphics[width=\textwidth]{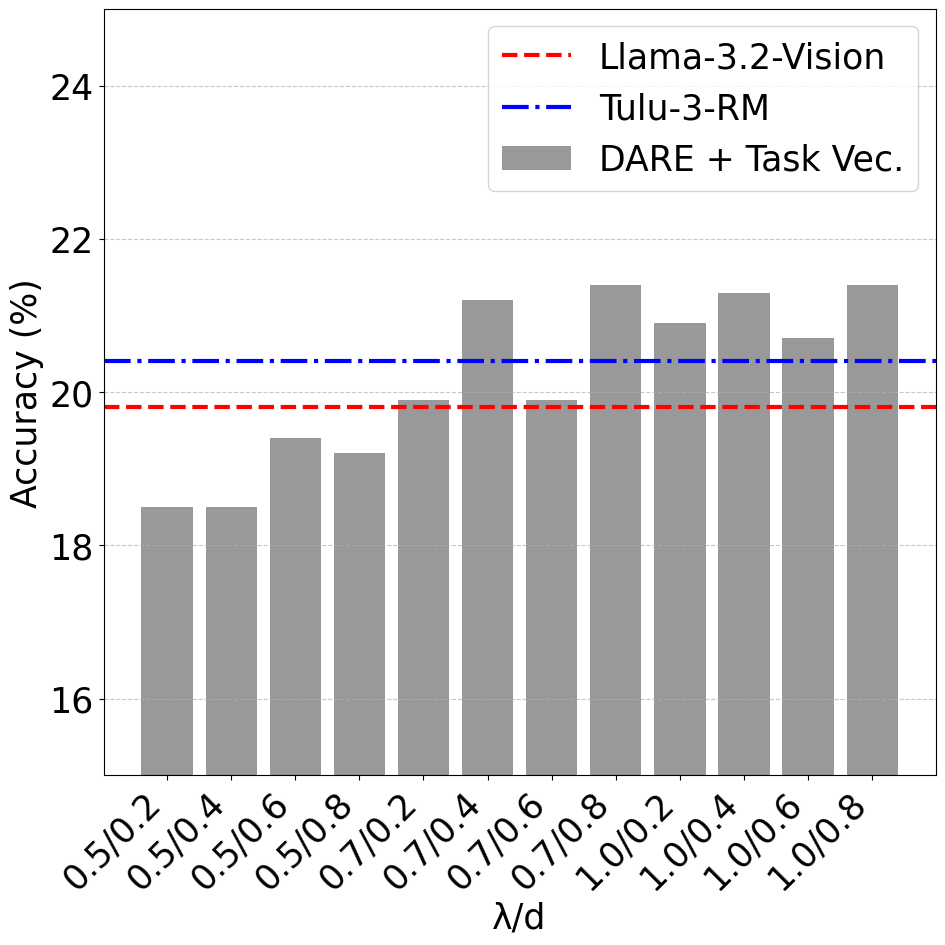}
         \caption{MMMU-Pro (Vision)}
     \end{subfigure}
        \caption{Full results of merging \texttt{Llama-3.2-Vision} and \texttt{Tulu-3-RM} (\texttt{DARE + Task Vec.})}
        \vspace{-10pt}
        \label{fig:full_tulu3_dare_tv}
\end{figure*}

\begin{figure*}[ht]
     \centering
     \begin{subfigure}[b]{0.245\linewidth}
         \centering
         \includegraphics[width=\textwidth]{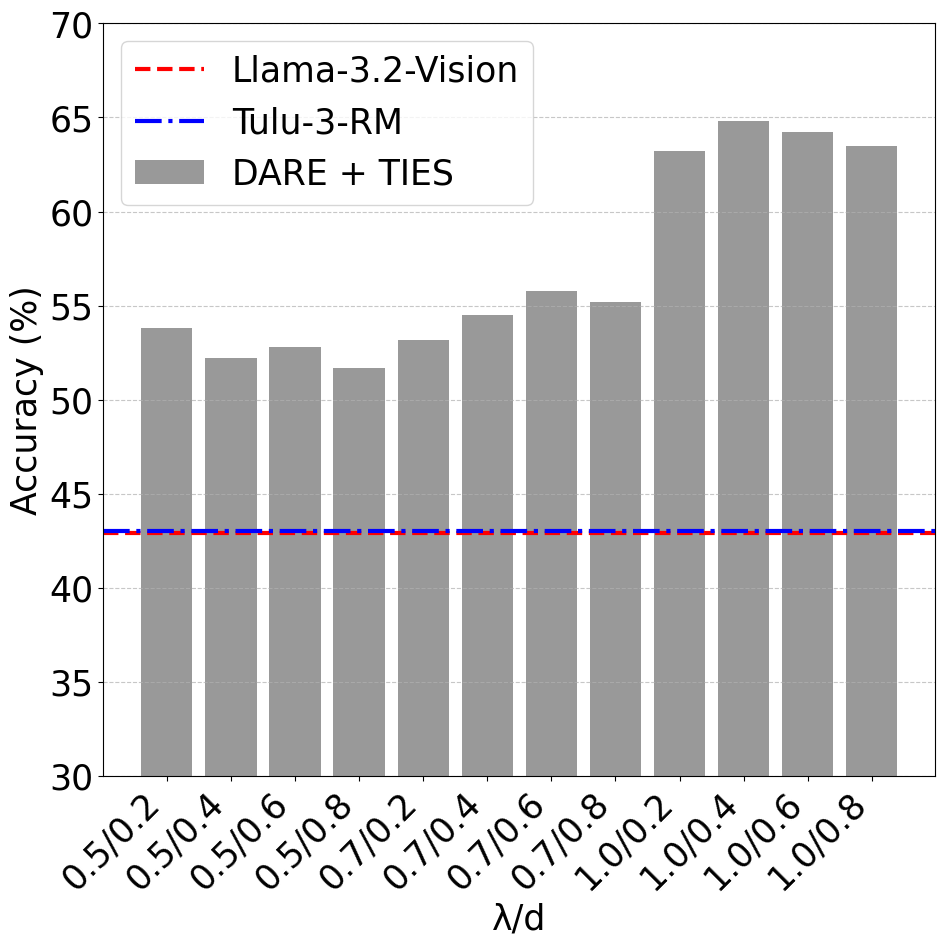}
         \caption{VL-RewardBench}
     \end{subfigure}
     \hfill
     \begin{subfigure}[b]{0.245\linewidth}
         \centering
         \includegraphics[width=\textwidth]{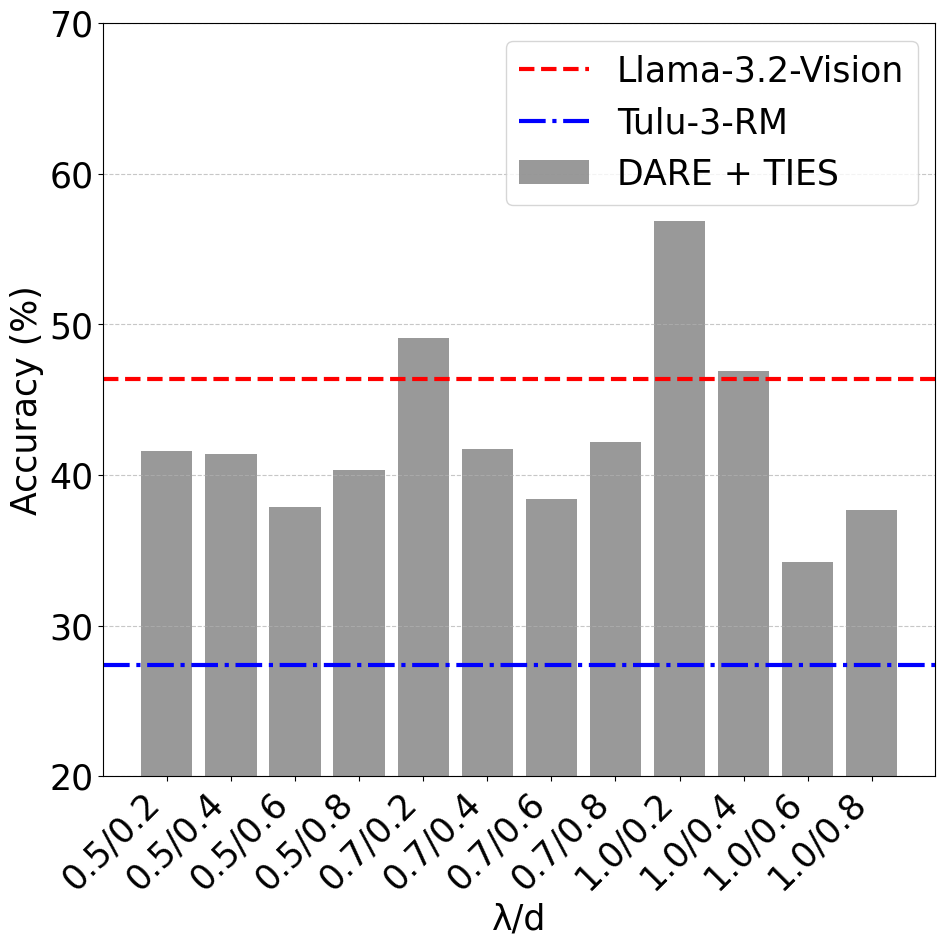}
         \caption{TextVQA}
     \end{subfigure}
     \hfill
      \begin{subfigure}[b]{0.245\linewidth}
         \centering
         \includegraphics[width=\textwidth]{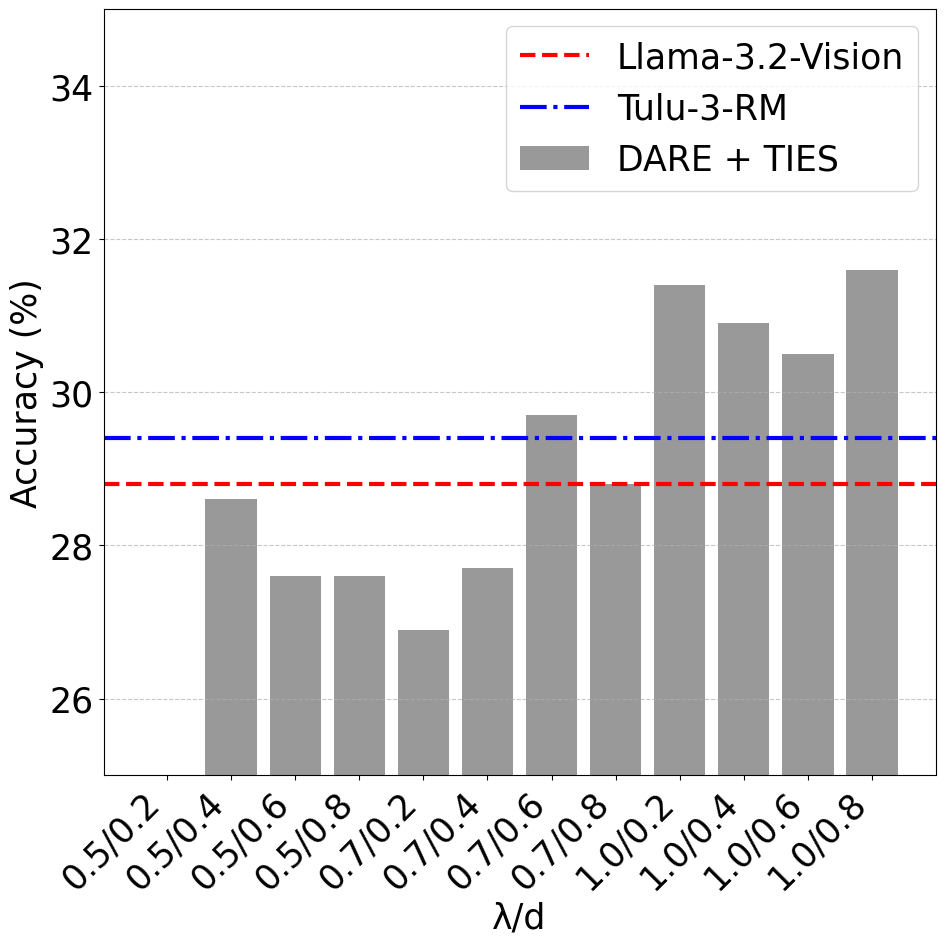}
         \caption{MMMU-Pro (Standard)}
     \end{subfigure}
     \hfill
     \begin{subfigure}[b]{0.245\linewidth}
         \centering
         \includegraphics[width=\textwidth]{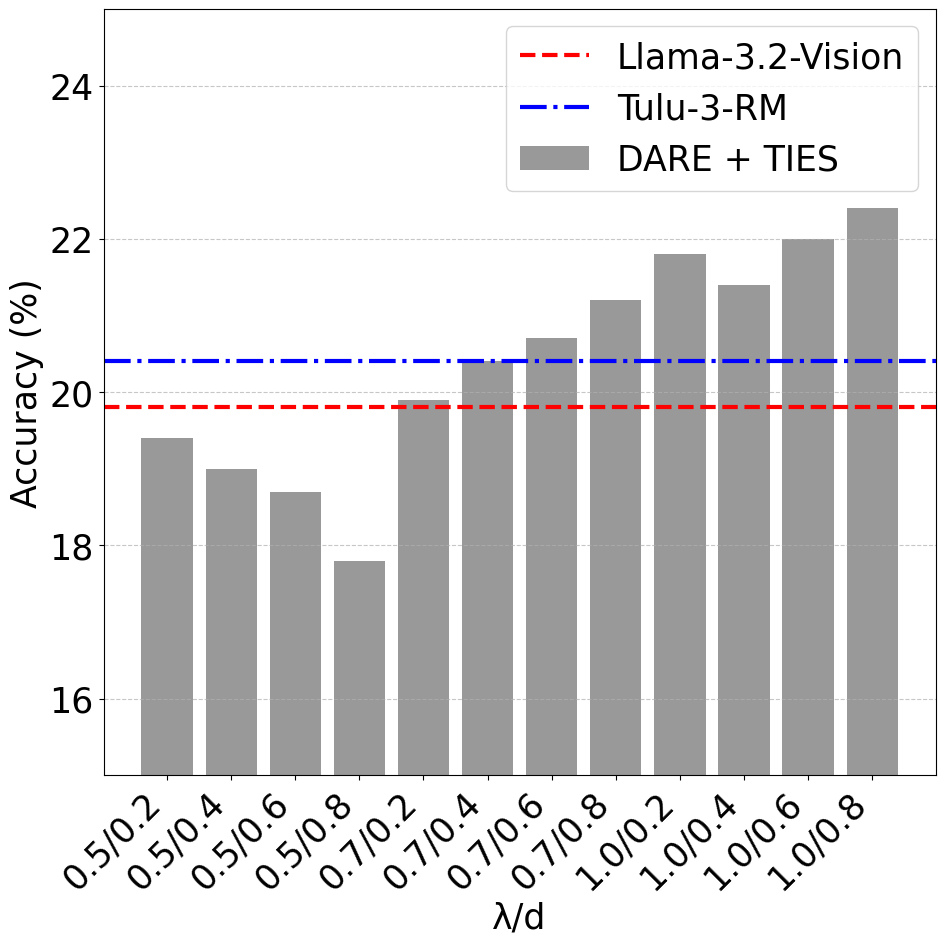}
         \caption{MMMU-Pro (Vision)}
     \end{subfigure}
        \caption{Full results of merging \texttt{Llama-3.2-Vision} and \texttt{Tulu-3-RM} (\texttt{DARE + TIES})}
        \vspace{-10pt}
        \label{fig:full_tulu3_dare_ties}
\end{figure*}

%% file: tables/hyper_select.tex
\begin{table*}[t]
\centering
    \resizebox{0.68\textwidth}{!}{
        \begin{tabular}{c | c c c c c c c c c c c}
            \toprule
            $\lambda$ & 0.0 & 0.1 & 0.2 & 0.3 & 0.4 & 0.5 & 0.6 & 0.7 & 0.8 & 0.9 & 1.0 \\
            \midrule
            Overall Acc. & 49.8 & 52.3 & 50.3 & \textbf{52.5} & 52.0 & 49.0 & 47.3 & 46.5 & 46.5 & 50.3 & 47.0 \\
            \bottomrule
        \end{tabular}
    }
    \caption{\texttt{Linear} merging using \texttt{Tulu-2.5-RM} as the text-based RM, evaluated on sampled RLAIF-V.}
    \label{tab:hyper_select_linear_tulu25}
    \vspace{1em}

    \resizebox{0.68\textwidth}{!}{
        \begin{tabular}{c | c c c c c c c c c c c}
            \toprule
            $\lambda$ & 0.0 & 0.1 & 0.2 & 0.3 & 0.4 & 0.5 & 0.6 & 0.7 & 0.8 & 0.9 & 1.0 \\
            \midrule
            Overall Acc. & \textbf{55.3} & 50.0 & 53.3 & 54.5 & 53.5 & 49.3 & 52.8 & 54.0 & 53.8 & 54.8 & \textbf{55.3}* \\
            \bottomrule
        \end{tabular}
    }
    \caption{\texttt{Task Vec.} merging using \texttt{Tulu-2.5-RM} as the text-based RM, evaluated on sampled RLAIF-V.}
    \label{tab:hyper_select_task_vector_tulu25}
    \vspace{1em}

    \resizebox{0.72\textwidth}{!}{
        \begin{tabular}{c | c c c c | c c c c | c c c c}
            \toprule
            $\lambda$ & \multicolumn{4}{c|}{1.0} & \multicolumn{4}{c|}{0.7} & \multicolumn{4}{c}{0.5} \\
            \midrule
            $d$ & 0.8 & 0.6 & 0.4 & 0.2 & 0.8 & 0.6 & 0.4 & 0.2 & 0.8 & 0.6 & 0.4 & 0.2 \\
            \midrule
            Overall Acc. & 53.5 & \textbf{53.8}* & 52.3 & 50.0 & 53.5 & \textbf{53.8} & 52.3 & 50.3 & 53.5 & \textbf{53.8} & 52.3 & 50.0 \\
            \bottomrule
        \end{tabular}
    }
    \caption{\texttt{TIES} merging using \texttt{Tulu-2.5-RM} as the text-based RM, evaluated on sampled RLAIF-V.}
    \label{tab:hyper_select_ties_tulu25}
    \vspace{1em}

    \resizebox{0.72\textwidth}{!}{
        \begin{tabular}{c | c c c c | c c c c | c c c c}
            \toprule
            $\lambda$ & \multicolumn{4}{c|}{1.0} & \multicolumn{4}{c|}{0.7} & \multicolumn{4}{c}{0.5} \\
            \midrule
            $d$ & 0.8 & 0.6 & 0.4 & 0.2 & 0.8 & 0.6 & 0.4 & 0.2 & 0.8 & 0.6 & 0.4 & 0.2 \\
            \midrule
            Overall Acc. & 55.3 & \textbf{56.5} & 54.5 & 55.3 & 54.5 & 54.0 & 53.5 & 55.8 & 49.0 & 49.3 & 51.8 & 54.8 \\
            \bottomrule
        \end{tabular}
    }
    \caption{\texttt{DARE} + \texttt{Task Vec.} merging using \texttt{Tulu-2.5-RM} as the text-based RM, evaluated on sampled RLAIF-V.}
    \label{tab:hyper_select_dare_linear_tulu25}
    \vspace{1em}

    \resizebox{0.72\textwidth}{!}{
        \begin{tabular}{c | c c c c | c c c c | c c c c}
            \toprule
            $\lambda$ & \multicolumn{4}{c|}{1.0} & \multicolumn{4}{c|}{0.7} & \multicolumn{4}{c}{0.5} \\
            \midrule
            $d$ & 0.8 & 0.6 & 0.4 & 0.2 & 0.8 & 0.6 & 0.4 & 0.2 & 0.8 & 0.6 & 0.4 & 0.2 \\
            \midrule
            Overall Acc. & 55.5 & \textbf{56.0}* & \textbf{56.0} & 55.5 & 53.3 & 54.3 & 53.8 & 52.3 & 51.5 & 49.8 & 51.5 & 51.8 \\
            \bottomrule
        \end{tabular}
    }
    \caption{\texttt{DARE} + \texttt{TIES} merging using \texttt{Tulu-2.5-RM} as the text-based RM, evaluated on sampled RLAIF-V.}
    \label{tab:hyper_select_dare_ties_tulu25}

    \vspace{1em}
    \resizebox{0.68\textwidth}{!}{
        \begin{tabular}{c | c c c c c c c c c c c}
            \toprule
            $\lambda$ & 0.0 & 0.1 & 0.2 & 0.3 & 0.4 & 0.5 & 0.6 & 0.7 & 0.8 & 0.9 & 1.0 \\
            \midrule
            Overall Acc. & 51.5 & 46.8 & 50.3 & 49.3 & \textbf{52.0} & 50.8 & 49.3 & 47.3 & 49.5 & 49.3 & 51.3 \\
            \bottomrule
        \end{tabular}
    }
    \caption{\texttt{Linear} merging using \texttt{Tulu-3-RM} as the text-based RM, evaluated on sampled RLAIF-V.}
    \label{tab:hyper_select_linear_tulu3}
    \vspace{1em}

    \resizebox{0.68\textwidth}{!}{
        \begin{tabular}{c | c c c c c c c c c c c}
            \toprule
            $\lambda$ & 0.0 & 0.1 & 0.2 & 0.3 & 0.4 & 0.5 & 0.6 & 0.7 & 0.8 & 0.9 & 1.0 \\
            \midrule
            Overall Acc. & 49.3 & 53.5 & 49.8 & 49.8 & 51.0 & 51.0 & 53.8 & 53.0 & 53.0 & 50.3 & \textbf{55.3} \\
            \bottomrule
        \end{tabular}
    }
    \caption{\texttt{Task Vec.} merging using \texttt{Tulu-3-RM} as the text-based RM, evaluated on sampled RLAIF-V.}
    \label{tab:hyper_select_task_vector_tulu3}
    \vspace{1em}

    \resizebox{0.72\textwidth}{!}{
        \begin{tabular}{c | c c c c | c c c c | c c c c}
            \toprule
            $\lambda$ & \multicolumn{4}{c|}{1.0} & \multicolumn{4}{c|}{0.7} & \multicolumn{4}{c}{0.5} \\
            \midrule
            $d$ & 0.8 & 0.6 & 0.4 & 0.2 & 0.8 & 0.6 & 0.4 & 0.2 & 0.8 & 0.6 & 0.4 & 0.2 \\
            \midrule
            Overall Acc. & 53.5 & 53.3 & 54.0 & 51.0 & 53.8 & \textbf{54.3} & \textbf{54.3}* & 51.5 & 53.5 & 53.3 & 54.0 & 51.0 \\
            \bottomrule
        \end{tabular}
    }
    \caption{\texttt{TIES} merging using \texttt{Tulu-3-RM} as the text-based RM, evaluated on sampled RLAIF-V.}
    \label{tab:hyper_select_ties_tulu3}
    \vspace{1em}

    \resizebox{0.72\textwidth}{!}{
        \begin{tabular}{c | c c c c | c c c c | c c c c}
            \toprule
            $\lambda$ & \multicolumn{4}{c|}{1.0} & \multicolumn{4}{c|}{0.7} & \multicolumn{4}{c}{0.5} \\
            \midrule
            $d$ & 0.8 & 0.6 & 0.4 & 0.2 & 0.8 & 0.6 & 0.4 & 0.2 & 0.8 & 0.6 & 0.4 & 0.2 \\
            \midrule
            Overall Acc. & 54.8 & 55.8 & 55.3 & \textbf{58.0} & 53.8 & 53.8 & 52.3 & 50.3 & 50.0 & 50.3 & 51.0 & 51.5 \\
            \bottomrule
        \end{tabular}
    }
    \caption{\texttt{DARE} + \texttt{Task Vec.} merging using \texttt{Tulu-3-RM} as the text-based RM, evaluated on sampled RLAIF-V.}
    \label{tab:hyper_select_dare_linear_tulu3}
    \vspace{1em}

    \resizebox{0.72\textwidth}{!}{
        \begin{tabular}{c | c c c c | c c c c | c c c c}
            \toprule
            $\lambda$ & \multicolumn{4}{c|}{1.0} & \multicolumn{4}{c|}{0.7} & \multicolumn{4}{c}{0.5} \\
            \midrule
            $d$ & 0.8 & 0.6 & 0.4 & 0.2 & 0.8 & 0.6 & 0.4 & 0.2 & 0.8 & 0.6 & 0.4 & 0.2 \\
            \midrule
            Overall Acc. & 55.8 & 55.8 & 56.0 & \textbf{56.8} & 52.8 & 52.5 & 52.5 & 52.3 & 55.3 & 53.8 & 48.0 & 54.5 \\
            \bottomrule
        \end{tabular}
    }
    \caption{\texttt{DARE} + \texttt{TIES} merging using \texttt{Tulu-3-RM} as the text-based RM, evaluated on sampled RLAIF-V.}
    \label{tab:hyper_select_dare_ties_tulu3}
\end{table*}

%% file: acl_latex.bbl
\begin{thebibliography}{51}
\providecommand{\natexlab}[1]{#1}

\bibitem[{Anthropic(2024)}]{anthropic2024claude35}
Anthropic. 2024.
\newblock \href {https://www.anthropic.com/news/claude-3-5-sonnet} {Claude 3.5 sonnet}.
\newblock Accessed: 2025-02-04.

\bibitem[{Bai et~al.(2023{\natexlab{a}})Bai, Bai, Chu, Cui, Dang, Deng, Fan, Ge, Han, Huang et~al.}]{bai2023qwen}
Jinze Bai, Shuai Bai, Yunfei Chu, Zeyu Cui, Kai Dang, Xiaodong Deng, Yang Fan, Wenbin Ge, Yu~Han, Fei Huang, et~al. 2023{\natexlab{a}}.
\newblock Qwen technical report.
\newblock \emph{arXiv preprint arXiv:2309.16609}.

\bibitem[{Bai et~al.(2023{\natexlab{b}})Bai, Bai, Yang, Wang, Tan, Wang, Lin, Zhou, and Zhou}]{bai2023qwenvl}
Jinze Bai, Shuai Bai, Shusheng Yang, Shijie Wang, Sinan Tan, Peng Wang, Junyang Lin, Chang Zhou, and Jingren Zhou. 2023{\natexlab{b}}.
\newblock Qwen-vl: A frontier large vision-language model with versatile abilities.
\newblock \emph{arXiv preprint arXiv:2308.12966}.

\bibitem[{Bai et~al.(2022)Bai, Jones, Ndousse, Askell, Chen, DasSarma, Drain, Fort, Ganguli, Henighan, Joseph, Kadavath, Kernion, Conerly, El-Showk, Elhage, Hatfield-Dodds, Hernandez, Hume, Johnston, Kravec, Lovitt, Nanda, Olsson, Amodei, Brown, Clark, McCandlish, Olah, Mann, and Kaplan}]{bai2022traininghelpfulharmlessassistant}
Yuntao Bai, Andy Jones, Kamal Ndousse, Amanda Askell, Anna Chen, Nova DasSarma, Dawn Drain, Stanislav Fort, Deep Ganguli, Tom Henighan, Nicholas Joseph, Saurav Kadavath, Jackson Kernion, Tom Conerly, Sheer El-Showk, Nelson Elhage, Zac Hatfield-Dodds, Danny Hernandez, Tristan Hume, Scott Johnston, Shauna Kravec, Liane Lovitt, Neel Nanda, Catherine Olsson, Dario Amodei, Tom Brown, Jack Clark, Sam McCandlish, Chris Olah, Ben Mann, and Jared Kaplan. 2022.
\newblock \href {https://arxiv.org/abs/2204.05862} {Training a helpful and harmless assistant with reinforcement learning from human feedback}.
\newblock \emph{Preprint}, arXiv:2204.05862.

\bibitem[{Bradley and Terry(1952)}]{bradley1952rank}
Ralph~Allan Bradley and Milton~E Terry. 1952.
\newblock Rank analysis of incomplete block designs: I. the method of paired comparisons.
\newblock \emph{Biometrika}, 39(3/4):324--345.

\bibitem[{Chen et~al.(2024{\natexlab{a}})Chen, Chen, Zhang, Wang, Liu, Zhou, Zhang, Wan, Zhou, and Sun}]{chen2024mllmasajudge}
Dongping Chen, Ruoxi Chen, Shilin Zhang, Yaochen Wang, Yinuo Liu, Huichi Zhou, Qihui Zhang, Yao Wan, Pan Zhou, and Lichao Sun. 2024{\natexlab{a}}.
\newblock \href {https://openreview.net/forum?id=dbFEFHAD79} {{MLLM}-as-a-judge: Assessing multimodal {LLM}-as-a-judge with vision-language benchmark}.
\newblock In \emph{Forty-first International Conference on Machine Learning}.

\bibitem[{Chen et~al.(2024{\natexlab{b}})Chen, Li, Dong, Zhang, He, Wang, Zhao, and Lin}]{chen2024sharegpt4v}
Lin Chen, Jinsong Li, Xiaoyi Dong, Pan Zhang, Conghui He, Jiaqi Wang, Feng Zhao, and Dahua Lin. 2024{\natexlab{b}}.
\newblock Sharegpt4v: Improving large multi-modal models with better captions.
\newblock In \emph{European Conference on Computer Vision}, pages 370--387. Springer.

\bibitem[{Cui et~al.(2024)Cui, Yuan, Ding, Yao, He, Zhu, Ni, Xie, Xie, Lin, Liu, and Sun}]{cui2024ultrafeedback}
Ganqu Cui, Lifan Yuan, Ning Ding, Guanming Yao, Bingxiang He, Wei Zhu, Yuan Ni, Guotong Xie, Ruobing Xie, Yankai Lin, Zhiyuan Liu, and Maosong Sun. 2024.
\newblock \href {https://openreview.net/forum?id=BOorDpKHiJ} {{ULTRAFEEDBACK}: Boosting language models with scaled {AI} feedback}.
\newblock In \emph{Forty-first International Conference on Machine Learning}.

\bibitem[{Deitke et~al.(2024)Deitke, Clark, Lee, Tripathi, Yang, Park, Salehi, Muennighoff, Lo, Soldaini et~al.}]{deitke2024molmo}
Matt Deitke, Christopher Clark, Sangho Lee, Rohun Tripathi, Yue Yang, Jae~Sung Park, Mohammadreza Salehi, Niklas Muennighoff, Kyle Lo, Luca Soldaini, et~al. 2024.
\newblock Molmo and pixmo: Open weights and open data for state-of-the-art multimodal models.
\newblock \emph{arXiv preprint arXiv:2409.17146}.

\bibitem[{Dubey et~al.(2024)Dubey, Jauhri, Pandey, Kadian, Al-Dahle, Letman, Mathur, Schelten, Yang, Fan et~al.}]{dubey2024llama}
Abhimanyu Dubey, Abhinav Jauhri, Abhinav Pandey, Abhishek Kadian, Ahmad Al-Dahle, Aiesha Letman, Akhil Mathur, Alan Schelten, Amy Yang, Angela Fan, et~al. 2024.
\newblock The llama 3 herd of models.
\newblock \emph{arXiv preprint arXiv:2407.21783}.

\bibitem[{Dubois et~al.(2023)Dubois, Li, Taori, Zhang, Gulrajani, Ba, Guestrin, Liang, and Hashimoto}]{dubois2023alpacafarm}
Yann Dubois, Xuechen Li, Rohan Taori, Tianyi Zhang, Ishaan Gulrajani, Jimmy Ba, Carlos Guestrin, Percy Liang, and Tatsunori Hashimoto. 2023.
\newblock \href {https://openreview.net/forum?id=4hturzLcKX} {Alpacafarm: A simulation framework for methods that learn from human feedback}.
\newblock In \emph{Thirty-seventh Conference on Neural Information Processing Systems}.

\bibitem[{Ethayarajh et~al.(2022)Ethayarajh, Choi, and Swayamdipta}]{pmlr-v162-ethayarajh22a}
Kawin Ethayarajh, Yejin Choi, and Swabha Swayamdipta. 2022.
\newblock Understanding dataset difficulty with $\mathcal{V}$-usable information.
\newblock In \emph{Proceedings of the 39th International Conference on Machine Learning}, volume 162 of \emph{Proceedings of Machine Learning Research}, pages 5988--6008. PMLR.

\bibitem[{Goddard et~al.(2024)Goddard, Siriwardhana, Ehghaghi, Meyers, Karpukhin, Benedict, McQuade, and Solawetz}]{goddard-etal-2024-arcees}
Charles Goddard, Shamane Siriwardhana, Malikeh Ehghaghi, Luke Meyers, Vladimir Karpukhin, Brian Benedict, Mark McQuade, and Jacob Solawetz. 2024.
\newblock \href {https://doi.org/10.18653/v1/2024.emnlp-industry.36} {Arcee`s {M}erge{K}it: A toolkit for merging large language models}.
\newblock In \emph{Proceedings of the 2024 Conference on Empirical Methods in Natural Language Processing: Industry Track}, pages 477--485, Miami, Florida, US. Association for Computational Linguistics.

\bibitem[{Hurst et~al.(2024)Hurst, Lerer, Goucher, Perelman, Ramesh, Clark, Ostrow, Welihinda, Hayes, Radford et~al.}]{hurst2024gpt}
Aaron Hurst, Adam Lerer, Adam~P Goucher, Adam Perelman, Aditya Ramesh, Aidan Clark, AJ~Ostrow, Akila Welihinda, Alan Hayes, Alec Radford, et~al. 2024.
\newblock Gpt-4o system card.
\newblock \emph{arXiv preprint arXiv:2410.21276}.

\bibitem[{Ilharco et~al.(2023)Ilharco, Ribeiro, Wortsman, Schmidt, Hajishirzi, and Farhadi}]{ilharco2023editing}
Gabriel Ilharco, Marco~Tulio Ribeiro, Mitchell Wortsman, Ludwig Schmidt, Hannaneh Hajishirzi, and Ali Farhadi. 2023.
\newblock \href {https://openreview.net/forum?id=6t0Kwf8-jrj} {Editing models with task arithmetic}.
\newblock In \emph{The Eleventh International Conference on Learning Representations}.

\bibitem[{Ivison et~al.(2024)Ivison, Wang, Liu, Wu, Pyatkin, Lambert, Smith, Choi, and Hajishirzi}]{ivison2024unpacking}
Hamish Ivison, Yizhong Wang, Jiacheng Liu, Zeqiu Wu, Valentina Pyatkin, Nathan Lambert, Noah~A. Smith, Yejin Choi, and Hannaneh Hajishirzi. 2024.
\newblock \href {https://openreview.net/forum?id=JMBWTlazjW} {Unpacking {DPO} and {PPO}: Disentangling best practices for learning from preference feedback}.
\newblock In \emph{The Thirty-eighth Annual Conference on Neural Information Processing Systems}.

\bibitem[{Ivison et~al.(2023)Ivison, Wang, Pyatkin, Lambert, Peters, Dasigi, Jang, Wadden, Smith, Beltagy et~al.}]{ivison2023camels}
Hamish Ivison, Yizhong Wang, Valentina Pyatkin, Nathan Lambert, Matthew Peters, Pradeep Dasigi, Joel Jang, David Wadden, Noah~A Smith, Iz~Beltagy, et~al. 2023.
\newblock Camels in a changing climate: Enhancing lm adaptation with tulu 2.
\newblock \emph{arXiv preprint arXiv:2311.10702}.

\bibitem[{Kim et~al.(2024)Kim, Suk, Longpre, Lin, Shin, Welleck, Neubig, Lee, Lee, and Seo}]{kim-etal-2024-prometheus}
Seungone Kim, Juyoung Suk, Shayne Longpre, Bill~Yuchen Lin, Jamin Shin, Sean Welleck, Graham Neubig, Moontae Lee, Kyungjae Lee, and Minjoon Seo. 2024.
\newblock \href {https://doi.org/10.18653/v1/2024.emnlp-main.248} {Prometheus 2: An open source language model specialized in evaluating other language models}.
\newblock In \emph{Proceedings of the 2024 Conference on Empirical Methods in Natural Language Processing}, pages 4334--4353, Miami, Florida, USA. Association for Computational Linguistics.

\bibitem[{K\"{o}pf et~al.(2023)K\"{o}pf, Kilcher, von R\"{u}tte, Anagnostidis, Tam, Stevens, Barhoum, Nguyen, Stanley, Nagyfi, ES, Suri, Glushkov, Dantuluri, Maguire, Schuhmann, Nguyen, and Mattick}]{oasst}
Andreas K\"{o}pf, Yannic Kilcher, Dimitri von R\"{u}tte, Sotiris Anagnostidis, Zhi~Rui Tam, Keith Stevens, Abdullah Barhoum, Duc Nguyen, Oliver Stanley, Rich\'{a}rd Nagyfi, Shahul ES, Sameer Suri, David Glushkov, Arnav Dantuluri, Andrew Maguire, Christoph Schuhmann, Huu Nguyen, and Alexander Mattick. 2023.
\newblock \href {https://proceedings.neurips.cc/paper_files/paper/2023/file/949f0f8f32267d297c2d4e3ee10a2e7e-Paper-Datasets_and_Benchmarks.pdf} {Openassistant conversations - democratizing large language model alignment}.
\newblock In \emph{Advances in Neural Information Processing Systems}, volume~36, pages 47669--47681. Curran Associates, Inc.

\bibitem[{Lambert et~al.(2024)Lambert, Morrison, Pyatkin, Huang, Ivison, Brahman, Miranda, Liu, Dziri, Lyu et~al.}]{lambert2024t}
Nathan Lambert, Jacob Morrison, Valentina Pyatkin, Shengyi Huang, Hamish Ivison, Faeze Brahman, Lester James~V Miranda, Alisa Liu, Nouha Dziri, Shane Lyu, et~al. 2024.
\newblock T$\backslash$" ulu 3: Pushing frontiers in open language model post-training.
\newblock \emph{arXiv preprint arXiv:2411.15124}.

\bibitem[{Lee et~al.(2024)Lee, Kim, Park, Kim, and Seo}]{lee-etal-2024-prometheus}
Seongyun Lee, Seungone Kim, Sue Park, Geewook Kim, and Minjoon Seo. 2024.
\newblock \href {https://doi.org/10.18653/v1/2024.findings-acl.672} {Prometheus-vision: Vision-language model as a judge for fine-grained evaluation}.
\newblock In \emph{Findings of the Association for Computational Linguistics: ACL 2024}, pages 11286--11315, Bangkok, Thailand. Association for Computational Linguistics.

\bibitem[{Li et~al.(2024{\natexlab{a}})Li, Wei, Xie, Yang, Song, Wang, An, Liu, Li, Lin et~al.}]{li2024vlrewardbench}
Lei Li, Yuancheng Wei, Zhihui Xie, Xuqing Yang, Yifan Song, Peiyi Wang, Chenxin An, Tianyu Liu, Sujian Li, Bill~Yuchen Lin, et~al. 2024{\natexlab{a}}.
\newblock Vlrewardbench: A challenging benchmark for vision-language generative reward models.
\newblock \emph{arXiv preprint arXiv:2411.17451}.

\bibitem[{Li et~al.(2024{\natexlab{b}})Li, Xie, Li, Chen, Wang, Chen, Yang, Wang, Kong, and Liu}]{li-etal-2024-vlfeedback}
Lei Li, Zhihui Xie, Mukai Li, Shunian Chen, Peiyi Wang, Liang Chen, Yazheng Yang, Benyou Wang, Lingpeng Kong, and Qi~Liu. 2024{\natexlab{b}}.
\newblock \href {https://doi.org/10.18653/v1/2024.emnlp-main.358} {{VLF}eedback: A large-scale {AI} feedback dataset for large vision-language models alignment}.
\newblock In \emph{Proceedings of the 2024 Conference on Empirical Methods in Natural Language Processing}, pages 6227--6246, Miami, Florida, USA. Association for Computational Linguistics.

\bibitem[{Li et~al.(2024{\natexlab{c}})Li, Xie, Li, Chen, Wang, Chen, Yang, Wang, Kong, and Liu}]{li2024vlfeedback}
Lei Li, Zhihui Xie, Mukai Li, Shunian Chen, Peiyi Wang, Liang Chen, Yazheng Yang, Benyou Wang, Lingpeng Kong, and Qi~Liu. 2024{\natexlab{c}}.
\newblock Vlfeedback: A large-scale ai feedback dataset for large vision-language models alignment.
\newblock In \emph{Proceedings of the 2024 Conference on Empirical Methods in Natural Language Processing}, pages 6227--6246.

\bibitem[{Lin et~al.(2024)Lin, Li, Lee, and Chen}]{lin-etal-2024-dogerm}
Tzu-Han Lin, Chen-An Li, Hung-yi Lee, and Yun-Nung Chen. 2024.
\newblock \href {https://doi.org/10.18653/v1/2024.emnlp-main.868} {{D}oge{RM}: Equipping reward models with domain knowledge through model merging}.
\newblock In \emph{Proceedings of the 2024 Conference on Empirical Methods in Natural Language Processing}, pages 15506--15524, Miami, Florida, USA. Association for Computational Linguistics.

\bibitem[{Lu et~al.(2024)Lu, Liu, Zhang, Wang, Dong, Liu, Sun, Ren, Li, Yang et~al.}]{lu2024deepseek}
Haoyu Lu, Wen Liu, Bo~Zhang, Bingxuan Wang, Kai Dong, Bo~Liu, Jingxiang Sun, Tongzheng Ren, Zhuoshu Li, Hao Yang, et~al. 2024.
\newblock Deepseek-vl: towards real-world vision-language understanding.
\newblock \emph{arXiv preprint arXiv:2403.05525}.

\bibitem[{OpenAI(2023)}]{openai2023gpt4v}
OpenAI. 2023.
\newblock \href {https://openai.com/index/gpt-4v-system-card/} {Gpt-4v system card}.
\newblock Accessed: 2025-02-04.

\bibitem[{Ouyang et~al.(2022)Ouyang, Wu, Jiang, Almeida, Wainwright, Mishkin, Zhang, Agarwal, Slama, Ray et~al.}]{ouyang2022training}
Long Ouyang, Jeffrey Wu, Xu~Jiang, Diogo Almeida, Carroll Wainwright, Pamela Mishkin, Chong Zhang, Sandhini Agarwal, Katarina Slama, Alex Ray, et~al. 2022.
\newblock Training language models to follow instructions with human feedback.
\newblock \emph{Advances in neural information processing systems}, 35:27730--27744.

\bibitem[{Rame et~al.(2024)Rame, Vieillard, Hussenot, Dadashi, Cideron, Bachem, and Ferret}]{rame2024warm}
Alexandre Rame, Nino Vieillard, Leonard Hussenot, Robert Dadashi, Geoffrey Cideron, Olivier Bachem, and Johan Ferret. 2024.
\newblock \href {https://openreview.net/forum?id=s7RDnNUJy6} {{WARM}: On the benefits of weight averaged reward models}.
\newblock In \emph{Forty-first International Conference on Machine Learning}.

\bibitem[{Schulman et~al.(2017)Schulman, Wolski, Dhariwal, Radford, and Klimov}]{schulman2017proximal}
John Schulman, Filip Wolski, Prafulla Dhariwal, Alec Radford, and Oleg Klimov. 2017.
\newblock Proximal policy optimization algorithms.
\newblock \emph{arXiv preprint arXiv:1707.06347}.

\bibitem[{Singh et~al.(2019)Singh, Natarajan, Shah, Jiang, Chen, Batra, Parikh, and Rohrbach}]{singh2019towards}
Amanpreet Singh, Vivek Natarajan, Meet Shah, Yu~Jiang, Xinlei Chen, Dhruv Batra, Devi Parikh, and Marcus Rohrbach. 2019.
\newblock Towards vqa models that can read.
\newblock In \emph{Proceedings of the IEEE/CVF conference on computer vision and pattern recognition}, pages 8317--8326.

\bibitem[{Stiennon et~al.(2020)Stiennon, Ouyang, Wu, Ziegler, Lowe, Voss, Radford, Amodei, and Christiano}]{learningtosummarize}
Nisan Stiennon, Long Ouyang, Jeffrey Wu, Daniel Ziegler, Ryan Lowe, Chelsea Voss, Alec Radford, Dario Amodei, and Paul~F Christiano. 2020.
\newblock \href {https://proceedings.neurips.cc/paper_files/paper/2020/file/1f89885d556929e98d3ef9b86448f951-Paper.pdf} {Learning to summarize with human feedback}.
\newblock In \emph{Advances in Neural Information Processing Systems}, volume~33, pages 3008--3021. Curran Associates, Inc.

\bibitem[{Sun et~al.(2024)Sun, Shen, Cao, Liu, Li, Shen, Gan, Gui, Wang, Yang, Keutzer, and Darrell}]{sun-etal-2024-aligning}
Zhiqing Sun, Sheng Shen, Shengcao Cao, Haotian Liu, Chunyuan Li, Yikang Shen, Chuang Gan, Liangyan Gui, Yu-Xiong Wang, Yiming Yang, Kurt Keutzer, and Trevor Darrell. 2024.
\newblock \href {https://doi.org/10.18653/v1/2024.findings-acl.775} {Aligning large multimodal models with factually augmented {RLHF}}.
\newblock In \emph{Findings of the Association for Computational Linguistics: ACL 2024}, pages 13088--13110, Bangkok, Thailand. Association for Computational Linguistics.

\bibitem[{Team et~al.(2024)Team, Georgiev, Lei, Burnell, Bai, Gulati, Tanzer, Vincent, Pan, Wang et~al.}]{team2024gemini}
Gemini Team, Petko Georgiev, Ving~Ian Lei, Ryan Burnell, Libin Bai, Anmol Gulati, Garrett Tanzer, Damien Vincent, Zhufeng Pan, Shibo Wang, et~al. 2024.
\newblock Gemini 1.5: Unlocking multimodal understanding across millions of tokens of context.
\newblock \emph{arXiv preprint arXiv:2403.05530}.

\bibitem[{Team(2025)}]{Qwen2.5-VL}
Qwen Team. 2025.
\newblock \href {https://qwenlm.github.io/blog/qwen2.5-vl/} {Qwen2.5-vl}.

\bibitem[{Wang et~al.(2024)Wang, Dong, Delalleau, Zeng, Shen, Egert, Zhang, Sreedhar, and Kuchaiev}]{wang2024helpsteer}
Zhilin Wang, Yi~Dong, Olivier Delalleau, Jiaqi Zeng, Gerald Shen, Daniel Egert, Jimmy~J. Zhang, Makesh~Narsimhan Sreedhar, and Oleksii Kuchaiev. 2024.
\newblock \href {https://openreview.net/forum?id=PvVKUFhaNy} {Helpsteer 2: Open-source dataset for training top-performing reward models}.
\newblock In \emph{The Thirty-eight Conference on Neural Information Processing Systems Datasets and Benchmarks Track}.

\bibitem[{Wijaya et~al.(2024)Wijaya, Nguyen, and Cheung}]{wijaya2024multimodalpreferencedatasynthetic}
Robert Wijaya, Ngoc-Bao Nguyen, and Ngai-Man Cheung. 2024.
\newblock \href {https://arxiv.org/abs/2412.17417} {Multimodal preference data synthetic alignment with reward model}.
\newblock \emph{Preprint}, arXiv:2412.17417.

\bibitem[{Wortsman et~al.(2022)Wortsman, Ilharco, Gadre, Roelofs, Gontijo-Lopes, Morcos, Namkoong, Farhadi, Carmon, Kornblith et~al.}]{wortsman2022model}
Mitchell Wortsman, Gabriel Ilharco, Samir~Ya Gadre, Rebecca Roelofs, Raphael Gontijo-Lopes, Ari~S Morcos, Hongseok Namkoong, Ali Farhadi, Yair Carmon, Simon Kornblith, et~al. 2022.
\newblock Model soups: averaging weights of multiple fine-tuned models improves accuracy without increasing inference time.
\newblock In \emph{International conference on machine learning}, pages 23965--23998. PMLR.

\bibitem[{Xiao et~al.(2024)Xiao, Huang, Gan, He, Li, Yu, Jiang, Wu, and Zhu}]{xiao2024detecting}
Wenyi Xiao, Ziwei Huang, Leilei Gan, Wanggui He, Haoyuan Li, Zhelun Yu, Hao Jiang, Fei Wu, and Linchao Zhu. 2024.
\newblock Detecting and mitigating hallucination in large vision language models via fine-grained ai feedback.
\newblock \emph{arXiv preprint arXiv:2404.14233}.

\bibitem[{Yadav et~al.(2024)Yadav, Tam, Choshen, Raffel, and Bansal}]{yadav2024ties}
Prateek Yadav, Derek Tam, Leshem Choshen, Colin~A Raffel, and Mohit Bansal. 2024.
\newblock Ties-merging: Resolving interference when merging models.
\newblock \emph{Advances in Neural Information Processing Systems}, 36.

\bibitem[{Yang et~al.(2024)Yang, Shen, Guo, Wang, Cao, Zhang, and Tao}]{yang2024modelmergingllmsmllms}
Enneng Yang, Li~Shen, Guibing Guo, Xingwei Wang, Xiaochun Cao, Jie Zhang, and Dacheng Tao. 2024.
\newblock \href {https://arxiv.org/abs/2408.07666} {Model merging in llms, mllms, and beyond: Methods, theories, applications and opportunities}.
\newblock \emph{Preprint}, arXiv:2408.07666.

\bibitem[{Yu et~al.(2024{\natexlab{a}})Yu, Yu, Yu, Huang, and Li}]{yu2024language}
Le~Yu, Bowen Yu, Haiyang Yu, Fei Huang, and Yongbin Li. 2024{\natexlab{a}}.
\newblock Language models are super mario: Absorbing abilities from homologous models as a free lunch.
\newblock In \emph{Forty-first International Conference on Machine Learning}.

\bibitem[{Yu et~al.(2024{\natexlab{b}})Yu, Yao, Zhang, He, Han, Cui, Hu, Liu, Zheng, Sun, and Chua}]{Yu_2024_CVPR}
Tianyu Yu, Yuan Yao, Haoye Zhang, Taiwen He, Yifeng Han, Ganqu Cui, Jinyi Hu, Zhiyuan Liu, Hai-Tao Zheng, Maosong Sun, and Tat-Seng Chua. 2024{\natexlab{b}}.
\newblock Rlhf-v: Towards trustworthy mllms via behavior alignment from fine-grained correctional human feedback.
\newblock In \emph{Proceedings of the IEEE/CVF Conference on Computer Vision and Pattern Recognition (CVPR)}, pages 13807--13816.

\bibitem[{Yu et~al.(2024{\natexlab{c}})Yu, Zhang, Yao, Dang, Chen, Lu, Cui, He, Liu, Chua, and Sun}]{yu2024rlaifv}
Tianyu Yu, Haoye Zhang, Yuan Yao, Yunkai Dang, Da~Chen, Xiaoman Lu, Ganqu Cui, Taiwen He, Zhiyuan Liu, Tat-Seng Chua, and Maosong Sun. 2024{\natexlab{c}}.
\newblock Rlaif-v: Aligning mllms through open-source ai feedback for super gpt-4v trustworthiness.
\newblock \emph{arXiv preprint arXiv:2405.17220}.

\bibitem[{Yue et~al.(2024{\natexlab{a}})Yue, Ni, Zhang, Zheng, Liu, Zhang, Stevens, Jiang, Ren, Sun et~al.}]{yue2024mmmu1}
Xiang Yue, Yuansheng Ni, Kai Zhang, Tianyu Zheng, Ruoqi Liu, Ge~Zhang, Samuel Stevens, Dongfu Jiang, Weiming Ren, Yuxuan Sun, et~al. 2024{\natexlab{a}}.
\newblock Mmmu: A massive multi-discipline multimodal understanding and reasoning benchmark for expert agi.
\newblock In \emph{Proceedings of the IEEE/CVF Conference on Computer Vision and Pattern Recognition}, pages 9556--9567.

\bibitem[{Yue et~al.(2024{\natexlab{b}})Yue, Zheng, Ni, Wang, Zhang, Tong, Sun, Yu, Zhang, Sun et~al.}]{yue2024mmmu}
Xiang Yue, Tianyu Zheng, Yuansheng Ni, Yubo Wang, Kai Zhang, Shengbang Tong, Yuxuan Sun, Botao Yu, Ge~Zhang, Huan Sun, et~al. 2024{\natexlab{b}}.
\newblock Mmmu-pro: A more robust multi-discipline multimodal understanding benchmark.
\newblock \emph{arXiv preprint arXiv:2409.02813}.

\bibitem[{Zhang et~al.(2024)Zhang, Li, Zhang, Pu, Cahyono, Hu, Liu, Zhang, Yang, Li et~al.}]{zhang2024lmms}
Kaichen Zhang, Bo~Li, Peiyuan Zhang, Fanyi Pu, Joshua~Adrian Cahyono, Kairui Hu, Shuai Liu, Yuanhan Zhang, Jingkang Yang, Chunyuan Li, et~al. 2024.
\newblock Lmms-eval: Reality check on the evaluation of large multimodal models.
\newblock \emph{arXiv preprint arXiv:2407.12772}.

\bibitem[{Zhao et~al.(2024)Zhao, Ren, Hessel, Cardie, Choi, and Deng}]{zhao2024wildchat}
Wenting Zhao, Xiang Ren, Jack Hessel, Claire Cardie, Yejin Choi, and Yuntian Deng. 2024.
\newblock \href {https://openreview.net/forum?id=Bl8u7ZRlbM} {Wildchat: 1m chat{GPT} interaction logs in the wild}.
\newblock In \emph{The Twelfth International Conference on Learning Representations}.

\bibitem[{Zhou et~al.(2024)Zhou, Cui, Rafailov, Finn, and Yao}]{zhou2024aligning}
Yiyang Zhou, Chenhang Cui, Rafael Rafailov, Chelsea Finn, and Huaxiu Yao. 2024.
\newblock Aligning modalities in vision large language models via preference fine-tuning.
\newblock In \emph{ICLR 2024 Workshop on Reliable and Responsible Foundation Models}.

\bibitem[{Zhu et~al.(2024)Zhu, Frick, Wu, Zhu, Ganesan, Chiang, Zhang, and Jiao}]{zhu2024starlingb}
Banghua Zhu, Evan Frick, Tianhao Wu, Hanlin Zhu, Karthik Ganesan, Wei-Lin Chiang, Jian Zhang, and Jiantao Jiao. 2024.
\newblock \href {https://openreview.net/forum?id=GqDntYTTbk} {Starling-7b: Improving helpfulness and harmlessness with {RLAIF}}.
\newblock In \emph{First Conference on Language Modeling}.

\bibitem[{Zhu et~al.(2025)Zhu, Song, Shen, Zhao, Yang, Zhang, and Wu}]{zhu2025remedy}
Didi Zhu, Yibing Song, Tao Shen, Ziyu Zhao, Jinluan Yang, Min Zhang, and Chao Wu. 2025.
\newblock \href {https://openreview.net/forum?id=iX7eHHE5Tx} {{REMEDY}: Recipe merging dynamics in large vision-language models}.
\newblock In \emph{The Thirteenth International Conference on Learning Representations}.

\end{thebibliography}
